\documentclass[11pt]{article}

\usepackage[final]{acl}

\usepackage{times}
\usepackage{latexsym}

\usepackage[T1]{fontenc}

\usepackage[utf8]{inputenc}

\usepackage{microtype}

\usepackage{inconsolata}

\usepackage{graphicx}

\usepackage{adjustbox}
\usepackage{amssymb}
\usepackage{booktabs}
\usepackage{hhline}
\usepackage{multirow}
\usepackage{arydshln}
\usepackage{amsmath}

\newcommand{\lenstat}[2]{#1\,{\scriptsize$\pm$}\,#2}

\usepackage{listings}
\lstdefinelanguage{json}{
  morestring=[b]",
  morestring=[d]',
  stringstyle=\color{blue!70!black},
  keywordstyle=\color{red!60!black},
  commentstyle=\color{gray},
  identifierstyle=\color{black},
}

\usepackage[most]{tcolorbox}
\tcbuselibrary{listings, skins}

\newtcblisting{myjson}{
  breakable,                
  enhanced,                 
  colback=gray!5,
  colframe=black!50,
  arc=0pt,
  outer arc=0pt,
  boxrule=0.5pt,
  listing only,
  top=0pt,                  
  bottom=0pt,
  left=5mm,                 
  listing options={
    language=json,    
    basicstyle=\ttfamily\small,
    numbers=left,
    numberstyle=\tiny,
    stepnumber=1,
    numbersep=5pt,
    showstringspaces=false,
    breaklines=true,
    frame=none,
    columns=fullflexible
  }
}

\usepackage{tcolorbox}
\tcbset{colback=white,colframe=black!30,boxrule=0.5pt,arc=2pt,left=6pt,right=6pt,top=6pt,bottom=6pt}

\usepackage{pifont}
\title{Cross-Domain, Multi-Task Data-to-Text Generation without In-Domain Training Data}

\author{
  \textbf{Yifei Song},  \textbf{Kun Efimov-Zhang}, 
  \textbf{Claire Gardent} \\
  \textsuperscript{}CNRS/LORIA and Université de Lorraine \\
  \texttt{yifei.song@loria.fr} \quad
  \texttt{kun.zhang@inria.fr} \quad
  \texttt{claire.gardent@loria.fr}
}

\begin{document}
\maketitle

\begin{abstract}
Structured data exists in many forms (tables, knowledge graphs, charts, and time series), and converting it into text may involve different generation tasks. However, most prior work on data-to-text (D2T) generation has focused on specific tasks and datasets, relying either on task-specific training data or on the zero-shot capabilities of large language models. We study
cross-domain D2T generation in a setting where neither in-domain training text nor test references are available, and where domains, generation goals, and input structures vary substantially. We compare data-driven knowledge distillation (DDKD) against zero-shot inference and fine-tuning on out-of-domain D2T data, and introduce structure-preserving augmentation via structural subsampling and perturbation. Experiments on five benchmarks show that, at constant model size (1.7B parameters), DDKD consistently outperforms both fine-tuning and zero-shot inference. Moreover, the resulting small models outperform a much larger finetuned model on two of the five domains, achieving comparable performance on the remaining three. We further construct QUINTD-5, a fivefold extension of QUINTD-1, and show that simply scaling real target-domain inputs yields only modest gains, whereas our augmentation strategy remains more effective and more cost-efficient for cross-domain distillation.
The code and data are publicly available in our GitHub repository.\footnote{\url{https://github.com/MeloS7/Cross-Domain-D2T-DDKD}}
\end{abstract}

\section{Introduction}
\label{sec:intro}

Generation from structured data (data-to-text generation, D2T) has many applications. It can be used, for instance, to generate weather forecasts from \textit{time series} \cite{belz-2007-probabilistic, angeli-etal-2010-simple}
; to verbalise \textit{knowledge graphs} \citep{lebret-etal-2016-neural,castro-ferreira-etal-2020-2020,gardent-etal-2017-webnlg,song-etal-2025-multilingual-verbalisation} or \textit{records} \citep{kasner-dusek-2020-data,novikova-etal-2017-e2e,dusek-etal-2018-findings} 
; to generate game reports from  \textit{statistics} describing basketball games \citep{wiseman-etal-2017-challenges,Puduppully_Dong_Lapata_2019,10.1007/978-3-030-45439-5_5}; or to generate captions based on \textit{chart data} \citep{obeid-hoque-2020-chart,DBLP:journals/corr/abs-2110-05633,DBLP:conf/ijcnn/LiGCZSL23}.

\begin{figure}[!t]
    \centering
    \includegraphics[width=\columnwidth]{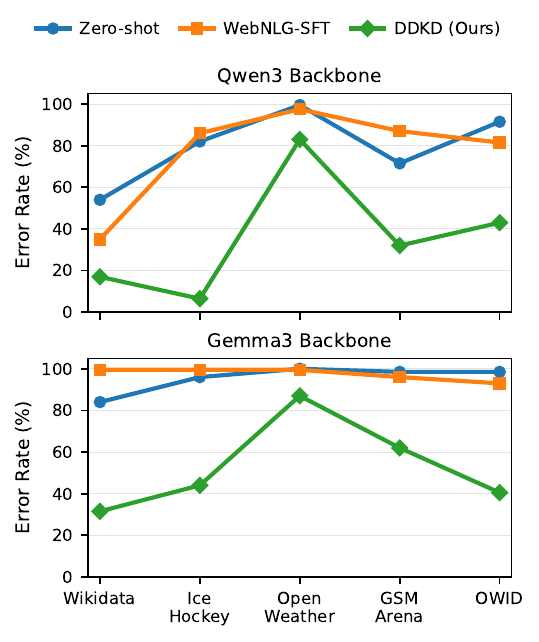}
    \caption{
    Per-domain error rate (\textbf{lower is better}) across the five QUINTD-1 target domains, averaged over two independent LLM judges (GPT-5.1 and Gemini-2.5-Pro).
    Results compare three \emph{small} models of identical scale:
    zero-shot inference, 
    fine-tuning on WebNLG, and our DDKD-distilled model.
    Across both Qwen3 and Gemma3 backbones, DDKD consistently yields substantially lower error rates, indicating improved cross-domain transfer without target-domain text supervision.
    }
    \label{fig:macro_error_main}
\end{figure}

In this work, we strive to generate text from different data structures and domains in a setting where neither training nor test reference texts are available. We confront several major challenges.
First, schema
diversity across domains leads to substantial structural differences
in input fields, nesting depth, and attribute semantics, making it
difficult for models to rely on a unified representation. Second, the
format heterogeneity (JSON, CSV, Markdown) introduces further
variation in how information is organised and encoded. Third, the absence of both training and test data precludes conventional supervised fine-tuning, reference-based evaluation, and large-scale self-training within the target domains. Fourth, the structured inputs in several
domains are often very long and information-dense, which makes
few-shot prompting impractical as fitting multiple in-context examples
alongside the long input exceeds the context limits of most models or
severely reduces the usable prompt space. Fifth, prior work by \citet{kasner-dusek-2024-beyond} has
shown that zero-shot generalisation across diverse domains and generation tasks 
remains limited, underscoring the need for methods that robustly
transfer D2T capabilities to new domains and formats. Specifically, they show that while the generated texts are generally fluent, 76 to 86\% of the texts generated by several open source 7B  models on five different domains contain at least one factual or semantic error.


Using the QUINTD-1 dataset introduced in \citet{kasner-dusek-2024-beyond}, we investigate four main approaches to generating English text from heterogeneous data types (i.e., time series, sets of triples/dictionaries, tables, and charts) and multiple domains (i.e., weather reports, sports game summaries, technical and generic object descriptions, and health data captioning). The approaches considered are: (i) zero-shot prompting (ZS), (ii) fine-tuning LLMs on out-of-domain data-to-text data (SFT), and (iii--iv) data-driven knowledge distillation (DDKD), where synthetic target-domain training data are generated either by a zero-shot LLM teacher (KD-ZS) or by a D2T-finetuned LLM teacher (KD-SFT). Beyond this comparison, we ask two additional questions: whether structure-preserving augmentation is more effective than simply scaling the amount of real target-domain structured inputs, and whether lower error rates are achieved without sacrificing content coverage.


Intuitively, these four approaches differ in the quantity and type of supervision they exploit. A zero-shot LLM relies only on D2T knowledge implicitly acquired during the LLM pre- and post-training stages. An SFT model additionally receives explicit out-of-domain D2T supervision, which in our case consists of 40K knowledge graph--text pairs from WebNLG \cite{gardent-etal-2017-webnlg}. Finally, distilled student models further benefit from synthetic target-domain D2T examples generated for each domain. 



We hypothesise that DDKD from a D2T-finetuned teacher will be the most effective setting, as it combines broad D2T supervision from WebNLG with target-domain, structure-specific synthetic examples generated for the domains of interest.

As shown in Figure~\ref{fig:macro_error_main}, across both model families (Gemma3-1B and Qwen3-1.7B), DDKD consistently yields lower error rates than the corresponding same-size zero-shot and WebNLG-finetuned baselines across all five target domains.


We make the following contributions:
\begin{itemize}
    \item We investigate Data-Driven Knowledge Distillation  approaches for cross-domain data-to-text generation without target-domain reference text. 
    \item We introduce structure-preserving target-domain input augmentations to improve robustness to schema and format heterogeneity, including structural subsampling and perturbation.
\item We construct \textbf{QUINTD-5}, a fivefold extension of QUINTD-1 and use it to show that structure-preserving augmentation is more effective and more cost-efficient than simply scaling real target-domain inputs.
\item We provide a rigorous reference-free evaluation combining error-taxonomy LLM-as-a-Judge assessment with two independent judges, human validation and agreement analysis, and a content-coverage sanity check showing that DDKD improves faithfulness without relying on conservative under-generation
\end{itemize}

\begin{figure*}
    \centering
    \includegraphics[width=0.95\linewidth]{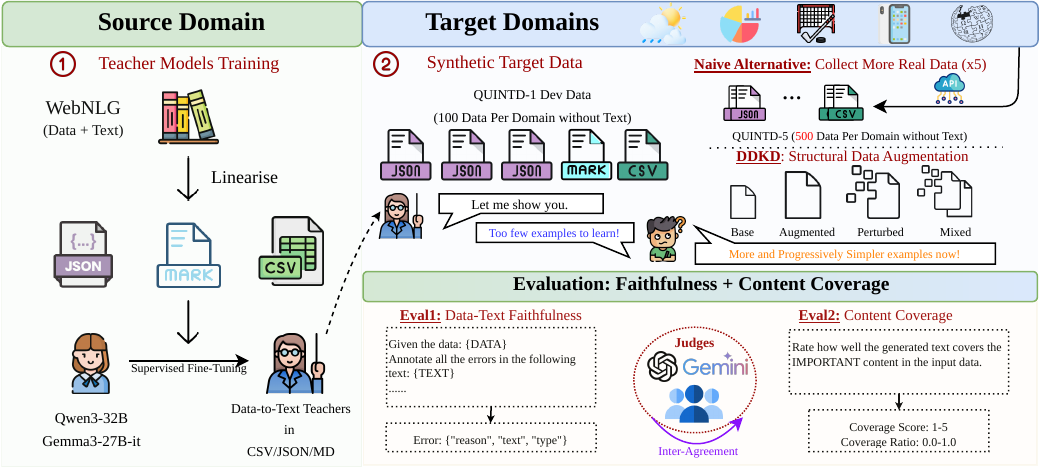}
    \caption{
\textbf{Overview of our cross-domain Data-Driven Knowledge Distillation framework.}
Large teacher models are first trained on source-domain WebNLG using multiple structured input linearizations.
Given limited target-domain structured inputs (QUINTD-1), the teacher generates synthetic texts for distillation.
We compare two ways of improving target-domain supervision: collecting more real structured inputs as a naive baseline (QUINTD-5), and our proposed \textbf{structural data augmentation}, which increases diversity and yields progressively simpler training examples.
A smaller student model is then distilled on the resulting data and evaluated with LLM judges and human annotation for faithfulness and content coverage, together with inter-judge agreement analysis.
}
    \label{fig:ddkd-approach}
\end{figure*}

\section{Related Work}
\label{sec:related}

\paragraph{Data-to-Text Generation.}
\label{subsec:d2t}
Research on D2T generation has considered various types of structured inputs, 
such as dependency structures \citep{mille-etal-2023-pipeline}, AMRs \citep{konstas-etal-2017-neural,knight2021abstract}, knowledge graphs \citep{gardent-etal-2017-webnlg,castro-ferreira-etal-2020-2020}, time series \citep{narasimhan2024time}, or tables \citep{table2txt2018wang,parikh-etal-2020-totto,xing-wan-2021-structure,11228570}. However, most of this work has focused on a single or similar types of structures\footnote{E.g., considering different types of data structures for datasets where the input instances (Knowledge Graph, Simple tables, Sets of records) can all be converted into sets of triples \citep{song-etal-2025-multilingual-verbalisation}}. 
Exceptions include \citet{xiang-etal-2022-asdot} who investigate decomposition strategies and multi-stage generation to improve generalisation and  
\citet{li-etal-2023-shot-data} who propose a unified framework to handle tables, graphs, and meaning representations through multi-source learning. Both approaches, however, rely on existing training data and are therefore limited in their applicability to novel D2T tasks involving real-world data. Closest to our work, \citet{kasner-dusek-2024-beyond} analyse the performance of LLMs across five domains with different input structures and on different tasks. We build on their work by investigating how different modelling strategies 
improve performance on the datasets they introduce. 

\paragraph{Knowledge Distillation.}
\label{subsec:kd}
Knowledge distillation, pioneered by \citet{hinton2015distillingknowledgeneuralnetwork}, enables compact student models to achieve competitive or superior performance by learning from teacher models.
This technique was widely applied in computer vision \citep{xu2020knowledgedistillationmeetsselfsupervision, yang2022mixskdselfknowledgedistillationmixup}, then extended to natural language processing tasks including compressing pre-trained language models \citep{sanh2020distilbertdistilledversionbert, jiao-etal-2020-tinybert, minilm}.
In natural language generation, distillation methods have been applied to machine translation \citep{kim-rush-2016-sequence, gu2018nonautoregressive}, summarisation \citep{liu-etal-2021-noisy, xu-etal-2023-inheritsumm, pham-etal-2023-select, song-etal-2023-enhancing}, dialogue generation \citep{zhu-etal-2021-combining-curriculum-learning,chae-etal-2023-dialogue}, and paraphrasing \citep{jung-etal-2024-impossible}. 
\citet{calderon-etal-2023-systematic} propose pseudo-target training as a general framework for task-specific distillation across NLG tasks. 
Recent LLM distillation introduces new approaches: \citet{hsieh-etal-2023-distilling} extract step-by-step rationales as additional supervision; 
\citet{agarwal2024onpolicy} develop on-policy learning from self-generated sequences; 
\citet{liu2024ddk} explore domain-specific transfer strategies.
For data-to-text, \citet{yang-etal-2024-effective} distil reasoning abilities from LLMs for scientific table-to-text, \citet{bai-etal-2025-reasoning} introduce reasoning knowledge filtering for logical table-to-text, and \citet{piedrahita2024enhancing} decompose graph-to-text into reasoning steps for chain-of-thought distillation.
We differ from these approaches in that we tackle cross-domain data-to-text generation without target-domain references, 
combining out-of-domain training, synthetic data generation, and structured perturbation 
to enable small models to generalise across diverse domains and formats.


\section{Cross-Domain D2T Generation}
\label{sec:task}

\begin{table*}[!htbp]
\small
\centering
\begin{tabular}{l l p{5.0cm} p{3.4cm} l}
\toprule
\textbf{Dataset} & \textbf{Format} & \textbf{Generation Goal} & \textbf{Structure Complexity} & \textbf{\# Input Tokens$^\dagger$} \\
\midrule
\textbf{Wikidata} & MD & Structured entity description & \#Props = 101, \#Vals = 418 & 122 (84-190) \\
\textbf{Ice Hockey} & JSON & Event-based game summary & \#Props = 20, \#Vals = 975 & 335 (315-366) \\
\textbf{OpenWeather} & JSON & Time-series weather forecast reporting & \#Props = 28, \#Vals = 3{,}782 & 3{,}294 (3{,}163-3{,}487) \\
\textbf{GSM Arena} & JSON & Entity-centric specification description & \#Props = 9, \#Vals = 2{,}361 & 1{,}280 (886-3{,}164) \\
\textbf{OWID} & CSV & Chart Captioning & \#Rows = 217, \#Cols = 2 & 3{,}742 (178-8{,}076) \\
\bottomrule
\end{tabular}
\caption{
\textbf{QUINTD-1 Benchmark and Generation Tasks}. \#Props, \#Vals: Number of distinct properties and Values present in each dataset. \# Input Tokens: Average (Min-Max) number of input tokens from Qwen3 tokenizer.
}
\label{tab:data_stats}
\end{table*}


\paragraph{Objective.}
Given a prompt $\mathcal{P}$ and some structured input $x$, the objective is to generate a fluent English text $y$ that faithfully reflects $x$'s factual content and follows $\mathcal{P}$'s instructions.

\paragraph{Input Data.}
We use the QUINTD-1 benchmark \citep{kasner-dusek-2024-beyond} which contains structured data collected from public APIs across five domains. QUINTD-1 does not include a training split: for each domain, it provides 100 development inputs and 100 test inputs, neither accompanied by in-domain reference texts. In this reference-free setting, we use the development inputs as seeds for constructing synthetic target-domain supervision and reserve the test inputs exclusively for evaluation. The resulting synthetic input--text pairs are internally split into training and validation subsets for model fitting and early stopping (Appendix~\ref{app:training_details}). 
As shown in Table~\ref{tab:data_stats}, these five domains vary substantially in terms of generation  objectives, input structures, and output formats.

\paragraph{Models.}
\label{sec:models}

We compare four methods to generate from different data structures in the absence of labelled data: zero-shot LLM prompting
, fine-tuning on data-to-text gold data 
and data-driven knowledge distillation where a small model (the student) is trained on synthetic data generated using either an LLM (DDKD from ZS Teacher) or an LLM finetuned on the WebNLG benchmark.

\paragraph{LLM Prompting (ZS).} \citet{kasner-dusek-2024-beyond} prompted three open weight 7B LLMs (Llama 2, Mistral, Zephyr) and GPT-3.5 on the QUINTD-1 benchmark, showing that according to both human annotators and GPT-4 as a judge, more than 80\% of the texts generated by open LLMs contain at least one semantic error. We re-run their experiment  using larger, up-to-date LLMs namely, Qwen3-32B and Gemma3-27B-IT \citep{yang2025qwen3technicalreport,gemmateam2025gemma3technicalreport}. 
We also use GPT-4.1 as a closed model for comparison
. However, since this model does not support reproducibility, we restrict fine-tuning and distillation experiments to open-weight models.

\paragraph{Fine-Tuning on WebNLG data.} Using LoRA, we fine-tune the open LLMs used in the ZS setting  on the WebNLG 3.0 training data (40K Knowledge Graph/English text pairs). 
This stage can be viewed as an initialization step, 
that equips the model with general data-to-text generation capabilities before adaptation to the five QUINTD-1 target domains.
We choose WebNLG because its human-written, fact-aligned references span diverse DBpedia categories, combining supervision quality with the semantic breadth needed for cross-domain transfer. By comparison, E2E~\citep{novikova-etal-2017-e2e} is also human-written but restricted to the restaurant domain, whereas KELM-Q1~\citep{song-gardent-2025-mucal} is a filtered Wikidata/Wikipedia subset whose references are automatically constructed and therefore likely to be noisy. A controlled source-initialization ablation using the same Qwen3-32B teacher (Appendix~\ref{app:init_ablation}) shows that WebNLG provides the best overall faithfulness--coverage trade-off: E2E yields competitive error counts on some simpler domains but substantially lower coverage, whereas KELM-Q1 performs worse on both metrics across most domains. 

\paragraph{DDKD from ZS and WebNLG-SFT Teachers.}
We adopt \textbf{Data-Driven Knowledge Distillation (DDKD)} to transfer cross-domain data-to-text generation ability from a large teacher model to a smaller student model. Figure~\ref{fig:ddkd-approach} illustrates our DDKD approach. 
Given a structured input $x$, the teacher first generates a synthetic target text $\tilde{y}$, and the student is then trained on the resulting $(x,\tilde{y})$ pairs.
Concretely, the student is optimized with standard maximum-likelihood estimation:
\(
\mathcal{L}_{\text{DDKD}} = - \log p_{\theta}(\tilde{y} \mid x),
\)
where $\theta$ denotes the student parameters.
Unlike logit-based or confidence-based distillation, DDKD relies solely on teacher-generated texts, making it simple and model-agnostic.
To construct the synthetic training data, we apply the teacher to target-domain structured inputs and further expand the resulting supervision data  using augmentation and perturbation methods (Section~\ref{sec:data}).
We consider two teacher variants: \textbf{DDKD from ZS teacher}, where the synthetic texts are generated by the zero-shot LLM, and \textbf{DDKD from WebNLG-SFT teacher}, where the synthetic texts are generated by the teacher fine-tuned on WebNLG.
This setup is particularly suitable for our target domains, where structured inputs are often long and information-dense (Table~\ref{tab:augmentation_stats}), making direct fine-tuning of large models substantially more expensive.
By shifting supervision transfer offline, the large teacher is only used for conditional generation, while all target-domain learning is performed on a compact student model that is much more efficient to train and deploy.

\section{Training Data}
\label{sec:data}

We use the WebNLG D2T data for fine-tuning. For DDKD, 
we create synthetic data and we collect additional real data from the QUINTD APIs.

\subsection{WebNLG Data.}
\label{subsec:fine-tuning-data}

The WebNLG training dataset contains 39,890 Knowledge Graph/English Text pairs spanning 16 distinct knowledge categories, where the graphs are extracted from DBpedia and paired with human-written texts describing their content. To reduce input-format mismatch between source-domain fine-tuning and target-domain transfer, we deterministically linearise each WebNLG graph into the three formats used in QUINTD-1: JSON, Markdown, and CSV. Table~\ref{tab:webnlg_linearisation_example} illustrates these representations for the same underlying graph.

\begin{table*}[t]
\centering
\small
\setlength{\tabcolsep}{6pt}
\renewcommand{\arraystretch}{1.08}
\begin{tabular}{p{0.16\linewidth} p{0.76\linewidth}}
\toprule
\textbf{Representation} & \textbf{Example} \\
\midrule

\textbf{WebNLG Triples}
&
\texttt{\{"subject": "Abilene\_Regional\_Airport", "property": "cityServed", "object": "Abilene,\_Texas"\}} \\
&
\texttt{\{"subject": "Abilene,\_Texas", "property": "isPartOf",  "object": "Texas"\}} \\

\midrule

\textbf{CSV}
&
\texttt{\# category: Airport} \\
&
\texttt{subject,property,object} \\
&
\texttt{Abilene Regional Airport,cityServed,Abilene; Texas} \\
&
\texttt{Abilene; Texas,isPartOf,Texas} \\

\midrule

\textbf{JSON}
&
\texttt{\{'Abilene\_Regional\_Airport': \{'cityServed': 'Abilene, Texas'\},} \\
&
\texttt{'Abilene,\_Texas': \{'isPartOf': 'Texas'\}\}} \\

\midrule

\textbf{Markdown}
&
\texttt{Abilene Regional Airport} \\
&
\texttt{-}\texttt{-}\texttt{-} \\
&
\texttt{- city served: Abilene, Texas} \\
&
\texttt{Abilene, Texas} \\
&
\texttt{-}\texttt{-}\texttt{-} \\
&
\texttt{- is part of: Texas} \\

\bottomrule
\end{tabular}
\caption{Example of deterministic WebNLG linearisation into the three target-domain input formats used in our experiments.}
\label{tab:webnlg_linearisation_example}
\end{table*}



\subsection{Target-domain Synthetic Data Generation}
To enable 
Data-Driven Knowledge Distillation, we construct synthetic training data from the QUINTD-1 development sets to  
increase both the quantity and structural diversity of available inputs while preserving data faithfulness.

\paragraph{Base Setting.}
As a baseline, we directly use the original 100 development instances from each QUINTD-1 domain without any augmentation. For each structured input record $x \in$ QUINTD-1, we employ the best-performing model (cf. Table~\ref{tab:final_ddkd_selection} and Section~\ref{app:model-selection}) finetuned on WebNLG to verbalize the input and generate a corresponding synthetic textual description $y$. This setting reflects a minimal cross-domain transfer scenario in which no additional data or structural variation is introduced. 

\begin{table}[ht]
\small
\centering
\setlength{\tabcolsep}{2pt}
\begin{tabular}{lcccr}
\toprule
\textbf{Domain} & \textbf{Base} & \textbf{Sub./Pert.} & \textbf{Mixed} & \textbf{Len. ($\mu \pm \sigma$)} \\
\midrule
Wikidata     & 100 & 32{,}628 & 65{,}256 & $173 \pm 52$ \\
Ice Hockey   & 100 & 25{,}500 & 51{,}000 & $421 \pm 28$ \\
OpenWeather  & 100 & 1{,}100  & 2{,}200  & $6{,}983 \pm 857$ \\
GSM Arena    & 100 & 1{,}381  & 2{,}762  & $1{,}601 \pm 643$ \\
OWID         & 100 & 1{,}000  & 2{,}000  & $5{,}067 \pm 3{,}355$ \\
\bottomrule
\end{tabular}
\caption{
\textbf{Statistics of the synthetic training data generated from the QUINTD-1 development sets under different augmentation strategies.} Base: the number of  QUINTD-1 instances. Sub./Pert.: the number of instances after  subsampling or structural perturbation. Mixed: the number of instances for the mixed strategy. Len: mean and standard deviation of input-output sequence lengths (in Qwen3 tokenizer tokens).
}
\label{tab:augmentation_stats}
\end{table}

\paragraph{Structured Subsampling Augmentation.}
To further enrich the target-domain data, we introduce a structured subsampling strategy that exploits the compositional nature of structured inputs. Given an input record consisting of an entity and a set of attribute-value pairs, we generate new training instances by extracting informative structural subsets.
Formally, let a structured input be represented as 
$x = (e, \mathcal{A}), \quad \mathcal{A} = \{(p_i, v_i)\}_{i=1}^{n}$, 
where $e$ denotes the central entity and $\mathcal{A}$ is a set of $n$ attribute--value pairs. We construct new inputs by selecting non-empty subsets $\mathcal{S} \subseteq \mathcal{A}$ and forming new instances
 $x_{\mathcal{S}} = (e, \mathcal{S}).$ 
Each $x_{\mathcal{S}}$ is 
then verbalized by the teacher model to produce a corresponding synthetic text $y_{\mathcal{S}}$.
Appendix \ref{app:sub_aug_detail} describes the sampling strategy used for each of the five domains in more detail.





\paragraph{Noise-based Structural Perturbation.}

While structured subsampling increases coverage over compositional input subsets, models trained solely on subsampled data may be subject to overfitting. 
To address this shortcoming, we introduce a noise-based structural perturbation strategy applied on top of the subsampled data. Our aim is to increase intra-sample variability, making individual data points less distinct from each other and preventing the network from fitting the training set too tightly.
For instance, for the OWID data, where inputs primarily consist of numeric values and temporal records, we apply in-instance perturbation by randomly modifying values within instances (e.g., modifying the birth rate for a given year).
 As for subsampling, we devise different perturbation strategies for the  various datasets. These are detailed in Appendix~\ref{app:noise_pert_detail}. In all cases, the perturbation ratio is fixed to 20\% of the selected instances. 





\paragraph{Combined Augmentation.}
Finally, we construct a mixed target-domain training set by combining the subsampled and perturbed synthetic instances and randomly shuffling them.
By distilling on a mixture of clean and structurally perturbed synthetic inputs, the student is exposed to diverse input distributions during training.
This encourages robustness to schema variation and partial structural noise, 
leading to more stable cross-domain generalisation in practice.

Table~\ref{tab:augmentation_stats} shows statistics for our synthetic data. 





\subsection{QUINTD-5: A Real-Data Scaling Control}
\label{subsec:quintd5}

Our main method strengthens target-domain distillation through structure-preserving augmentation. To compare this strategy against a more direct alternative which simply increases the size of the synthetic training data, we additionally construct \textbf{QUINTD-5}, a fivefold extension of QUINTD-1 that increases the number of real target-domain structured inputs from 100 to 500 per domain. We create QUINTD-5 using the same data collection pipeline, public APIs, and preprocessing as QUINTD-1, thereby preserving task definitions, input formats, and domain coverage (see Appendix \ref{app:quintd-5}).
QUINTD-5 permits assessing whether gains in cross-domain distillation are better obtained by adding more real target-domain inputs or by generating structure-preserving augmented supervision from a smaller seed set.

\section{Experiments}
\label{sec:experiments}

\subsection{Model Families and Ablation Studies}
\label{subsec:model_families}

Our experiments are conducted on two open-source model families, Qwen3 and Gemma3. Hyperparameters, experimental settings, and computational costs for each approach are detailed in Appendix~\ref{app:training_details}.

To study the effect of structured data augmentation and distillation strategies in detail, we perform a comprehensive ablation over augmentation variants using Qwen3 as our primary analysis model family. Based on these results (Table~\ref{tab:llm_judge_all_errnum_5domains_gpt}), we identify a set of robust and consistently effective augmentation strategies. For Gemma3, we adopt the best-performing augmentation configurations identified on Qwen3 and focus on evaluating cross-domain transfer performance rather than repeating the full ablation (Table~\ref{tab:llm_judge_all_errnum_5domains_gpt_gemma3}). This design allows us to assess the generality of our approach across two model families while keeping the experimental budget manageable. In addition to the main QUINTD-1 setting, we also evaluate a controlled real-data scaling condition based on QUINTD-5 and perform a content-coverage sanity check to further analyse the source of DDKD gains.




\subsection{Automatic Evaluation}
\label{subsec:evaln}
Since four of our five domains involve summarisation tasks (cf. Table~\ref{tab:data_stats}), 
and there are no reference texts against which we could evaluate the generated output, our primary automatic evaluation targets \emph{faithfulness} to the structured input. 
However, since lower error rates could in principle be achieved by conservative under-generation, for instance by producing shorter or less informative outputs, we conduct an additional \emph{content coverage} sanity check on the same subset used for human evaluation, using the same two LLM judges (GPT-5.1 and Gemini-2.5-Pro). Full prompts, scoring details, and per-domain results  for the coverage analysis are reported in Appendix~\ref{app:coverage}.


\paragraph{Evaluating Faithfulness.}
As the QUINTD-1 dataset does not provide reference texts,  
reference-based metrics such as BLEU or ROUGE are not applicable. 
Following prior work by \citet{kasner-dusek-2024-beyond}, we therefore adopt an \emph{LLM-as-a-Judge} protocol that evaluates generated outputs by directly comparing them against their corresponding structured inputs employing an error taxonomy which covers  four main types of errors: \emph{Incorrect Fact}, \emph{Not Checkable}, \emph{Misleading}, and \emph{Other}.
Appendix \ref{app:error-types} defines these error types and provides examples.
Multiple error types may co-occur within a single output. 
For each generated text, the judge identifies erroneous spans and assigns an error type to each.
We then aggregate these annotations and report the \emph{average number of errors per output} as the primary automatic evaluation metric.

\paragraph{Judges.}
We use GPT-5.1~\citep{singh2025openaigpt5card} as the primary judge for reporting results in the main paper.
To validate the robustness of LLM-based evaluation, we additionally evaluate all systems using a second independent reasoning-oriented judge, Gemini-2.5-Pro.
Compared to chat-oriented models, these reasoning-focused LLMs exhibit more stable behaviour when performing structured input–output comparison and error categorisation, particularly under heterogeneous schemas and long-context inputs. Detailed prompts and annotation guidelines are provided in Appendix~\ref{app:llm_judge_prompt}.

\begin{table*}[!htbp]
\centering
\small
\begin{tabular}{lcccccc}
\toprule
\textbf{System} &
\textbf{Wikidata} &
\textbf{Ice Hockey} &
\textbf{OpenWeather} &
\textbf{GSM Arena} &
\textbf{OWID} &
\textbf{NormAvg} $\downarrow$ \\
\midrule

\multicolumn{7}{l}{\textbf{\textit{Zero-shot (ZS)}}} \\
GPT-4.1-ZS      & 0.12 & 0.37 & 0.76 & 1.05 & 0.53 & 0.13 \\\hdashline
Qwen3-1.7B-ZS   & 0.74 & 1.38 & 6.10 & 1.22 & 2.11 & 0.64 \\ 
Qwen3-8B-ZS     & 0.27 & 0.59 & 3.37 & 1.17 & 1.20 & 0.30 \\
Qwen3-32B-ZS    & 0.45 & 0.63 & 2.53 & 0.49 & 0.99 & 0.25 \\

\hdashline
\multicolumn{7}{l}{\textbf{\textit{WebNLG-SFT}}} \\
Qwen3-1.7B-SFT  & 0.37 & 1.49 & 16.66 & 2.16 & 2.61 & 0.81 \\
Qwen3-8B-SFT    & 0.17 & 0.16 & 5.21 & 1.36 & 0.73 & 0.23 \\
Qwen3-32B-SFT   & \underline{0.09} & \underline{0.02} & \underline{1.86} & 0.65 & 0.43 & \underline{0.04} \\

\hdashline
\multicolumn{7}{l}{\textbf{\textit{DDKD from ZS teacher}}} \\
Qwen3-1.7B-DDKD-Base (100)   & 0.53 & 0.91 & 4.05 & 1.03 & 1.88 & 0.45 \\
Qwen3-1.7B-DDKD-Sub    & 0.45 & 0.66 & 2.68 & 0.67 & 1.34 & 0.30 \\
Qwen3-1.7B-DDKD-Pert   & 0.29 & 0.89 & 2.68 & \underline{\textbf{0.47}} & 1.49 & 0.27 \\
Qwen3-1.7B-DDKD-Mixed  & 0.33 & 0.73 & 2.88 & 0.65 & 1.39 & 0.28 \\
Qwen3-1.7B-DDKD-Base (500)   & 0.61 & 1.00 & 4.08 & 0.72 & 1.29 & 0.42 \\

\hdashline
\multicolumn{7}{l}{\textbf{\textit{DDKD from WebNLG-SFT teacher}}} \\
Qwen3-1.7B-DDKD-Base (100)& 0.25 & 0.26 & 6.78 & 1.38 & 3.94 & 0.47 \\
Qwen3-1.7B-DDKD-Sub    & 0.21 & 0.04 & \textbf{2.01} & 1.12 & 1.52 & 0.20 \\
Qwen3-1.7B-DDKD-Pert   & 0.18 & 0.12 & 4.88 & 1.25 & 0.45 & 0.19 \\
Qwen3-1.7B-DDKD-Mixed  & \textbf{0.10} & \textbf{0.03} & 4.47 & 0.87 & \underline{\textbf{0.36}} & \textbf{0.10} \\
Qwen3-1.7B-DDKD-Base (500)   & 0.26 & 0.04 & 2.79 & 1.14 & 1.12 & 0.20 \\

\bottomrule
\end{tabular}
\caption{
\textbf{LLM-as-a-judge evaluation (\textbf{all categories}): average number of errors per output on the five QUINTD-1 target domains (\emph{lower is better}).}
Results are based on GPT-5.1 as the judge.
For DDKD, unless otherwise specified, results use 100 real seed instances (QUINTD-1 setting); rows marked with (500) use 500 real seed instances (QUINTD-5 setting).
\textbf{Bold} indicates the best small model (1.7B) per domain and for NormAvg;
\underline{underline} indicates the best overall system.
\textbf{NormAvg} denotes the mean of per-domain min-max normalised error counts, where normalisation is performed independently for each domain across all systems.
}
\label{tab:llm_judge_all_errnum_5domains_gpt}
\end{table*}

\paragraph{LLM Judge Agreement}

To assess the reliability of LLM-based evaluation, we analyse the agreement between the two LLM judges (GPT-5.1 and Gemini-2.5-Pro) at the example and system levels based on Qwen-based outputs. 
All agreement scores are computed separately for each target domain and error category.
Detailed definitions and full agreement statistics are provided in Appendix~\ref{app:llm_judge_agree} and in Table~\ref{tab:llm_judge_agreement_full}. 

\paragraph{Coverage Sanity Check.}
\label{subsec:coverage_main}
We conduct an additional \emph{content coverage} sanity check on the same models used for human evaluation, using the same two LLM judges. 
Full prompts, scoring details, and per-domain results are reported in Appendix~\ref{app:coverage}.

\subsection{Human Evaluation}
\label{subsec:human-eval}

To further validate the reliability of LLM-based evaluation, we conduct a human evaluation on a carefully constructed diagnostic subset designed to support analysis along multiple dimensions, including comparisons across systems, domains, and error categories.
We sample 60 inputs (12 per domain) via stratified sampling over error categories, using GPT-5.1 annotations as anchors, and collect outputs from the best model among four main approaches (a small zero-shot model, a small source-domain SFT model, the best distilled model and the best large teacher) resulting in 240 outputs in total. In this way, we can compare zero-shot, finetuned and distilled models at constant size and with respect to a large LLM. 
Human annotators label each output at the example level by identifying the presence and severity of different error types (including those annotated by the LLM judges), marking error spans.
Details of  the model selection process, of the annotation protocol, sampling strategy, distributional statistics and correlation analysis methodology are provided in Appendix~\ref{app:human_eval}.

\section{Results and Discussion}
\label{sec:results}

\subsection{Automatic Evaluation Results}

Table~\ref{tab:llm_judge_all_errnum_5domains_gpt} reports the performance of all evaluated model variants under the GPT-5.1 judge.

\paragraph{At constant size (1.7B parameters), distilled models consistently outperform fine-tuning and zero-shot prompting.}
Remarkably, we also find that a compact distilled model outperforms much larger LLMs (Qwen3-32B and GPT-4.1
) on four of the five domains (Wikidata, Ice Hockey, GSM Arena, and OWID). 
More broadly, our results demonstrate that a 1.7B-parameter model can acquire strong cross-domain data-to-text generation capabilities through data-driven knowledge distillation, substantially narrowing the performance gap with significantly larger teacher models.
Crucially, these gains are achieved without any human-written or task-specific reference texts in the target domains. Instead, the distilled models rely solely on structured inputs and synthetic outputs generated by a large teacher. 

\paragraph{Overall, the 500-real condition remains clearly behind the best augmented models:} the  NormAvg (lower is better) is 0.42 when training on 500 instances vs.\ 0.27 using perturbed instances under the ZS teacher, and  0.20 vs.\ 0.10 for the mixed setting under the WebNLG-SFT teacher. 
We conjecture that augmentation is more effective for two complementary reasons. First, it is substantially more \emph{data-efficient}: structural subsampling generates many more supervision instances from a small seed set than simply adding a limited number of new real inputs. Second, structural subsampling provides simplified but still faithful versions of complex inputs, effectively decomposing difficult structured examples into easier teacher-verbalization problems. In cross-domain D2T, where long and heterogeneous inputs make faithful generation difficult even for strong teachers, these simplified demonstrations appear to be more beneficial than merely exposing the teacher to additional raw inputs with greater knowledge diversity. Overall, the results suggest that, for distillation, \emph{more numerous and structurally simplified examples} are more effective than \emph{more real but similarly complex examples}.

\paragraph{The structured data augmentation variants systematically outperform base configuration,} showing that data augmentation plays a crucial role in improving distillation quality.
Either using only perturbation or mixed subsampling/perturbation data helps improve performance on four of the five datasets.  It is less effective for tasks requiring strong aggregation and summarisation, such as weather forecasting, where teacher models already abstract away fine-grained attribute variations.


\paragraph{Differences between domains.} 
The results reveal substantial performance differences across domains, reflecting their varying levels of task complexity. Across all models, the weather domain consistently yields the worst scores, which aligns with the high complexity of the task (i.e., summarising weather data spanning 15 data points). For the remaining four domains, their rankings vary depending on the training approach, with a systematic influence of the teacher model on student performance.

Specifically, for ZS inference and DDKD with a ZS teacher, the best-performing domains are, in descending order, Wikidata, GSM Arena, Ice Hockey, and OWID, whereas, for SFT and DDKD with an SFT teacher, the ranking shifts to Ice Hockey, Wikidata, OWID, and GSM Arena. We conjecture that these differences reflect fundamental distinctions between the two teacher models.
For example, GSM Arena primarily involves entity-centric knowledge about consumer devices, much of which is likely already internalized by large language models during pretraining. In this case, additional source-domain fine-tuning may bias the teacher toward overly structured, data-driven generation, thereby reducing its effectiveness compared to a zero-shot teacher that can more directly leverage pretrained knowledge. Conversely, for domains closely aligned with WebNLG finetuning data such as Wikidata entity descriptions, D2T supervision substantially improves performance by reinforcing structured graph-to-text verbalization patterns.

\paragraph{Error types per Model.}
Table~\ref{tab:llm_judge_error_types_agg_gpt51} shows the macro-averaged error counts per output
across all target domains. Among small models (1.7B), we find that the best distilled model (DDKD from SFT) hallucinates substantially less (Not Checkable: 0.05) than both the zero-shot (0.53) and the finetuned (0.70) models. 

Compared to the best distilled  1.7B models, the large 32B zero-shot models still hallucinate more frequently 
while the 32B fine-tuned model shows comparable hallucination rate. 
However, all three large models are overall more faithful to the input (Incorrect: 0.14, 0.50, 0.41) than the best distilled model (1.13), indicating that distillation needs to be further improved to better condition on the input structure.

\paragraph{Lower error rates are not explained by conservative under-generation.}
Figure~\ref{fig:coverage_normavg} shows that across both judges, DDKD consistently achieves higher coverage than the same-size zero-shot and source-domain SFT baselines, while remaining close to the selected best teacher. Moreover, inter-judge agreement is very high at the system level ($r > 0.95$ for both score- and ratio-based judgments; $p < 0.001$ when pooling all domain-system pairs), indicating that this conclusion is robust to the choice of judge. These results show that DDKD improves faithfulness without relying on conservative omission.

\subsection{Human Evaluation Results}
For lack of space, we move the discussion of the human evaluation results to Appendix~\ref{app:human_eval_dis_summary}. These can briefly be summarised as follows.  For semantic faithfulness, Inter Annotator Agreement among the three human annotators is strong at both the instance and system levels, but it varies substantially across error categories. While Incorrect Fact and Other exhibit high agreement ($\alpha$ = 0.810 and 0.871),  with Not Checkable and Misleading showing much lower and even negative correlation than Incorrect Fact and Other reflecting a higher degree of subjectivity. 
Importantly, the agreement between human evaluation and our primary automatic judge (GPT-5.1) is high at the system level (Pearson r $> 0.90$) confirming that GPT-5.1 reliably reproduces human system rankings. 
Finally, Appendix \ref{app:human_eval_res} and Table~\ref{tab:human_eval_results} 
reveal broadly consistent trends across the two evaluation settings (automatic vs. human-based). In both cases, knowledge distillation ranks first overall, performance is lowest for all models on OpenWeather and the small distilled models outperform the large teacher model on several domains.

\section{Conclusion}
\label{sec:conclusion}

We studied data-to-text generation across multiple input structures and generation objectives. 
Overall, our findings suggest that cross-domain generalisation for data-to-text generation can be effectively transferred to compact models when large LLMs are leveraged as intermediate generators rather than as final deployment targets; and that data augmentation and data-driven knowledge distillation can help mitigate the absence of training data, enabling the study of new generation tasks based on existing real-world data.

\section{Limitations}
\label{sec:limitations}
\paragraph{Model coverage.} We evaluate DDKD on the Qwen3 and Gemma3 families, primarily focusing on small-scale models (1B--2B). Consequently, the transferability of the observed gains to significantly larger architectures is not yet fully established. We prioritize the Qwen3/Gemma3 families because they offer robust open-weight models with exceptional long-context support (>10k tokens) in the small parameter regime, whereas many other sub-2B models are constrained by shorter context windows. This focus addresses the practical bottleneck of cross-domain adaptation: fine-tuning, inference, and synthetic data generation become computationally prohibitive at long sequence lengths, making efficiency improvements on small, long-context models particularly high-value.

We only use closed-source frontier models for comparison (GPT-4.1) and in the \emph{evaluation} stage as LLM-as-a-judge annotators (GPT-5.1), because our setting is entirely reference-free and we aim to obtain the most reliable automated assessments available. Importantly, these closed-source models are not used to generate training data, provide supervision, or guide distillation, so that the full training and data-generation pipeline remains auditable and reproducible.
Extending the study to additional open-weight families and scales, and exploring how comparable results can be obtained with open evaluators, are promising directions for future work.

\paragraph{Heuristic design in structural augmentation.} While our structural augmentation (subsampling and controlled perturbations) is intended to improve model performance, its instantiation relies on heuristic choices, such as defining atomic units and enforcing exchangeability constraints. These choices necessitate minimal structural assumptions regarding the input data and may not be optimal for all conceivable domains. Nevertheless, our empirical results demonstrate that DDKD remains highly effective under generic and lightweight instantiations of these heuristics, suggesting broad applicability despite the reliance on inductive biases.

\paragraph{Cost and bias of LLM-as-a-Judge.} LLM-based judging can be domain-sensitive, with different judges exhibiting different preferences, and it can be costly—especially when using reasoning-capable models. We mitigate this by employing two strong judges (GPT-5.1 and Gemini-2.5-Pro) and by corroborating the main findings with human evaluation, which yields consistent trends. We expect that future open-model-based metrics will reduce cost and improve accessibility.

\section*{Acknowledgments}
We thank the anonymous reviewers for their feedback. This work received government funding managed by the French National Research Agency under France 2030, reference number ``ANR-23-IACL-0004'' (AI Chair Gardent: ``Semantically Consistent LLM Based Text Generation''). Experiments presented in this paper were carried out using the Grid'5000 testbed, supported by a scientific interest group hosted by Inria and including CNRS, RENATER, and several universities as well as other organizations (see \url{https://www.grid5000.fr}). This work was also granted access to the HPC resources of IDRIS under the allocation AD011016561 made by GENCI.


\bibliography{custom}

\appendix

\section{Implementation and Training Details}
\label{app:training_details}

\subsection{Training Hyperparameters}

\paragraph{Training Framework.}
All experiments are conducted using the \textsc{MS-SWIFT} framework \cite{zhao2024swiftascalablelightweightinfrastructure}.
We adopt parameter-efficient fine-tuning via LoRA \cite{hu2021loralowrankadaptationlarge} and apply early stopping based on validation loss.
Unless otherwise specified, all runs use a fixed random seed (42) and identical optimisation settings.

\paragraph{Source-Domain Fine-Tuning on WebNLG.}
For source-domain supervision, base models are finetuned on WebNLG using LoRA with rank 8 and $\alpha=32$.
Training is performed for up to 10 epochs with early stopping.
We use a learning rate of $1\times10^{-4}$, a batch size of 8, and a maximum sequence length of 13{,}000 tokens.
Using long context lengths during source-domain training ensures consistency with downstream target-domain inference, especially for domains with long inputs.

\paragraph{Target-Domain Fine-Tuning.}
For target domains, we consider a \emph{Base} setting and \emph{Augmented} settings.
In the Base setting, only 100 synthetic examples are available; models are trained for up to 500 steps with evaluation every 5 steps, using a batch size of 4 and a learning rate of $5\times10^{-5}$.
For Augmented settings (Sub, Pert, Mixed), larger synthetic datasets are used; models are trained for up to 10 epochs with evaluation every 50 steps, using a batch size of 8 and a learning rate of $1\times10^{-4}$.
Across all target-domain experiments, the maximum length is set to 13{,}000 tokens to match source-domain training and inference conditions.

\begin{table}[t]
\small
\centering
\setlength{\tabcolsep}{4pt}
\begin{tabular}{lcc}
\toprule
\textbf{Hyperparameter} & \textbf{WebNLG-SFT} & \textbf{Target-Domain SFT} \\
\midrule
Train. type        & LoRA & LoRA \\
LoRA rank          & 8    & 8 \\
LoRA $\alpha$      & 32   & 32 \\
Target mods        & all-linear & all-linear \\
LR                 & $1\!\times\!10^{-4}$ & $5\!\times\!10^{-5}$ / $1\!\times\!10^{-4}$ \\
Train BS           & 8    & 4 / 8 \\
Eval BS            & 8    & 8 / 16 \\
Grad. accum.       & 4    & 4 \\
Max len            & 13{,}000 & 8{,}192 / 13{,}000 \\
Warmup             & 0.05 & 0.05 \\
Eval freq.         & 50 steps & 5 / 50 steps \\
Early stop         & 5 evals & 5 evals \\
Val. split         & 0.15 & 0.15 \\
Seed               & 42   & 42 \\
Best metric        & Val. loss & Val. loss \\
\bottomrule
\end{tabular}
\caption{
Training hyperparameters used for source-domain (WebNLG) and target-domain fine-tuning.
For target-domain SFT, values are reported as
\emph{Base / Augmented} when they differ.
For consistency, long maximum sequence lengths are used across training stages to match downstream inference conditions.
}
\label{tab:training_hyperparams}
\end{table}

\subsection{Training and Inference Cost}

\paragraph{Computational Setup.}
All training and inference experiments are conducted on a single NVIDIA A100 GPU with 80\,GB memory.

\paragraph{Training Cost.}
Table~\ref{tab:training_cost} reports peak GPU memory usage and wall-clock training time for representative configurations.
For WebNLG-SFT, we use the Markdown linearisation format, which induces the largest memory footprint among considered input formats.
For DDKD models, costs are reported on OWID to reflect the longest target-domain inputs and thus an upper bound on resource requirements. DDKD (Best) refers to the best-performing augmentation variant per domain.
Zero-shot models incur no training cost.

\paragraph{Inference Cost.}
Inference is performed using the vLLM backend with greedy decoding (temperature $=0$).
For small distilled models (1.7B / 1B), inference over the full 100-example test set completes within 5 minutes across all domains.
Inference with large teacher models under long-context settings is substantially more expensive; generating synthetic supervision for a single target domain can take up to 12 hours.

\begin{table}[!htbp]
\centering
\small
\setlength{\tabcolsep}{3.5pt}
\renewcommand{\arraystretch}{1.05}
\begin{tabular}{@{}lcc@{}}
\toprule
\textbf{System} & \textbf{Peak Mem.} & \textbf{Train Time} \\
\midrule

\multicolumn{3}{@{}l}{\textbf{Qwen3}} \\
\hdashline
\multicolumn{3}{@{}l}{\textit{Zero-shot (ZS)}} \\
Qwen3-1.7B-ZS          & --      & --    \\
Qwen3-8B-ZS            & --      & --    \\
Qwen3-32B-ZS           & --      & --    \\
\hdashline
\multicolumn{3}{@{}l}{\textit{WebNLG-SFT}} \\
Qwen3-1.7B-SFT         & 50.0\,GB & 1.5\,h \\
Qwen3-8B-SFT           & 62.7\,GB & 10.4\,h \\
Qwen3-32B-SFT          & 78.1\,GB & 20.3\,h \\
\hdashline
\multicolumn{3}{@{}l}{\textit{DDKD (Best)}} \\
Qwen3-1.7B-DDKD-Best   & 74.8\,GB & 1.2\,h \\
\addlinespace[3pt]

\midrule

\multicolumn{3}{@{}l}{\textbf{Gemma3}} \\
\hdashline
\multicolumn{3}{@{}l}{\textit{Zero-shot (ZS)}} \\
Gemma3-1B-IT-ZS        & --      & --    \\
Gemma3-27B-IT-ZS       & --      & --    \\
\hdashline
\multicolumn{3}{@{}l}{\textit{WebNLG-SFT}} \\
Gemma3-1B-IT-SFT       & 70.3\,GB & 1.1\,h \\
Gemma3-27B-IT-SFT      & 78.1\,GB & 14.4\,h \\
\hdashline
\multicolumn{3}{@{}l}{\textit{DDKD (Best)}} \\
Gemma3-1B-IT-DDKD-Best & 78.7\,GB & 2.0\,h \\
\bottomrule
\end{tabular}

\caption{
Training memory and time cost on a single NVIDIA A100 (80\,GB).
DDKD costs are reported on OWID to reflect the maximum input-length setting.
Zero-shot models incur no training cost.
}
\label{tab:training_cost}
\end{table}

\subsection{LLM-as-a-Judge API Details}
\label{app:api_judge_details}

\paragraph{Decoding and reproducibility.}
For all API-based judgements, we use deterministic greedy decoding with temperature $=0$ and a fixed seed (42) to ensure reproducibility.
We set the maximum generation budget to $\texttt{max\_tokens}=16{,}384$ to allow the judge to provide complete rationales and category-wise annotations for long-context outputs. 

\paragraph{Robustness to incomplete generations.}
In practice, API calls may occasionally terminate early or exhibit degenerate behaviours (e.g., repetition loops) under long outputs.
To mitigate this, our evaluation pipeline implements an automatic recovery mechanism: we enforce structured JSON output and continually validate partial generations.
If a response becomes incomplete, we retain the latest syntactically valid JSON instance parsed from the stream and treat it as the final judgment for that example.
This design prevents evaluation failures from propagating and ensures that every system output receives a usable judgment.

\section{Subsampling Structural Augmentation Details}
\label{app:sub_aug_detail}

We describe the domain-specific subsampling strategies used to construct structurally subsampled inputs for teacher verbalization. 
Across domains, subsampling operates over \emph{atomic units} that preserve internal structure (e.g., top-level JSON blocks, attribute--value pairs, or table rows), while keeping domain identifiers and high-level metadata unchanged. 
Depending on the input granularity and combinatorial complexity, we either enumerate all non-empty subsets or sample a fixed number of subsets at multiple sparsity levels.

\paragraph{Wikidata domain.}
Given an entity $e$ with a set of attribute--value pairs $\{(A_i,V_i)\}$, we keep the entity identifier $e$ and enumerate all non-empty subsets of the attribute--value pairs as distinct inputs.

\paragraph{Ice Hockey domain.}
Each instance is a JSON dictionary with exactly $n=8$ top-level key--value blocks, where each block may contain nested structure.
We treat each top-level block as an atomic unit and enumerate all non-empty subsets of these units, yielding $2^8-1=255$ variants per instance (including the full input), consistent with Table~\ref{tab:augmentation_stats}.

\paragraph{OpenWeather domain.}
For long and dense inputs, enumerating all subsets is impractical. 
We therefore apply a constrained subsampling scheme: we restrict structural subsampling to core weather-related fields and perform rule-based temporal thinning over the time series.
Concretely, instead of using all 3-hourly records, we select a fixed set of representative time points per day (e.g., 06:00, 12:00, 18:00), which reduces input length while maintaining coverage across days.

\paragraph{GSM Arena domain.}
Each instance is a JSON record with meta fields (e.g., \texttt{id}, \texttt{name}, and \texttt{details.name}/\texttt{details.img}) and a set of specification blocks under \texttt{details}. 
We subsample at the level of specification \emph{blocks}: (i) each non-empty \texttt{details.detailSpec} category block (e.g., \texttt{Network}, \texttt{Display}), whose within-category key--value specifications are kept intact; and (ii) the entire \texttt{details.quickSpec} list when present, treated as a single block.
For a device record with $M$ available blocks, we generate $M$ subsampled variants by sampling one subset for each subset size $k \in \{1,\ldots,M\}$ uniformly without replacement (with $k=M$ corresponding to the full set). 
Each variant keeps all meta fields and includes only the selected blocks before teacher verbalization.

\paragraph{OWID domain.}
OWID instances contain a markdown code block with metadata lines (prefixed by ``\# ''), followed by a \texttt{date,value} header and a sequence of time-series rows.
We construct variants by subsampling the \texttt{date,value} rows while keeping the metadata and header unchanged. 
Specifically, for each instance we create 10 variants with retention ratios $r \in \{0.1,0.2,\ldots,1.0\}$. 
For $r<1$, given $N$ rows we set $k=\max(1,\mathrm{round}(rN))$ and, when $N>1$, cap $k$ at $N-1$ to avoid duplicating the full table; we then sample $k$ rows uniformly without replacement using a deterministic seed (dependent on the instance index and ratio), and restore chronological order by sorting sampled indices. 
For $r=1.0$, we keep all $N$ rows.

\section{Noise-based Structural Perturbation Strategies}
\label{app:noise_pert_detail}

We consider two complementary types of perturbations depending on the data format and domain characteristics:
\emph{in-instance perturbation} and \emph{cross-instance perturbation}.

For CSV-based tabular data (e.g., OWID), where inputs primarily consist of numeric values and temporal records, we apply in-instance perturbation by randomly modifying values within instances (e.g., modifying the birth rate for a given year).
This design avoids introducing excessive noise while preserving global structural coherence, which is crucial for aggregation-sensitive tasks such as chart captioning.

For the hierarchical JSON-formatted data, we perform category- or block-level cross-instance perturbation.
Concretely, we identify semantically aligned structural blocks across different instances (e.g., weather records corresponding to the same day and time slot) and exchange their values across instances.
This strategy introduces plausible structural variation while maintaining schema consistency.

For Markdown-based inputs, which are inherently more descriptive and easier for language models to interpret, we directly perform cross-instance swapping of selected property-value pairs across instances.
This enables the model to observe diverse attribute configurations while remaining within a valid structural space.

\begin{table}[!htbp]
\small
\centering
\setlength{\tabcolsep}{4pt}
\begin{tabular}{lcc}
\toprule
\textbf{Domain} & \textbf{In-instance} & \textbf{Cross-instance} \\
\midrule
Wikidata     &  & \checkmark \\
Ice Hockey   &  & \checkmark \\
OpenWeather  &  & \checkmark \\
GSM Arena    &  & \checkmark \\
OWID         & \checkmark &  \\
\bottomrule
\end{tabular}
\caption{
Structural perturbation strategies adopted for each target domain.
}
\label{tab:perturbation_types}
\end{table}

\section{QUINTD-5 Construction}
\label{app:quintd-5}

\begin{table*}[t]
\centering
\small
\setlength{\tabcolsep}{4pt}
\renewcommand{\arraystretch}{1.05}
\begin{tabular}{lcccccc}
\toprule
\multirow{2}{*}{\textbf{Domain}} &
\multicolumn{3}{c}{\textbf{QUINTD-1}} &
\multicolumn{3}{c}{\textbf{QUINTD-5}} \\
\cmidrule(lr){2-4} \cmidrule(lr){5-7}
&
\#Props / Rows &
\#Vals / Cols &
Input Len &
\#Props / Rows &
\#Vals / Cols &
Input Len \\
\midrule

\textbf{Wikidata}
& 95
& 416
& 122 (83–196)
& 176
& 1,762
& 123 (82–206)

\\

\textbf{Ice Hockey}
& 20
& 971
& 333 (314–373)
& 20
& 3,047
& 332 (314–382)

\\

\textbf{OpenWeather}
& 28
& 4,833
& 6,415 (6,214–6,749)
& 28
& 7,986
& 6,401 (6,159–6,886)

\\

\textbf{GSM Arena}
& 9
& 2,432
& 1,307 (811–3,164)
& 9
& 8,140
& 1,273 (773–3,164)

\\

\textbf{OWID}
& 236
& 2
& 4,004 (648–8,111)
& 273
& 2
& 4,465 (135–8,088)

\\

\bottomrule
\end{tabular}

\caption{Comparison of \textbf{QUINTD-1} and \textbf{QUINTD-5} development inputs.
\#Props / Rows denotes the number of distinct properties (JSON/Markdown)
or rows (CSV), and \#Vals denotes the number of distinct values.
Input Len shows average input tokens (min–max, Qwen3 tokenizer). \textbf{QUINTD-5} preserves similar input length distributions while substantially increasing knowledge diversity.}
\label{tab:quintd1_vs_quintd5}
\end{table*}

\subsection{Overview}

To complement our structural augmentation study, we construct an extended version of the QUINTD-1 benchmark \citep{kasner-dusek-2024-beyond}, denoted \textbf{QUINTD-5}, containing five times more development instances per domain.
This provides a controlled setting for comparing two strategies for improving cross-domain data-to-text generation without gold supervision:
(i) scaling real structured inputs and
(ii) structure-preserving augmentation.

QUINTD-5 is built using the official QUINTD-1 data collection pipelines, querying the same public APIs with identical preprocessing.
Each domain contains 500 instances, compared to 100 in QUINTD-1, ensuring identical schemas, distributions, and task definitions, differing only in data volume.


\subsection{Data Analysis}

We compare the statistical properties of QUINTD-5 and QUINTD-1 development data.

Table~\ref{tab:quintd1_vs_quintd5} shows that average input lengths remain highly stable despite the fivefold increase in data size (e.g., OpenWeather: 6,415 vs. 6,401 tokens; Wikidata: 122 vs. 123).
This confirms that QUINTD-5 preserves the original input length scale, data structure, and task formulation. At the same time, QUINTD-5 substantially increases \textbf{knowledge diversity}, reflected by the large growth in the number of distinct values and entities (e.g., Wikidata: 416 → 1,762; GSM Arena: 2,432 → 8,140). This increase reflects broader coverage of the underlying knowledge space rather than structural changes.
In a Markdown-based domain such as Wikidata, the input schema remains fixed, while additional instances introduce new entities and predicates.\footnote{Here, ``\#Props'' counts the number of distinct predicates observed in the dataset rather than the schema itself.}
Similarly, for CSV-based domains such as OWID, each instance contains the same metadata fields, while additional instances correspond to new measurements and numerical values rather than new structural fields.


\subsection{Instance-level Distribution Comparison}

\begin{figure*}
    \centering
    \includegraphics[width=0.95\linewidth]{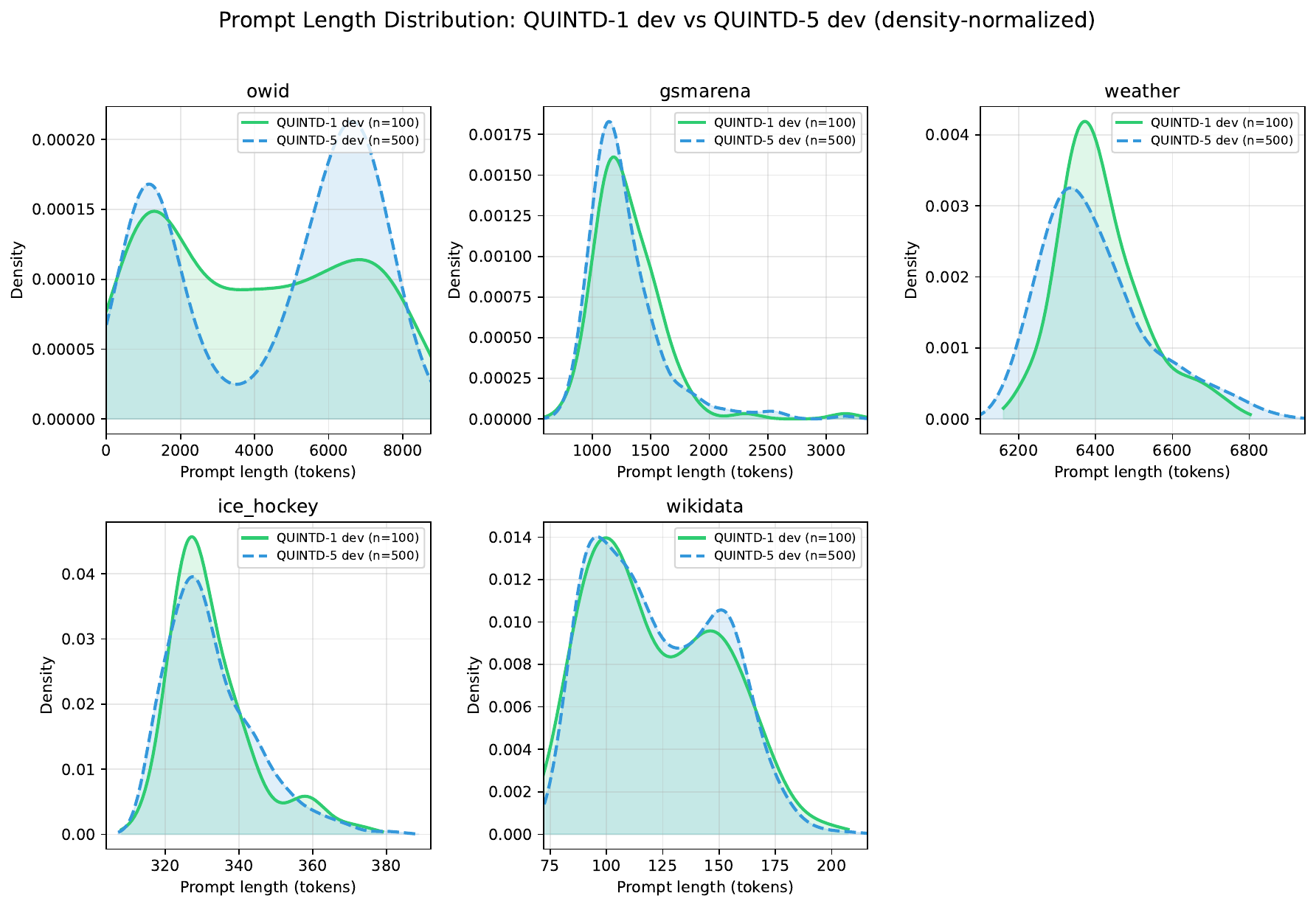}
    \caption{Kernel density estimates (KDE) of prompt token length distributions for \textbf{QUINTD-1} and \textbf{QUINTD-5} dev sets across five domains. Density normalization enables direct comparison despite differing sample sizes (100 vs. 500).}
    \label{fig:quintd_distribution}
\end{figure*}

Figure~\ref{fig:quintd_distribution} further examines instance-level length distributions. Across Wikidata, Ice Hockey, OpenWeather, and GSM Arena, QUINTD-1 and QUINTD-5 exhibit highly similar distributions with closely aligned modes and spread.
This confirms that increasing real data volume does not significantly alter the structural or length characteristics of the inputs.

OWID shows a more pronounced bimodal pattern in QUINTD-5, indicating the inclusion of additional extreme cases that were underrepresented in QUINTD-1.
This suggests that scaling real-world data increases coverage of the underlying data space and improves knowledge diversity, while potentially introducing greater distributional heterogeneity.

Overall, \textbf{QUINTD-5 increases knowledge diversity while preserving input structure, task formulation, and length distributions.}
This makes it a controlled benchmark for isolating the effect of scaling real data volume compared to structure-preserving augmentation.

\section{LLM-as-a-Judge Prompt and Guidelines}
\label{app:llm_judge_prompt}

\subsection{System Prompt}
\label{subsec:prompts}

\begin{lstlisting}
You are an expert data-to-text error annotation system. You understand structured data and you can correctly operate with units and numerical values. You are designed to output span-level annotations in JSON.
\end{lstlisting}

\subsection{User Prompt Template}
\label{subsec:user-prompt}

\begin{lstlisting}
Given the data:
```
{data}
```
Annotate all the errors in the following text:
```
{text}
```
Output the errors as a JSON list "errors" in which each object contains fields  "reason", "text", and "type". The value of "text" is the text of the error. The value of "reason" is the reason for the error. The value of "type" is one of {{0, 1, 2, 3}} based on the following list:
- 0: Incorrect fact: The fact in the text contradicts the data.
- 1: Not checkable: The fact in the text cannot be checked in the data.
- 2: Misleading: The fact in the text is misleading in the given context.
- 3: Other: The text is problematic for another reason, e.g. grammatically or stylistically incorrect, irrelevant, or repetitive.

The list should be sorted by the position of the error in the text.

*Example:*
data:
```
[ [ "Aditi Bhagwat", "occupation", "television actor" ], [ "Aditi Bhagwat", "date of birth", "18 January 1981" ] ]     
```
text:
```
Aditi Bhagwat, born on January 18, 1991, used to be a popular Indian television actor. The data comes from a knowledge graph.
```
output:
```
{{ "errors": [{{"reason": "The data mentions that the actor was born on 1981", "text": "1991", "type": 0}}, {{"reason": "Misleadingly suggests that the actor is not alive", "text": "used to be", type: 2}}, {{"reason": "Popularity is not mentioned in the data", "text": "popular", type: 1}}, {{"reason": "Nationality is not mentioned in the data", "text": "Indian", type: 1}}, {{"reason": "The note is superfluous", "text": "The data comes from a knowledge graph.", type: 3}}] }}
```

Notes:
- Some details may not be mentioned in the text: do not count omissions as errors.
- Do not be too strict: some facts can be less specific than in the data (rounded values, shortened or abbreviated text, etc.), and these should not be counted as errors.
- If there are no errors in the text, "errors" must be an empty list.
- Each reason should be concise (a short phrase) and describe the exact error.
- For repeated or stylistically poor information, annotate it as a single error. The "text" field must contain only a minimal representative span of the error (no more than 10 words), not the entire repeated or problematic passage.
- Minor differences in spelling, transliteration, or the use/omission of accents and diacritics in names are acceptable as long as it is clear that they refer to the same entity and should not be counted as errors.
\end{lstlisting}

\section{Cost of LLM-as-a-Judge Evaluation}
\label{app:judge_cost}

We conduct reference-free evaluation using two independent LLM judges (GPT-5.1 and Gemini-2.5-Pro) accessed via OpenRouter. OpenRouter bills requests token-wise with separate rates for \emph{input} (prompt) tokens and \emph{output} (completion) tokens, quoted in USD \emph{per million tokens}. 

Overall, our evaluation is cost-intensive primarily because each judgment prompt includes (i) long structured inputs (often spanning many thousands of tokens), (ii) the candidate generations to be assessed, and (iii) a detailed error-taxonomy rubric. Using two judges for robustness further doubles the number of judgment calls. Across all systems, domains, and test instances, the total evaluation cost amounts to approximately \$500.

\section{Error Types}
\label{app:error-types}

This section presents the four error types used in the LLM-as-a-Judge evaluation protocol described in Section~\ref{subsec:evaln} and provides illustrative examples for each category.
All error types are applied at the span level, and multiple error types may co-occur within a single generated output.

\paragraph{Incorrect Fact.}
A text span is annotated as \emph{Incorrect Fact} if it directly contradicts the input structured data.
This includes incorrect attribute values, relations, numerical quantities, or temporal information.
For example, stating a date of birth that differs from the value provided in the input data constitutes an incorrect fact.

\paragraph{Not Checkable.}
A text span is annotated as \emph{Not Checkable} if it cannot be verified based on the input data.
This category captures hallucinated or unsupported statements that are neither confirmed nor contradicted by the structured input.
For instance, asserting a person’s nationality or popularity when such information is absent from the data is considered not checkable.

\paragraph{Misleading.}
A text span is annotated as \emph{Misleading} if it is likely to induce an incorrect or biased interpretation of the input data, even if it is not strictly false.
This includes phrasing or implications that distort the intended meaning of the data.
For example, expressions such as “used to be” may misleadingly suggest a change of status not supported by the input.

\paragraph{Other.}
The \emph{Other} category captures problematic spans that do not fall into the previous categories, including grammatical or stylistic issues, irrelevant content, or unnecessary meta-statements.
For example, superfluous remarks about the data source or repetitive content are annotated under this category.

\paragraph{Annotation Boundary Conditions.}
Omissions of information present in the input data are not annotated as errors.
Annotators do not penalize acceptable paraphrases, minor differences in specificity, or variations in spelling or transliteration that clearly refer to the same entity.
Repeated or stylistically poor content is annotated as a single error using a minimal representative span.

\section{Additional Analysis of LLM Judge Agreement}
\label{app:llm_judge_agree}

\subsection{Token-level Agreement}
\label{subsec:token-agr}
For each generated output, each judge produces a binary vector
\[
\mathbf{w}^{(j)} \in \{0,1\}^{T},
\]
marking whether each token participates in an error of the given type.  
We concatenate these vectors across all outputs from all systems (14 models), producing long sequences 
$\mathbf{w}^{(A)}$ and $\mathbf{w}^{(B)}$ for two judges $A$ and $B$.  
The token-level agreement is defined as:
\[
r_{\text{token}} = \mathrm{corr}\bigl(\mathbf{w}^{(A)},\, \mathbf{w}^{(B)}\bigr).
\]

\subsection{Example-level and System-level Definitions}
\label{subsec:agr-def}
\medskip
\noindent\textbf{Example-level.}
For each example $i$, each judge $j$ outputs the number of errors of the given category:
\[
e^{(j)}_i = \text{error count for the category in example } i.
\]
Concatenating the values across all examples from all systems yields vectors 
$\mathbf{e}^{(A)}$ and $\mathbf{e}^{(B)}$.  
The example-level agreement is:
\[
r_{\text{example}} = \mathrm{corr}\bigl(\mathbf{e}^{(A)},\, \mathbf{e}^{(B)}\bigr).
\]

\medskip
\noindent\textbf{System-level.} 
For each system $s$, and judge $j$, we compute the average number of errors:
\[
m^{(j)}_s = \frac{1}{N_s} \sum_{i \in s} e^{(j)}_i,
\]
where $N_s$ is the number of outputs generated by system $s$.  
This results in two vectors $\mathbf{m}^{(A)}$ and $\mathbf{m}^{(B)}$ (each of length equal to the number of evaluated systems), and the system-level agreement is:
\[
r_{\text{system}} = \mathrm{corr}\bigl(\mathbf{m}^{(A)},\, \mathbf{m}^{(B)}\bigr).
\]

\medskip
\noindent
This multi-granularity formulation enables us to evaluate agreement at fine (token), medium (example), and coarse (system) levels in the same domain/task.

\subsection{LLM Judge Agreement Results}
Table~\ref{tab:llm_judge_agreement_cross_domain} reports agreement aggregated across all domains,
while Table~\ref{tab:llm_judge_agreement_full} presents domain-specific results.

We observe that the two judges exhibit medium agreement at the example level (Pearson’s r=0.666) and very high agreement at the system level (Pearson’s r=0.955) across all domains. These results support the robustness of LLM-based evaluation for comparing data-to-text systems at the system level and for assessing individual-generated outputs at the example level.

\begin{table*}[!htbp]
\centering
\setlength{\tabcolsep}{5pt}
\begin{tabular}{lccccc}
\toprule
\textbf{Level} &
\textbf{Incorrect Fact} &
\textbf{Not Checkable} &
\textbf{Misleading} &
\textbf{Other} &
\textbf{All (Total Errors)} \\
\midrule
Token    & 0.618 & 0.490 & 0.169 & 0.580 & 0.582 \\
Example  & 0.680 & 0.256 & 0.334 & 0.459 & 0.666 \\
System   & \textbf{0.941} & \textbf{0.831} & \textbf{0.781} & \textbf{0.949} & \textbf{0.955} \\
\bottomrule
\end{tabular}
\caption{
Cross-domain inter-judge agreement between GPT-5.1 and Gemini-2.5-Pro,
macro-averaged across all target domains.
\textbf{All (Total Errors)} is computed by correlating summed error counts across categories.
}
\label{tab:llm_judge_agreement_cross_domain}
\end{table*}

\begin{table*}[!htbp]
\centering
\small
\setlength{\tabcolsep}{4pt}
\begin{tabular}{llccccc}
\toprule
\textbf{Domain} & \textbf{Level} 
& \textbf{Incorrect Fact} 
& \textbf{Not Checkable} 
& \textbf{Misleading} 
& \textbf{Other} 
& \textbf{All (Total Errors)} \\
\midrule

\multirow{3}{*}{Wikidata}
& Token    & 0.563 & 0.663 & 0.147 & 0.255 & 0.523 \\
& Example & 0.521 & 0.748 & 0.088 & 0.280 & 0.562 \\
& System  & 0.591 & \textbf{0.930} & 0.716 & 0.395 & \textbf{0.932} \\

\midrule
\multirow{3}{*}{OpenWeather}
& Token    & 0.566 & 0.183 & 0.173 & 0.588 & 0.531 \\
& Example & 0.557 & 0.172 & 0.291 & 0.361 & 0.550 \\
& System  & \textbf{0.915} & 0.444 & 0.075 & 0.826 & \textbf{0.913} \\

\midrule
\multirow{3}{*}{OWID}
& Token    & 0.532 & 0.359 & 0.128 & 0.559 & 0.622 \\
& Example & 0.738 & 0.175 & 0.313 & 0.535 & 0.704 \\
& System  & \textbf{0.942} & 0.451 & \textbf{0.955} & 0.917 & \textbf{0.957} \\

\midrule
\multirow{3}{*}{Ice Hockey}
& Token    & 0.507 & 0.403 & 0.161 & 0.521 & 0.482 \\
& Example & 0.877 & 0.518 & 0.089 & 0.262 & 0.719 \\
& System  & \textbf{0.996} & \textbf{0.960} & 0.780 & 0.652 & \textbf{0.919} \\

\midrule
\multirow{3}{*}{GSM Arena}
& Token    & 0.716 & 0.595 & 0.196 & 0.535 & 0.572 \\
& Example & 0.814 & 0.665 & 0.228 & 0.579 & 0.758 \\
& System  & \textbf{0.964} & 0.726 & 0.401 & \textbf{0.954} & \textbf{0.953} \\

\bottomrule
\end{tabular}
\caption{
Inter-judge Pearson correlation ($r$) between GPT-5.1 and Gemini-2.5-Pro at different aggregation levels,
computed separately for each domain and error category.
\textbf{All (Total Errors)} is computed by correlating the summed error counts across all error categories.
System-level agreement is consistently higher than Token- and Example-level agreement,
indicating robust consistency in system ranking across judges.
}
\label{tab:llm_judge_agreement_full}
\end{table*}

\begin{table*}[!htbp]
\centering
\begin{tabular}{llll}
\toprule
\textbf{Target Domain} &
\textbf{Selected DDKD Model} &
\textbf{Teacher} \\
\midrule

Wikidata &
Qwen3-1.7B-DDKD-Mixed &
Qwen3-32B-SFT  \\

Ice Hockey &
Qwen3-1.7B-DDKD-Mixed &
Qwen3-32B-SFT  \\

OpenWeather &
Qwen3-1.7B-DDKD-Sub &
Qwen3-32B-SFT  \\

GSM Arena &
Qwen3-1.7B-DDKD-Mixed &
Qwen3-32B-ZS  \\

OWID &
Qwen3-1.7B-DDKD-Mixed &
Qwen3-32B-SFT  \\

\bottomrule
\end{tabular}
\caption{
\textbf{Final selection of distilled models per target domain.}
Model selection is primarily based on the average number of errors per output under
GPT-5.1, and further informed by error rate and consistency with an independent
judge (Gemini-2.5-Pro).
Teacher models are selected on a per-domain basis according to their domain-level
performance.
}
\label{tab:final_ddkd_selection}
\end{table*}

\section{Human Evaluation Details}
\label{app:human_eval}

This section provides detailed information on the construction of the human evaluation set, the annotation protocol, and additional statistics that complement the main experimental results, which we discuss in Section~\ref{app:human_eval_res} through Section~\ref{app:llm_human_eval_agree}.

\subsection{Evaluation Set Construction}
\label{subsec:dev-set}
To validate the reliability of LLM-based evaluation, we construct a diagnostic human evaluation set covering diverse error phenomena across domains.
The evaluation set is built at the \emph{input level}, ensuring that all compared systems are evaluated on the same inputs.

\paragraph{Sampling Overview.}
We select 60 structured inputs in total, with 12 inputs per target domain in QUINTD-1.
For each selected input, we collect outputs from four representative systems:
(i) a small zero-shot model,
(ii) a small source-domain SFT model,
(iii) the best-performing large teacher model, and
(iv) the best distilled model.
This results in 240 outputs for human annotation.

\subsection{Stratified Sampling with Dual Anchors}
\label{subsec:sampling}
The human evaluation set is constructed via stratified sampling over error categories using LLM-judge annotations as anchors.
For each domain, we aim to select 12 inputs, corresponding to three instances per error category:
\emph{Incorrect Fact}, \emph{Not Checkable}, \emph{Misleading}, and \emph{Other}.

\paragraph{Primary Anchor.}
We use the outputs of the best-performing large teacher model as the primary anchor for sampling.
For each domain and error category, we prioritize inputs whose teacher outputs:
(i) contain only the target error category,
(ii) exhibit a small number of errors (preferably 1--3 instances),
and (iii) otherwise contain the target error category under relaxed constraints.

\paragraph{Secondary Anchor.}
In cases where a given error category is insufficiently represented in the teacher outputs, we employ a secondary anchor based on the zero-shot small model.
This fallback strategy allows us to supplement rare error categories while avoiding systematic bias toward the weakest system.
The final selection ensures that each domain contains coverage of all error categories whenever possible.

All sampling decisions are performed at the input level.
Once the final set of inputs is determined, outputs from all four systems are collected on the same inputs to enable fair comparison.

\subsection{Selection of the  best Final Distilled Models.}
\label{app:model-selection}

To select the final distilled models used in downstream analysis, we adopt a
principled, multi-criteria strategy.
Model selection is primarily guided by the average number of errors per output
under GPT-5.1, and further informed by
(i) the error rate, defined as the proportion of outputs containing at least one error,
and (ii) consistency with evaluations from an independent judge, Gemini-2.5-Pro.

For most target domains, this procedure yields a clear and consistent choice.
On Wikidata, Ice Hockey, and OWID, \textsc{Qwen3-1.7B-DDKD-Mixed} distilled from the
WebNLG-SFT teacher achieves the lowest error counts under GPT-5.1 while remaining
competitive under Gemini-2.5-Pro, and is therefore selected as the final model.
Similarly, for OpenWeather, \textsc{Qwen3-1.7B-DDKD-Sub} distilled from the WebNLG-SFT
teacher exhibits the most stable performance, achieving the best GPT-5.1 error
counts and strong agreement across judges.

GSM Arena exhibits a different behaviour.
As shown in Table~\ref{tab:llm_judge_all_errnum_5domains_gpt}, the strongest teacher
on this domain is \textsc{Qwen3-32B-ZS}, which outperforms the corresponding
WebNLG-SFT model. We discuss this difference between domains in Section~\ref{sec:results}.
Accordingly, the selected DDKD model for GSM Arena is distilled from the ZS teacher.
While \textsc{Qwen3-1.7B-DDKD-Pert} achieves the lowest average error count under GPT-5.1,
\textsc{Qwen3-1.7B-DDKD-Mixed} distilled from the ZS teacher achieves the lowest error
rate and exhibits more consistent behaviour across both GPT-5.1 and Gemini-2.5-Pro.
Given the small absolute differences in error counts and the stronger cross-judge
consistency, we select \textsc{Qwen3-1.7B-DDKD-Mixed} distilled from the ZS teacher
as the final model for GSM Arena.

We summarise our final distilled model selections in Table~\ref{tab:final_ddkd_selection}.




\begin{table*}[t]
\centering
\small
\setlength{\tabcolsep}{6pt} 
\begin{tabular}{l c c c c c} 
\toprule
\multirow{2}{*}{\textbf{Granularity}} & \textbf{Overall} & \multicolumn{4}{c}{\textbf{Breakdown by Error Type ($\alpha$)}} \\

\cmidrule(lr){3-6} 
 & \textbf{Agreement} & \textbf{Incorrect Fact} & \textbf{Not Checkable} & \textbf{Misleading} & \textbf{Other} \\
\midrule
\multicolumn{6}{l}{\textit{Metric: Nominal $\alpha$ (Span-level detection)}} \\
Token-level & 0.369 & 0.370 & 0.399 & 0.020 & 0.342 \\
\addlinespace 

\midrule
\multicolumn{6}{l}{\textit{Metric: Interval $\alpha$ (Error counts)}} \\
Instance-level & 0.578 & 0.549 & 0.530 & 0.033 & 0.710 \\
\textbf{System-level} & \textbf{0.902} & \textbf{0.810} & 0.498 & -0.072 & \textbf{0.871} \\
\addlinespace

\midrule
\multicolumn{6}{l}{\textit{Metric: Nominal $\alpha$ (Summary Level)}} \\
summarisation & 0.693 & \multicolumn{4}{c}{\textit{N/A (Categorical: Short / Medium / No Aggregation)}} \\
\bottomrule
\end{tabular}
\caption{\textbf{Inter-Annotator Agreement (Krippendorff's $\alpha$) Analysis.} 
We report the agreement scores for the overall dataset and a breakdown by specific error types. 
The \textbf{Overall} column represents the aggregate agreement calculated over all error types. 
The breakdown columns show the agreement when isolating each error type (binary distinction for token-level, counts for instance/system-level).
Note that \textit{Misleading} errors show near-zero or negative agreement, indicating high subjectivity, whereas \textit{Incorrect Fact} and \textit{Other} show high reliability at the system level.}
\label{tab:iaa_detailed}
\end{table*}

\subsection{Human Annotation Protocol}
\label{subsec:human-eval-protocol}

\paragraph{Annotators.}
Because of the complexity of the annotation task, we did not use crowdsourcing. Instead, the annotations were carried out on a voluntary basis by three researchers involved in this work. All of them have 
substantial research experience in data-to-text generation and related natural language generation tasks, are fluent English speakers, and routinely work with English-language scientific and technical texts.
Their expertise enables reliable identification of factual errors, unverifiable statements, and misleading content in structured-data-to-text outputs.

Human annotators evaluate each generated output independently.
For each output, annotators are instructed to identify all error spans present in the text and assign each span to one of the predefined error categories:
\emph{Incorrect Fact}, \emph{Not Checkable}, \emph{Misleading}, or \emph{Other}.
Multiple error spans and multiple error categories may be assigned to the same output.

For each output, the final annotation consists of:
(i) the set of error spans,
(ii) the corresponding error categories for each span, and
(iii) the resulting number of errors per category.
These annotations support analyses at multiple aggregation levels, including example-level, system-level, and domain-level comparisons, as well as comparisons with LLM-based judgments.

\subsection{Error Annotation Coverage and Distribution}
\label{subsec:error-dist}
To assess the coverage and balance of the constructed evaluation set, we analyse the distribution of error types across domains and LLM judges.
For each domain, we aggregate error annotations over all four systems and report both the number of error spans identified by the judge and the number of examples containing at least one error of a given type.

\noindent Table~\ref{tab:human_error_distribution} summarises the error distributions for OWID, GSM Arena, OpenWeather, Wikidata, and Ice Hockey under GPT-5.1 and Gemini-2.5-Pro.

\begin{table*}[t]
\centering
\small
\setlength{\tabcolsep}{4pt}
\begin{tabular}{llcccc}
\toprule
\textbf{Domain} & \textbf{Error Type} 
& \multicolumn{2}{c}{\textbf{GPT-5.1}} 
& \multicolumn{2}{c}{\textbf{Gemini-2.5-Pro}} \\
\cmidrule(lr){3-4} \cmidrule(lr){5-6}
 &  & \textbf{Span \#} & \textbf{Example \#} 
    & \textbf{Span \#} & \textbf{Example \#} \\
\midrule

\multirow{4}{*}{OWID}
& Incorrect Fact   & 62 & 23 & 69 & 28 \\
& Not Checkable    & 13 & 8  & 2  & 2  \\
& Misleading       & 10 & 7  & 9  & 9  \\
& Other            & 12 & 11 & 28 & 25 \\
\midrule

\multirow{4}{*}{GSM Arena}
& Incorrect Fact   & 23 & 16 & 31 & 20 \\
& Not Checkable    & 19 & 16 & 8  & 8  \\
& Misleading       & 8  & 8  & 9  & 9  \\
& Other            & 23 & 12 & 17 & 11 \\
\midrule

\multirow{4}{*}{OpenWeather}
& Incorrect Fact   & 167 & 35 & 218 & 40 \\
& Not Checkable    & 20  & 10 & 1   & 1  \\
& Misleading       & 7   & 7  & 57  & 17 \\
& Other            & 10  & 9  & 10  & 10 \\
\midrule

\multirow{4}{*}{Wikidata}
& Incorrect Fact   & 10 & 10 & 22 & 20 \\
& Not Checkable    & 20 & 17 & 9  & 9  \\
& Misleading       & 6  & 6  & 12 & 10 \\
& Other            & 4  & 4  & 8  & 8  \\
\midrule

\multirow{4}{*}{Ice Hockey}
& Incorrect Fact   & 22 & 9  & 31 & 15 \\
& Not Checkable    & 8  & 6  & 5  & 5  \\
& Misleading       & 4  & 4  & 8  & 7  \\
& Other            & 4  & 4  & 12 & 9  \\
\bottomrule
\end{tabular}
\caption{
Distribution of error annotations in the human evaluation set, aggregated over four systems.
For each target domain and error category, we report the total number of error spans (\emph{Span \#}) and the number of examples containing at least one such error (\emph{Example \#}), as identified by GPT-5.1 and Gemini-2.5-Pro.
All domains contain 48 annotated outputs (12 inputs $\times$ 4 systems).
}
\label{tab:human_error_distribution}
\end{table*}

\subsection{Human Annotation Toolkit Interface}
\label{subsec:annot-toolkit}
To facilitate consistent span-level error annotation, we developed a lightweight human annotation toolkit implemented in \textsc{Streamlit}.
The toolkit is designed to support structured, fine-grained inspection of data-to-text outputs while minimizing annotator overhead.

\begin{figure*}
    \centering
    \includegraphics[width=0.75\linewidth]{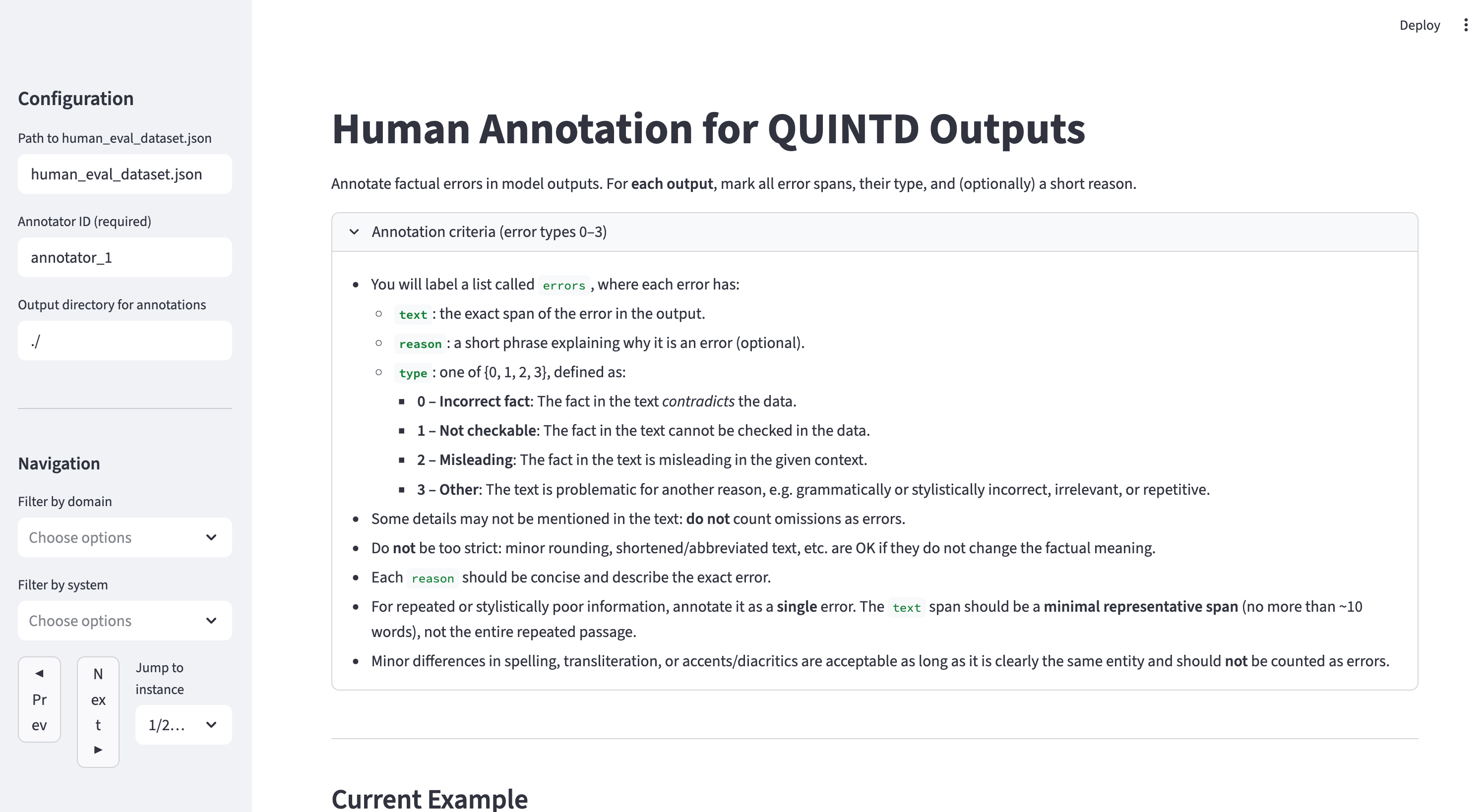}
    \caption{Human Annotation Toolkit Interface - Screenshot 1}
    \label{fig:tool_screen_1}
\end{figure*}

As shown in Fig.~\ref{fig:tool_screen_1}, the interface first presents annotators with a concise set of annotation guidelines, including the definitions of the four error categories and practical instructions for span selection.

\begin{figure*}
    \centering
    \includegraphics[width=0.75\linewidth]{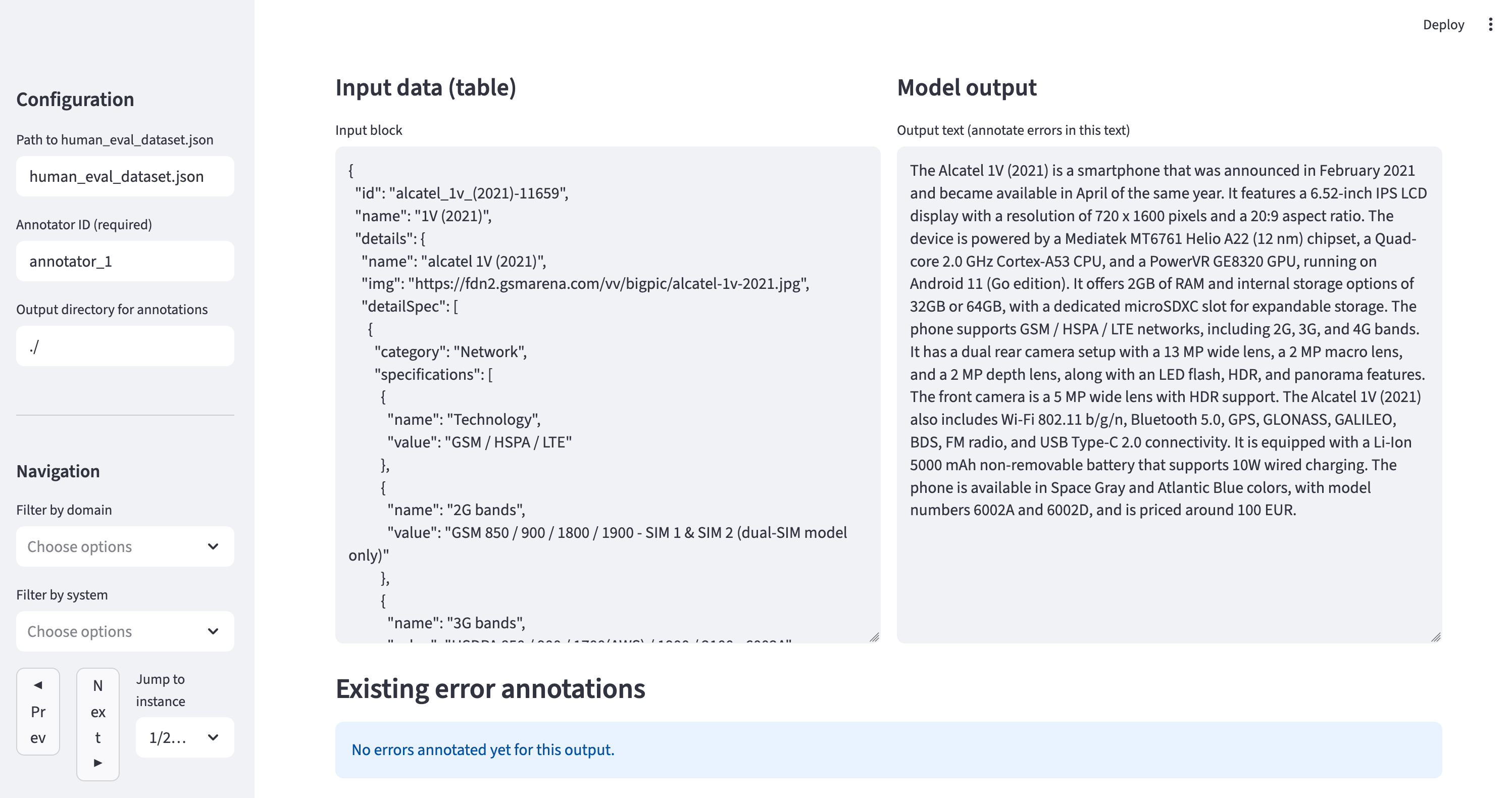}
    \caption{Human Annotation Toolkit Interface - Screenshot 2}
    \label{fig:tool_screen_2}
\end{figure*}

Fig.~\ref{fig:tool_screen_2} illustrates the main annotation view.
For each instance, the structured input (rendered in its original format) is displayed alongside the corresponding model output.
Annotators inspect the output text and identify error spans by copying the minimal representative text span, following the guidelines described in Appendix~\ref{app:human_eval}.

\begin{figure*}
    \centering
    \includegraphics[width=0.75\linewidth]{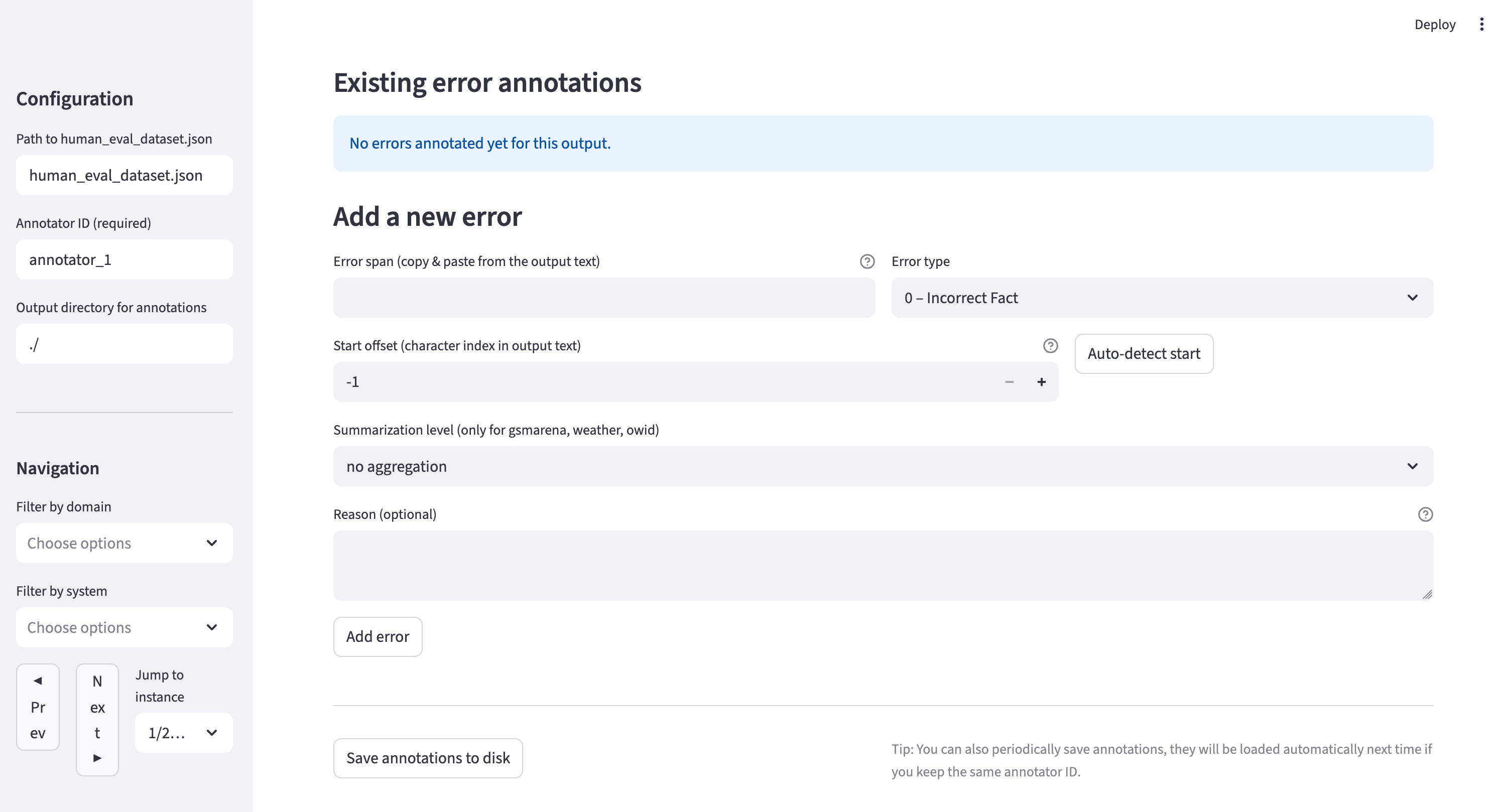}
    \caption{Human Annotation Toolkit Interface - Screenshot 3}
    \label{fig:tool_screen_3}
\end{figure*}

The error annotation form is shown in Fig.~\ref{fig:tool_screen_3}.
For each identified span, annotators assign one of the predefined error categories (\emph{Incorrect Fact}, \emph{Not Checkable}, \emph{Misleading}, or \emph{Other}), optionally provide a short explanation, and specify the span offset to ensure reproducibility.

\subsection{Human Evaluation Results}
\label{app:human_eval_res}

Table~\ref{tab:human_eval_results} reports the average number of annotated errors per output for each system, aggregated over three human annotators.
Lower values indicate fewer identified factual or verifiability-related issues.

Across all evaluated domains, DDKD achieves the lowest error counts among the 1.7B models and the lowest normalized average (NormAvg) overall.
Domain-level variations are observed across systems, reflecting differences in domain complexity and error prevalence.

We discuss the results in Section~\ref{sec:results}.

\begin{table*}[t]
\centering
\small
\setlength{\tabcolsep}{5pt}
\renewcommand{\arraystretch}{1.05}
\begin{tabular}{lccccc c}
\toprule
\textbf{System} &
\textbf{GSM-Arena} &
\textbf{Ice Hockey} &
\textbf{OWID} &
\textbf{Weather} &
\textbf{Wikidata} &
\textbf{NormAvg}$\downarrow$ \\
\midrule
\textit{Large Model (32B)} \\
Teacher Model        & 1.06 & 0.42 & 1.25 & 0.78 & \textbf{0.44} & 0.15 \\
\midrule
\textit{Small Models (1.7B)} \\
Zero-Shot                & 1.47 & 1.50 & 1.92 & 3.89 & 1.11 & 0.77 \\
WebNLG-SFT        & 2.03 & 4.47 & 1.94 & 3.42 & 0.92 & 0.91 \\
DDKD (Ours)              & \textbf{0.69} & \textbf{0.22} & \textbf{0.72} & \textbf{1.67} & 0.56 & \textbf{0.09} \\
\bottomrule
\end{tabular}
\caption{
\textbf{Human evaluation results (average error count per output; lower is better).}
For each domain, we report the mean number of annotated errors per output, averaged over the three human annotators on the 12-example diagnostic set.
\textbf{NormAvg} is the mean of per-domain min--max normalized scores computed across the four evaluated systems (lower is better), to reduce scale differences between domains.
}
\label{tab:human_eval_results}
\end{table*}

\subsection{Inter-Annotator Agreement}
\label{sec:iaa}

To assess the reliability of our human evaluation, we calculated Inter-Annotator Agreement (IAA) among the three annotators. We utilise \textbf{Krippendorff's} $\alpha$ \cite{krippendorff1computing}, which is well-suited for multiple annotators ($N=3$) and can handle different data scales (nominal vs.\ interval). We compute agreement at four granularities to capture different aspects of the annotation quality:

\paragraph{Methodology.}
Based on the nature of the annotation tasks, we employed different metric functions for $\alpha$:
\begin{itemize}
    \item \textbf{Token-level (Nominal):} We treat the error span annotation as a token-classification task. For each token, we assign a label corresponding to the error type (0 for no error, 1-4 for specific error types). We use the \textit{nominal} metric, which treats any disagreement in error type as a mismatch.
    \item \textbf{Instance-level Counts (Interval):} We calculate the total number of errors annotated per example. Here, we use the \textit{interval} metric to account for the magnitude of disagreement (e.g., a disagreement between 1 and 2 errors is penalized less than 1 vs.\ 5 errors).
    \item \textbf{System-level Scoring (Interval):} We aggregate the error counts to the system level to verify if annotators consistently rank the models. The \textit{interval} metric is used on the average error counts per system.
    \item \textbf{Summarisation Label (Nominal):} For the overall summarisation quality tags (short, medium, no aggregation), we use the \textit{nominal} metric.
\end{itemize}

\paragraph{Results.}
The IAA results are detailed in Table~\ref{tab:iaa_detailed}. 
We observe \textbf{strong agreement} ($\alpha=0.90$) at the system level, indicating that annotators are highly consistent in assessing the overall quality and ranking of different systems.
The breakdown analysis reveals that this reliability is driven by \textbf{factual errors} (``Incorrect Fact'', $\alpha_{\text{sys}}=0.81$) and formatting issues (``Other'', $\alpha_{\text{sys}}=0.87$), which constitute the core of the evaluation.
In contrast, the \textbf{``Misleading''} category exhibits near-zero agreement ($\alpha \approx 0$), highlighting the inherent subjectivity of detecting subtle misleading information compared to hard factual errors.

At the \textbf{token level}, agreement is fair ($\alpha=0.37$). This lower score is attributed to two factors: 
(1) the subjectivity of the ``Misleading'' category, and 
(2) the typical difficulty of defining precise span boundaries in open-ended generation (e.g., selecting ``5000 mAh'' vs.\ ``Li-Ion 5000 mAh'').
Crucially, the high system-level agreement confirms that neither the span boundary fuzziness nor the subjectivity of the minority ``Misleading'' class compromises the validity of the final model rankings.


\subsection{LLM-Human Evaluation Agreement}
\label{app:llm_human_eval_agree}

We assess the reliability of LLM-based evaluation by comparing our primary automatic judge (GPT-5.1) with human judgments.
To validate this choice, we measure its agreement with human annotations using Pearson correlation.

\paragraph{Evaluation granularity.}
We focus on \emph{system-level} agreement rather than instance-level agreement.
This choice is motivated by two factors.
First, our primary goal is to compare the relative performance of different systems, making system-level ranking the most relevant signal.
Second, for each domain-system pair, only 12 test instances are available, which makes instance-level correlation highly unstable and sensitive to outliers.
In addition, when both human annotators and GPT-5.1 assign zero errors to all instances of a system, instance-level correlation becomes undefined.

\paragraph{Human annotation aggregation.}
Each instance is annotated by three human annotators.
Because error spans are not directly aggregatable across annotators, we compute agreement based on the \emph{average number of errors per system}.
This aggregation enables consistent comparison with GPT-5.1 predictions, while token-level or span-level agreement is infeasible under this annotation setup.

\paragraph{Correlation analysis.}
We compute Pearson correlation ($r$) between human-annotated and GPT-5.1-predicted error counts across four evaluated systems
(a small zero-shot model, a small source-domain SFT model, the best-performing large teacher model, and the best DDKD model).
A high system-level correlation indicates that the LLM produces a system ranking consistent with human judgment.

Table~\ref{tab:human_llm_agreement} reports the results.
We observe a strong agreement ($r > 0.90$) in the aggregate \textbf{All Errors} category for GSM-Arena, OWID, and Weather, confirming that GPT-5.1 reliably reproduces human system rankings in these domains.
Wikidata ($r = 0.83$) and Ice Hockey ($r = 0.73$) also exhibit positive alignment, though with moderately weaker correlation.

\paragraph{Breakdown by error type.}
Examining individual error categories reveals additional nuances.
For \textbf{Incorrect Fact} errors, correlations are exceptionally high in GSM-Arena ($r = 0.99$) and OWID ($r = 0.95$), indicating that GPT-5.1 is particularly effective at identifying factual hallucinations in structured settings.
In contrast, correlations are weaker for domains such as Wikidata, where factual errors are rare and system-level statistics are computed from a small number of evaluation instances, making correlation estimates less stable.
Finally, the \textbf{Misleading} category exhibits high variability and even negative correlations in some domains.
This aligns with our inter-annotator agreement analysis, suggesting that misleading errors are inherently subjective and difficult to evaluate consistently, even for human annotators.

Overall, the strong system-level agreement on aggregate error counts supports the use of GPT-5.1 as a reliable and scalable evaluation proxy in our experiments.

\begin{table}[!htbp]
\centering
\small
\setlength{\tabcolsep}{3.5pt}
\begin{tabular}{l c | c c c c}
\toprule
\multirow{2}{*}{\textbf{Domain}} & \textbf{Overall} & \multicolumn{4}{c}{\textbf{Breakdown by Error Type ($r$)}} \\
\cmidrule(lr){3-6}
& \textbf{All Errors} & \textbf{Fact} & \textbf{Unchk} & \textbf{Mis.} & \textbf{Oth.} \\
\midrule
\textbf{GSM-Arena}  & \textbf{0.941} & \textbf{0.997} & 0.502  & -0.316 & \textbf{0.998} \\
\textbf{OWID}       & \textbf{0.904} & \textbf{0.950} & 0.668  & 0.823  & \textbf{0.952} \\
\textbf{Weather}    & \textbf{0.941} & 0.730          & \textbf{0.960} & -0.422 & 0.884 \\
\textbf{Wikidata}   & 0.829          & -0.180         & 0.400  & 0.667  & 0.781 \\
\textbf{Ice Hockey} & 0.732          & 0.829          & \textbf{0.990} & \textbf{0.943} & 0.947 \\
\bottomrule
\end{tabular}
\caption{\textbf{System-Level Human--LLM Agreement (Pearson $r$).} Correlation between human and GPT-5.1 evaluations across 4 systems. \textbf{All Errors} is computed on the total error count. Breakdown columns correspond to: \textbf{Fact} (Incorrect Fact), \textbf{Unchk} (Not Checkable), \textbf{Mis.} (Misleading), and \textbf{Oth.} (Other). High correlations ($>0.9$) in \textbf{All Errors} indicate that GPT-5.1 largely reproduces the human system ranking.}
\label{tab:human_llm_agreement}
\end{table}

\subsection{Human Evaluation Discussion Summary}
\label{app:human_eval_dis_summary}

We can see (Table \ref{tab:iaa_detailed}) that Inter Annotator Agreement (IAA) between the three human annotators is strong at both the instance (Krippendorff $\alpha$ = 0.578) and the system (Krippendorff $\alpha$ = 0.902) level but that agreement varies substantially across error categories. While Incorrect Fact and Other exhibit high agreement ($\alpha$ = 0.810 and 0.871), Not Checkable and Misleading show much lower and even negative correlation reflecting a higher degree of subjectivity. 

Table \ref{tab:human_llm_agreement} further shows that the agreement between human evaluation and our primary automatic judge (GPT-5.1) is high at the system level (Pearson r $> 0.90$) confirming that GPT-5.1 reliably reproduces human system rankings. A detailed per-category analysis of human–LLM agreement is provided in Appendix \ref{app:llm_human_eval_agree}.

Finally, Appendix \ref{app:human_eval_res} and Table~\ref{tab:human_eval_results} reveal broadly consistent trends across the two evaluation settings (automatic vs. human-based). In both cases, knowledge distillation ranks first overall (cf. NormAvg column in Table \ref{tab:human_eval_results}), performance is lowest for all models on OpenWeather, and the small distilled models outperform the large teacher model on several domains. 
Some discrepancies are also observed, which we conjecture may stem from differences in evaluation scale (e.g., the domain exhibiting the fewest errors is WikiData under automatic evaluation but Ice Hockey under human evaluation) or from differences in model coverage, as the human evaluation considers a single distilled model per domain (cf. ~\ref{app:model-selection}), while the automatic evaluation includes multiple distilled variants. This latter difference may explain, for instance, why the performance gap between the distilled model and the large teacher appears larger in human evaluation.


\section{Qualitative Examples of Structured Inputs and Generated Texts}
\label{app:qualitative_example}
For each domain, we present one representative example consisting of the structured input and the generated text.
All outputs are produced by the final distilled model selected for that domain (Appendix~\ref{app:model-selection}),
which reflects the best-performing small model under our automatic and human evaluation protocols.

\subsection{Wikidata (Entity Description)}
\label{subsec:wikidata}
\noindent\textbf{Input:}
\begin{lstlisting}
Qin dynasty
---
- basic form of government: absolute monarchy
- capital: Xianyang
- continent: Asia
- instance of: Chinese dynasty
- instance of: historical Chinese state
- topic's main category: Qin dynasty
\end{lstlisting}
\noindent\textbf{Output:}
\par\noindent
The Qin dynasty was a Chinese dynasty and a historical Chinese state. It was located in Asia and had an absolute monarchy as its basic form of government. The capital of the Qin dynasty was Xianyang.
\par

\subsection{Ice Hockey (Game Summary)}
\label{subsec:ice-hockey}
\noindent\textbf{Input:}
\begin{myjson}
{
    'season': {'name': 'KHL 23/24', 'year': '23/24', 'id': 52523}, 
    'homeTeam': {'name': 'Sibir Novosibirsk', 'shortName': 'Sibir', 'nameCode': 'SIB', 'id': 3939, 'country': 'Russia'}, 
    'awayTeam': {'name': 'Severstal Cherepovets', 'shortName': 'Severstal', 'nameCode': 'SCH', 'id': 3943, 'country': 'Russia'}, 
    'homeScore': {'display': 1, 'period1': 0, 'period2': 1, 'period3': 0, 'normaltime': 1}, 
    'awayScore': {'display': 2, 'period1': 1, 'period2': 0, 'period3': 1, 'normaltime': 2}, 
    'id': 11386718, 'startTimestamp': 1701174600, 'startDatetime': '2023-11-28 12:30:00' 
}
\end{myjson}
\noindent\textbf{Output:}
\par\noindent
Sibir Novosibirsk (Russia) played against Severstal Cherepovets (Russia) in the KHL 23/24 season. The game started on 28 November 2023 at 12:30. The final score was 1-2 to Severstal Cherepovets.
\par

\subsection{OpenWeather (Weather Forecast)}

\noindent\textbf{Input:}
We have the OpenWeather input in Figure~\ref{fig:weather_input_pic}.

\begin{figure*}
\small
    \begin{myjson}
    {
        'city': {'id': 4157095, 'name': 'Goulding', 'coord': {'lat': 30.443, 'lon': -87.2225}, 'country': 'US'}, 
        'units': {'temp': '$^\circ$C', 'wind': 'm/s', 'pressure': 'hPa', 'rain': 'mm', 'snow': 'mm'}, 
        'list': [
            {'main': {'temp': 7.77, 'feels_like': 5.04, 'temp_min': 7.63, 'temp_max': 7.77, 'pressure': 1017}, 'weather': [{'id': 501, 'main': 'Rain', 'description': 'moderate rain', 'icon': '10d'}], 'clouds': {'all': 100}, 'wind': {'speed': 4.45, 'deg': 69, 'gust': 7.14}, 'rain': {'3h': 3.12}, 'dt_txt': '2024-01-03 12:00:00'}, 
            {'main': {'temp': 7.46, 'feels_like': 4.48, 'temp_min': 7.3, 'temp_max': 7.46, 'pressure': 1018}, 'weather': [{'id': 804, 'main': 'Clouds', 'description': 'overcast clouds', 'icon': '04n'}], 'clouds': {'all': 100}, 'wind': {'speed': 4.85, 'deg': 19, 'gust': 8.2}, 'dt_txt': '2024-01-03 18:00:00'}, 
            {'main': {'temp': 5.18, 'feels_like': 2.53, 'temp_min': 5.18, 'temp_max': 5.18, 'pressure': 1020}, 'weather': [{'id': 802, 'main': 'Clouds', 'description': 'scattered clouds', 'icon': '03n'}], 'clouds': {'all': 40}, 'wind': {'speed': 3.28, 'deg': 359, 'gust': 6.39}, 'dt_txt': '2024-01-04 00:00:00'}, 
            {'main': {'temp': 3.52, 'feels_like': -0.05, 'temp_min': 3.52, 'temp_max': 3.52, 'pressure': 1021}, 'weather': [{'id': 800, 'main': 'Clear', 'description': 'clear sky', 'icon': '01n'}], 'clouds': {'all': 3}, 'wind': {'speed': 4.15, 'deg': 4, 'gust': 9.34}, 'dt_txt': '2024-01-04 06:00:00'}, 
            {'main': {'temp': 12.12, 'feels_like': 10.54, 'temp_min': 12.12, 'temp_max': 12.12, 'pressure': 1022}, 'weather': [{'id': 800, 'main': 'Clear', 'description': 'clear sky', 'icon': '01d'}], 'clouds': {'all': 1}, 'wind': {'speed': 4.13, 'deg': 14, 'gust': 5.25}, 'dt_txt': '2024-01-04 12:00:00'}, 
            ...
            {'main': {'temp': 11.91, 'feels_like': 11.17, 'temp_min': 11.91, 'temp_max': 11.91, 'pressure': 1016}, 'weather': [{'id': 804, 'main': 'Clouds', 'description': 'overcast clouds', 'icon': '04n'}], 'clouds': {'all': 100}, 'wind': {'speed': 2.22, 'deg': 279, 'gust': 3.04}, 'dt_txt': '2024-01-07 00:00:00'}, 
            {'main': {'temp': 10.37, 'feels_like': 9.66, 'temp_min': 10.37, 'temp_max': 10.37, 'pressure': 1019}, 'weather': [{'id': 804, 'main': 'Clouds', 'description': 'overcast clouds', 'icon': '04n'}], 'clouds': {'all': 98}, 'wind': {'speed': 2.64, 'deg': 330, 'gust': 5.01}, 'dt_txt': '2024-01-07 06:00:00'}, 
            {'main': {'temp': 16.6, 'feels_like': 15.62, 'temp_min': 16.6, 'temp_max': 16.6, 'pressure': 1020}, 'weather': [{'id': 804, 'main': 'Clouds', 'description': 'overcast clouds', 'icon': '04d'}], 'clouds': {'all': 95}, 'wind': {'speed': 2.77, 'deg': 359, 'gust': 3.82}, 'dt_txt': '2024-01-07 12:00:00'}, 
            {'main': {'temp': 14.31, 'feels_like': 13.47, 'temp_min': 14.31, 'temp_max': 14.31, 'pressure': 1021}, 'weather': [{'id': 804, 'main': 'Clouds', 'description': 'overcast clouds', 'icon': '04n'}], 'clouds': {'all': 99}, 'wind': {'speed': 2.65, 'deg': 29, 'gust': 3.49}, 'dt_txt': '2024-01-07 18:00:00'}, 
            {'main': {'temp': 12.52, 'feels_like': 11.63, 'temp_min': 12.52, 'temp_max': 12.52, 'pressure': 1021}, 'weather': [{'id': 804, 'main': 'Clouds', 'description': 'overcast clouds', 'icon': '04n'}], 'clouds': {'all': 100}, 'wind': {'speed': 4.15, 'deg': 76, 'gust': 7.91}, 'dt_txt': '2024-01-08 00:00:00'}, 
            {'main': {'temp': 11.81, 'feels_like': 10.8, 'temp_min': 11.81, 'temp_max': 11.81, 'pressure': 1020}, 'weather': [{'id': 804, 'main': 'Clouds', 'description': 'overcast clouds', 'icon': '04n'}], 'clouds': {'all': 97}, 'wind': {'speed': 5.12, 'deg': 85, 'gust': 10.49}, 'dt_txt': '2024-01-08 06:00:00'}
        ]
    }
    \end{myjson}
\caption{Example of OpenWeather Input}
\label{fig:weather_input_pic}
\end{figure*}

\noindent\textbf{Output:}
\par\noindent
Goulding, US is expected to have a mix of rain and clouds over the next five days. On the 3rd of January, the temperature will be 7.77°C with moderate rain. The following day, the temperature will be 7.46°C with overcast clouds. On the 5th of January, the temperature will be 15.66°C with broken clouds. The next day, the temperature will be 16.88°C with light rain. On the 7th of January, the temperature will be 17.19°C with heavy intensity rain.
\par

\subsection{GSM Arena (Product Description)}
\label{subsec:gsm-arena}

\noindent\textbf{Input:}
We have the GSM Arena input in Figure~\ref{fig:gsmarena_input_pic}.


\begin{figure*}
\small
    \begin{myjson}
    {
        'id': 'lava_yuva_pro-11928', 'name': 'Yuva Pro', 
        'details': {
            'name': 'Lava Yuva Pro', 'img': 'https://fdn2.gsmarena.com/vv/bigpic/lava-yuva-pro.jpg', 
            'detailSpec': [
                {'category': 'Network', 'specifications': [{'name': 'Technology', 'value': 'GSM / HSPA / LTE'}, {'name': '2G bands', 'value': 'GSM 900 / 1800 - SIM 1 & SIM 2'}, {'name': '3G bands', 'value': 'HSDPA 900 / 2100 '}, {'name': '4G bands', 'value': '1, 3, 5, 8, 40, 41'}, {'name': 'Speed', 'value': 'HSPA, LTE'}]}, 
                {'category': 'Launch', 'specifications': [{'name': 'Announced', 'value': '2022, October 11'}, {'name': 'Status', 'value': 'Available. Released 2022, October 11'}]},
                {'category': 'Body', 'specifications': [{'name': 'Dimensions', 'value': '164.4 x 75.8 x 8.9 mm (6.47 x 2.98 x 0.35 in)'}, {'name': 'Weight', 'value': '191 g (6.74 oz)'}, {'name': 'SIM', 'value': 'Dual SIM (Nano-SIM, dual stand-by)'}]},
                {'category': 'Display', 'specifications': [{'name': 'Type', 'value': 'IPS LCD'}, {'name': 'Size', 'value': '6.52 inches, 102.6 cm2 (~82.4% screen-to-body ratio)'}, {'name': 'Resolution', 'value': '720 x 1600 pixels, 20:9 ratio (~269 ppi density)'}]}, 
                {'category': 'Platform', 'specifications': [{'name': 'OS', 'value': 'Android 12'}]}, 
                {'category': 'Memory', 'specifications': [{'name': 'Card slot', 'value': 'microSDXC (dedicated slot)'}, {'name': 'Internal', 'value': '32GB 3GB RAM'}, {'name': '\\xa0', 'value': 'eMMC 5.1'}]}, 
                {'category': 'Main Camera', 'specifications': [{'name': 'Triple', 'value': '13 MP, (wide), AF\\n0.08 MP, (depth)\\n0.08 MP'}, {'name': 'Features', 'value': 'LED flash'}, {'name': 'Video', 'value': '1080p@30fps'}]}, 
                {'category': 'Selfie camera', 'specifications': [{'name': 'Single', 'value': '8 MP'}, {'name': 'Video', 'value': 'Yes'}]}, 
                {'category': 'Sound', 'specifications': [{'name': 'Loudspeaker ', 'value': 'Yes'}, {'name': '3.5mm jack ', 'value': 'Yes'}]}, 
                {'category': 'Comms', 'specifications': [{'name': 'WLAN', 'value': 'Wi-Fi 802.11 b/g/n/ac'}, {'name': 'Bluetooth', 'value': '5.0, A2DP, LE'}, {'name': 'Positioning', 'value': 'GPS'}, {'name': 'NFC', 'value': 'No'}, {'name': 'Radio', 'value': 'FM radio'}, {'name': 'USB', 'value': 'USB Type-C 2.0, OTG'}]}, 
                {'category': 'Features', 'specifications': [{'name': 'Sensors', 'value': 'Fingerprint (side-mounted), accelerometer, proximity'}]}, 
                {'category': 'Battery', 'specifications': [{'name': 'Type', 'value': 'Li-Po 5000 mAh, non-removable'}, {'name': 'Charging', 'value': '10W wired'}]}, 
                {'category': 'Misc', 'specifications': [{'name': 'Colors', 'value': 'Metallic Black, Metallic Blue, Metallic Grey'}, {'name': 'Models', 'value': 'LIX402'}, {'name': 'Price', 'value': 'About 100 EUR'}]}
            ], 
            'quickSpec': [
                {'name': 'Display size', 'value': '6.52\"'}, 
                {'name': 'Display resolution', 'value': '720x1600 pixels'}, 
                {'name': 'Camera pixels', 'value': '13MP\\n      '}, 
                {'name': 'Video pixels', 'value': '1080p'}, 
                {'name': 'RAM size', 'value': '3GB RAM'}, 
                {'name': 'Chipset', 'value': ''}, 
                {'name': 'Battery size', 'value': '5000mAh'}, 
                {'name': 'Battery type', 'value': 'Li-Po'}
            ]
        }
    }
    \end{myjson}
\caption{Example of GSM Arena Input}
\label{fig:gsmarena_input_pic}
\end{figure*}

\noindent\textbf{Output:}
\par\noindent
The Lava Yuva Pro is a smartphone that was announced on October 11, 2022, and is currently available. It features a 6.52-inch IPS LCD display with a resolution of 720 x 1600 pixels and a 20:9 aspect ratio. The device is powered by Android 12 and comes with 3GB of RAM and 32GB of internal storage, which can be expanded using a microSDXC card. The Yuva Pro is equipped with a triple rear camera setup consisting of a 13 MP wide-angle lens, a 0.08 MP depth sensor, and another 0.08 MP lens, along with an 8 MP front-facing camera. It supports 4G LTE connectivity with bands 1, 3, 5, 8, 40, and 41, and also includes Wi-Fi 802.11 b/g/n/ac, Bluetooth 5.0, GPS, FM radio, and USB Type-C 2.0 with OTG support. The phone is powered by a non-removable Li-Po 5000 mAh battery that supports 10W wired charging. It includes a side-mounted fingerprint sensor, accelerometer, and proximity sensor. The Yuva Pro is available in Metallic Black, Metallic Blue, and Metallic Grey colors and is priced around 100 EUR.
\par

\subsection{OWID (Chart Captioning)}
\label{subsec:owid}
\noindent\textbf{Input:}
\begin{lstlisting}[basicstyle=\ttfamily\small]
# country: Thailand\n# title: People vaccinated against COVID-19 per hundred
# description: Share of the population that received a first COVID-19 vaccine dose.
# unit: people
date,value
2021-02-27,0.0
2021-03-02,0.01
2021-03-13,0.08
2021-03-30,0.21
2021-04-02,0.26
...
2021-10-13,50.07
...
2022-03-30,77.36
2022-06-30,79.10
2022-09-30,79.51
\end{lstlisting}

\noindent\textbf{Output:}
\par\noindent
The number of people vaccinated against COVID-19 per hundred in Thailand increased from 0.0 on 27th February 2021 to 79.51 on 30th September 2022.
\par

\section{Content Coverage Evaluation}
\label{app:coverage}

Our primary evaluation focuses on hallucination-related errors. To ensure that error reductions are not obtained by degenerate strategies (e.g., abnormally short outputs or systematically omitting content), we additionally conduct a sanity check on \emph{content coverage} between the structured input and the generated text, complementing the output-length analysis in Table~\ref{tab:output_length_tokens_5domains}.
Because coverage evaluation for cross-domain data-to-text is underexplored and non-trivial across heterogeneous schemas and formats, we use LLM-as-a-judge to obtain a coarse but scalable coverage signal.
We employ two strong judges, \textbf{GPT-5.1} and \textbf{Gemini-2.5-Pro}, and evaluate the same set of systems as in our human study (Section~\ref{subsec:human-eval}): a small zero-shot model, a small source-domain SFT model, our best distilled model (\textbf{DDKD-Best}), and the selected best large teacher.

\subsection{LLM Judge Setup}
\label{app:coverage_setup}

\subsubsection{Prompt}
\label{app:coverage_prompt}

\begin{tcolorbox}[breakable,enhanced]
\small
\textbf{System:} You are an expert evaluator for data-to-text generation. Your job is to judge \textbf{CONTENT COVERAGE only} (i.e., omissions / missing content), not fluency or style.

\vspace{1mm}
\textbf{User:}

\textbf{Task flag:}
\begin{itemize}\itemsep0.1em
    \item \textbf{DOMAIN} = \texttt{\{DOMAIN\}}
    \item \textbf{Generation goal (task hint):}
    \begin{itemize}\itemsep0.1em
        \item Wikidata: Structured entity description (entity profile; aim for completeness).
        \item Ice Hockey: Event-based game summary (cover teams, score, key stats/events).
        \item OpenWeather: Time-series weather forecast reporting (cover location/time range and salient conditions/extremes/trends).
        \item GSM Arena: Entity-centric specification description (cover key specs and distinguishing attributes).
        \item OWID: Chart captioning from tabular data (cover what is measured, where/when, and salient numeric patterns/trends).
    \end{itemize}
\end{itemize}

\textbf{Rules:}
\begin{itemize}\itemsep0.1em
    \item If \textbf{DOMAIN = Wikidata}, treat \emph{all} input facts as important and judge completeness.
    \item Otherwise, judge coverage over important content consistent with the generation goal above.
\end{itemize}

\textbf{[Input structured data]}
\begin{quote}\small\ttfamily
<<< \{INPUT\_DATA\} >>>
\end{quote}

\textbf{[Generated text]}
\begin{quote}\small\ttfamily
<<< \{GEN\_TEXT\} >>>
\end{quote}

\textbf{Task:} Rate how well the generated text covers the \textbf{IMPORTANT} content in the input data.

\textbf{General rules:}
\begin{itemize}\itemsep0.1em
    \item Focus on content coverage only. Ignore writing quality and minor paraphrasing differences.
    \item Do \textbf{not} reward verbosity. Longer text is not better unless it covers more important content.
    \item If the text repeats the same fact, count it once.
    \item For very long numeric inputs (e.g., OWID-like tables), evaluate whether the text covers:
    (i) what the data is about (metadata such as metric/topic/region/time range), and
    (ii) salient numeric patterns (e.g., latest value, max/min, notable change/trend),
    rather than exhaustive row-by-row coverage.
\end{itemize}

\textbf{Provide:}
\begin{enumerate}\itemsep0.1em
    \item A \textbf{coverage score} from 1 to 5 using the rubric below.
    \item An \textbf{estimated coverage ratio} in $[0,1]$, consistent with the score.
    For Wikidata: proportion of all input facts expressed; for other domains: proportion of important content points expressed.
    \item Up to 3 most important missing points ($\le$20 words each), labeled as \texttt{ABSENT} or \texttt{PARTIAL}.
\end{enumerate}

\textbf{Rubric (score + approximate proportion):}
\begin{itemize}\itemsep0.1em
    \item 5: Almost all important content covered; no notable omissions ($\sim$90--100\%).
    \item 4: Most important content covered; a few omissions or missing details ($\sim$70--90\%).
    \item 3: Some important content covered, but several key omissions ($\sim$40--70\%).
    \item 2: Little important content covered; most key information omitted ($\sim$10--40\%).
    \item 1: No meaningful coverage (empty/degenerate/off-topic; $<$10\%).
\end{itemize}

\textbf{Output JSON only:}
\begin{quote}\small\ttfamily
\{\\
\ \ "domain": "\{DOMAIN\}",\\
\ \ "coverage\_score": 1--5,\\
\ \ "estimated\_coverage\_ratio": 0.0--1.0,\\
\ \ "missing\_points": [\{"point": "...", "status": "ABSENT|PARTIAL"\}],\\
\ \ "notes": "one-sentence justification focused on coverage"\\
\}
\end{quote}
\end{tcolorbox}

\subsubsection{Scoring and Normalization}
\label{app:coverage_scoring}

For each instance, the judge returns (i) an ordinal \textbf{Coverage Score} (1--5) and (ii) a continuous \textbf{Estimated Coverage Ratio} in $[0,1]$.
For Wikidata entity descriptions, the judge is instructed to evaluate \emph{completeness} over all provided facts; for other domains, the judge evaluates coverage over task-relevant important content (salient entities/attributes and salient numeric patterns) without requiring exhaustive row-level matching for long tabular inputs.

We report mean Score and mean Ratio per domain and system (Tables~\ref{tab:coverage_gpt51}--\ref{tab:coverage_gemini25}).
To enable a clearer \emph{cross-system} comparison aggregated across heterogeneous domains, we additionally compute \textbf{NormAvg}.
For a fixed judge and a fixed domain, we min--max normalize the \emph{mean coverage scores} across the systems:
\[
\mathrm{Norm}(s,d) = \frac{\mu(s,d)-\min_{s'}\mu(s',d)}{\max_{s'}\mu(s',d)-\min_{s'}\mu(s',d)} ,
\]
where $\mu(s,d)$ is the mean Coverage Score for system $s$ on domain $d$.
We then average the normalized values across domains:
\[
\mathrm{NormAvg}(s)=\frac{1}{|D|}\sum_{d\in D}\mathrm{Norm}(s,d).
\]
NormAvg is used only as a compact summary for cross-system comparison (higher is better).

\subsubsection{LLM Judge Results}
\label{app:coverage_results}

\begin{figure*}[t]
    \centering
    \includegraphics[width=\linewidth]{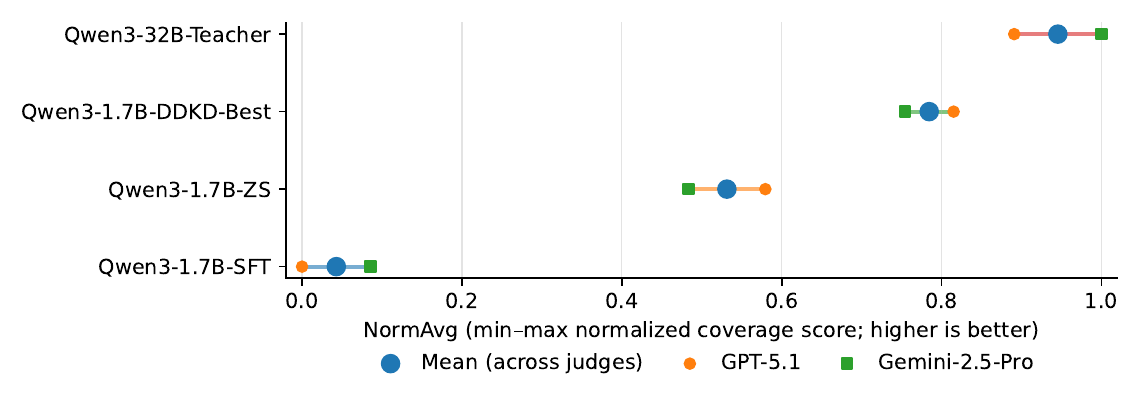}
    \caption{System-level content coverage sanity check summarized by NormAvg (domain-wise min--max normalized coverage \emph{scores} averaged across domains). Each row shows the mean NormAvg across two LLM judges (dot) and the judge-specific values (markers). DDKD-Best consistently outperforms same-size baselines and remains close to the selected best teacher, indicating gains are not achieved via content omission.}
    \label{fig:coverage_normavg}
\end{figure*}

Figure~\ref{fig:coverage_normavg} summarizes coverage using NormAvg.
We observe consistent rankings across the two judges: the selected best teacher achieves the highest coverage, while \textbf{DDKD-Best} remains close to the teacher and clearly outperforms same-size baselines (zero-shot and source-domain SFT), suggesting that performance gains are not obtained by systematically omitting content.
Full per-domain Score/Ratio results are provided in Tables~\ref{tab:coverage_gpt51}--\ref{tab:coverage_gemini25}.

\subsubsection{Inter-Judge Agreement}
\label{app:coverage_agreement}

We assess the robustness of our coverage sanity check to the choice of LLM judge by measuring agreement between \textbf{GPT-5.1} and \textbf{Gemini-2.5-Pro} at two granularities.
\textbf{(i) Instance-level agreement} pools aligned instances across all matched systems within each domain and computes Krippendorff's $\alpha$ (Table~\ref{tab:coverage_agreement_instance}).
We use an ordinal metric for Coverage Score (1--5) and an interval metric for the continuous Estimated Coverage Ratio (0--1).
Across domains, the judges exhibit moderate-to-strong agreement at the instance level.
Agreement is highest on Wikidata ($\alpha_{\text{ord}}{=}0.744$, $\alpha_{\text{int}}{=}0.835$), and lower on Ice Hockey and Weather, reflecting greater subjectivity in selecting task-salient content for narrative summaries and long numeric inputs.
Importantly, when pooling all instances across domains and systems, agreement increases to $\alpha_{\text{ord}}{=}0.732$ (Score) and $\alpha_{\text{int}}{=}0.794$ (Ratio), indicating that the two judges are broadly consistent overall despite domain-specific ambiguity.
\textbf{(ii) System-level agreement} compares the two judges on \emph{per-system mean} coverage judgments (Table~\ref{tab:coverage_agreement_system}), reporting Pearson $r$ and Spearman $\rho$.
Despite the small number of systems per domain (4), we observe strong cross-judge consistency in most domains (typically $r,\rho \ge 0.8$), and very high pooled agreement across all (domain, system) pairs ($r$=0.959 and $\rho$=0.951 for Score; $r$=0.961 and $\rho$=0.959 for Ratio).
All pooled correlations are statistically significant ($p < 0.001$, two-tailed; computed using \texttt{scipy.stats}).
On Wikidata, system-level mean \emph{scores} show limited variance (near saturation), which reduces correlation for the discrete score, while the continuous ratio still preserves meaningful rank agreement ($r$=0.789, $\rho$=0.800).
Overall, these results indicate that our coverage sanity check is not overly sensitive to the specific judge and supports the qualitative conclusions drawn from NormAvg visualisations.

\begin{table}[t]
\centering
\small
\begin{tabular}{@{}lcccc@{}}
\toprule
\textbf{Domain} & $\alpha_{\text{ord}}$ & $\alpha_{\text{int}}$ & \textbf{\#Pairs} & \textbf{\#Skip} \\
\midrule
OWID       & 0.525 & 0.504 & 396 & 4 \\
GSM Arena  & 0.617 & 0.670 & 396 & 4 \\
Weather    & 0.373 & 0.472 & 391 & 9 \\
Ice Hockey & 0.283 & 0.524 & 395 & 5 \\
Wikidata   & 0.744 & 0.835 & 398 & 2 \\
\midrule
\textbf{ALL} & \textbf{0.732} & \textbf{0.794} & \textbf{1,976} & 20 \\
\bottomrule
\end{tabular}
\caption{Instance-level inter-judge agreement between \textbf{GPT-5.1} and \textbf{Gemini-2.5-Pro}. $\alpha_{\text{ord}}$ is computed on Coverage Score (1--5, ordinal) and $\alpha_{\text{int}}$ on Estimated Coverage Ratio (0--1, interval). \#Skip counts instances excluded due to parsing errors or missing values.}
\label{tab:coverage_agreement_instance}
\end{table}

\begin{table}[t]
\centering
\small
\begin{tabular}{@{}lccccc@{}}
\toprule
\textbf{Domain} & $r_s$ & $\rho_s$ & $r_r$ & $\rho_r$ & \textbf{\#Sys} \\
\midrule
OWID       & 0.924 & 1.000 & 0.916 & 1.000 & 4 \\
GSM Arena  & 0.974 & 0.800 & 0.987 & 1.000 & 4 \\
Weather    & 0.963 & 0.800 & 0.975 & 0.800 & 4 \\
Ice Hockey & 1.000 & 1.000 & 1.000 & 1.000 & 4 \\
Wikidata   & 0.000 & 0.000 & 0.789 & 0.800 & 4 \\
\midrule
\textbf{ALL} & \textbf{0.959} & \textbf{0.951} & \textbf{0.961} & \textbf{0.959} & \textbf{20} \\
\bottomrule
\end{tabular}
\caption{System-level inter-judge agreement between \textbf{GPT-5.1} and \textbf{Gemini-2.5-Pro}. $r_s/\rho_s$ are correlations on per-system mean \emph{scores} and $r_r/\rho_r$ on per-system mean \emph{ratios}.}
\label{tab:coverage_agreement_system}
\end{table}

\section{Initialization Dataset Ablation}
\label{app:init_ablation}

\subsection{Motivation and Setup}
\label{app:init_setup}

Our main experiments use WebNLG as the source-domain dataset for initializing teacher models before transfer to QUINTD-1.
To assess the sensitivity of cross-domain transfer to this design choice, we compare three source-domain initialization datasets: \textbf{WebNLG}, \textbf{E2E}, and \textbf{KELM-Q1}.
For each dataset, we fine-tune the same teacher model (\texttt{Qwen3-32B}) under the same training setup, and directly evaluate the resulting source-initialized model on the five QUINTD-1 target domains.
We report two complementary metrics, following the main evaluation setup: \textbf{average error numbers} from the faithfulness evaluation (lower is better) and \textbf{content coverage score} from the \textbf{GPT-5.1} judge (higher is better).
For simplicity, we report GPT-5.1 coverage scores in this ablation, as GPT-5.1 and Gemini-2.5-Pro showed highly consistent rankings in our main inter-judge agreement analysis.
This experiment isolates the effect of the source-domain initialization dataset, without involving additional target-domain distillation or augmentation.

\subsection{Compared Source Datasets}
\label{app:init_datasets}

Table~\ref{tab:init_dataset_summary} summarizes the three source-domain datasets compared in our initialization ablation.

\begin{table}[t]
\centering
\small
\begin{tabular}{@{}lccc@{}}
\toprule
\textbf{Dataset} & \textbf{\#Inst.} & \textbf{Domain Scope} & \textbf{Reference} \\
\midrule
WebNLG  & 39,890 & 16 DBpedia categories & Human \\
E2E     & 46,733 & Restaurant only       & Human \\
KELM-Q1 & 18,723 & Wikidata/Wikipedia    & Automatic \\
\bottomrule
\end{tabular}
\caption{Comparison of source-domain initialization datasets used in our ablation study. Statistics correspond to the actual training set used in our experiments.}
\label{tab:init_dataset_summary}
\end{table}

\paragraph{WebNLG.}
WebNLG is a graph-to-text benchmark with human-written, fact-aligned references derived from DBpedia triples.
Its broad semantic coverage and structured input format make it a strong candidate for learning general data-to-text generation capabilities. We describe this dataset in Section~\ref{sec:data}.

\paragraph{E2E.}
E2E is a meaning-representation-to-text benchmark in the restaurant domain.
Although its references are also human-written and high-quality, its semantic scope is substantially narrower than WebNLG, which may limit transfer to heterogeneous target domains.

\paragraph{KELM-Q1.}
We do not use the full KELM dataset, whose scale is much larger and whose automatically constructed texts are often noisy.
Instead, following \citet{song-gardent-2025-mucal}, we use \textbf{KELM-Q1}, a filtered subset selected using embedding-based and QA-based metrics to improve data-text alignment.
Despite this filtering, its references remain automatically constructed and generally less reliable than the human-written references in WebNLG and E2E.

\begin{table*}[t]
\centering
\small
\begin{tabular}{@{}lcccccccccccc@{}}
\toprule
\multirow{2}{*}{\textbf{Source Init.}} &
\multicolumn{2}{c}{\textbf{Wikidata}} &
\multicolumn{2}{c}{\textbf{IceHockey}} &
\multicolumn{2}{c}{\textbf{OpenWeather}} &
\multicolumn{2}{c}{\textbf{GSMArena}} &
\multicolumn{2}{c}{\textbf{OWID}} &
\multirow{2}{*}{\textbf{NE}$\uparrow$} &
\multirow{2}{*}{\textbf{NC}$\uparrow$} \\
\cmidrule(lr){2-3}\cmidrule(lr){4-5}\cmidrule(lr){6-7}\cmidrule(lr){8-9}\cmidrule(lr){10-11}
& \textbf{Err}$\downarrow$ & \textbf{Cov}$\uparrow$
& \textbf{Err}$\downarrow$ & \textbf{Cov}$\uparrow$
& \textbf{Err}$\downarrow$ & \textbf{Cov}$\uparrow$
& \textbf{Err}$\downarrow$ & \textbf{Cov}$\uparrow$
& \textbf{Err}$\downarrow$ & \textbf{Cov}$\uparrow$
&  &  \\
\midrule
\textbf{WebNLG}
& \textbf{0.09} & \textbf{4.61}
& 0.02 & \textbf{3.99}
& \textbf{1.86} & \textbf{2.76}
& 0.65 & \textbf{3.23}
& \textbf{0.43} & \textbf{3.41}
& \textbf{0.954} & \textbf{1.000} \\

\textbf{E2E}
& \textbf{0.09} & 4.38
& \textbf{0.00} & 2.56
& 3.84 & 2.07
& \textbf{0.39} & 2.43
& 0.90 & 2.18
& 0.828 & 0.066 \\

\textbf{KELM}
& 0.31 & 4.40
& 0.39 & 2.93
& 5.06 & 2.08
& 1.86 & 2.21
& 2.39 & 2.02
& 0.000 & 0.072 \\
\bottomrule
\end{tabular}
\caption{
Sensitivity to the source-domain initialization dataset.
We fine-tune the same teacher model (\texttt{Qwen3-32B}) on three source datasets and directly evaluate transfer to the five QUINTD-1 target domains.
\textbf{Err} denotes the average error number from our faithfulness evaluation (lower is better), and \textbf{Cov} denotes the content coverage score assigned by \textbf{GPT-5.1} (higher is better).
\textbf{NE} and \textbf{NC} denote domain-wise min--max normalized averages for Err and Cov, respectively; for Err, reversed min--max normalization is used so that higher is better.
WebNLG provides the strongest overall initialization, achieving the most favorable trade-off between factual correctness and coverage across heterogeneous target domains.
}
\label{tab:init_dataset_ablation}
\end{table*}

\subsection{Results}

Table~\ref{tab:init_dataset_ablation} shows that \textbf{WebNLG} provides the strongest overall source-domain initialization, achieving the best aggregated performance under both normalized summaries (\textbf{NE}=0.954, \textbf{NC}=1.000).
This indicates that WebNLG offers the most favorable trade-off between factual correctness and content coverage when transferring to the heterogeneous QUINTD-1 target domains.

Compared with WebNLG, \textbf{E2E} yields competitive or even slightly lower error counts on several relatively shorter or structurally simpler domains, such as Wikidata, Ice Hockey, and GSM Arena.
However, these gains are accompanied by substantially lower content coverage, especially on Ice Hockey (3.99 vs.\ 2.56) and GSM Arena (3.23 vs.\ 2.43).
This pattern suggests that E2E-initialized teachers may benefit from producing shorter and simpler outputs, but at the cost of increased content omission.
A plausible explanation is that E2E itself is narrowly scoped and consists of relatively short meaning representations and references, which may encourage conservative generation behavior that transfers poorly to more information-dense domains.

\textbf{KELM-Q1} performs worse than both WebNLG and E2E across most domains in terms of both error rate and coverage.
This is consistent with the fact that, even after filtering, KELM-Q1 remains based on automatically constructed text, which likely provides noisier and less tightly aligned supervision than human-written datasets such as WebNLG and E2E.

Overall, these results support that our choice of \textbf{WebNLG} is not arbitrary.
Among the compared source datasets, WebNLG provides the best balance between \emph{broad semantic coverage} and \emph{high-quality, fact-aligned references}, making it particularly suitable for learning transferable data-to-text generation capabilities rather than overfitting to narrow source-domain patterns.

\section{LLM-as-a-Judge Results}
We present all LLM judge tables in this section, including faithfulness/hallucination-related evaluation and content coverage sanity check. The result tables are as follows:
\begin{itemize}
    \item Table~\ref{tab:llm_judge_error_types_agg_gpt51}-Table~\ref{tab:output_length_tokens_5domains}: Qwen3 and Gemma3 result summary plus output length analysis.
    \item Table~\ref{tab:wikidata_error_counts}-Table~\ref{tab:error_rate_wikidata_gemini_gemma3}: Qwen3 and Gemma3 error analysis on Wikidata.
    \item Table~\ref{tab:error_count_icehockey_gpt_qwen3}-Table~\ref{tab:error_rate_icehockey_gemini_gemma3}: Qwen3 and Gemma3 error analysis on Ice Hockey.
    \item Table~\ref{tab:error_count_weather_gpt_qwen3}-Table~\ref{tab:error_rate_weather_gemini_gemma3}: Qwen3 and Gemma3 error analysis on OpenWeather.
    \item Table~\ref{tab:error_count_gsmarena_gpt_qwen3}-Table~\ref{tab:error_rate_gsmarena_gemini_gemma3}: Qwen3 and Gemma3 error analysis on GSMArena.
    \item Table~\ref{tab:error_count_owid_gpt_qwen3}-Table~\ref{tab:error_rate_owid_gemini_gemma3}: Qwen3 and Gemma3 error analysis on OWID.
    \item Table~\ref{tab:coverage_gpt51}-Table~\ref{tab:coverage_gemini25}: Content coverage across domains.
    
\end{itemize}

\clearpage
\begin{table*}[!htbp]
\centering
\small
\setlength{\tabcolsep}{4pt}
\renewcommand{\arraystretch}{1.05}
\begin{tabular}{lcccccc}
\toprule
\textbf{System} &
\textbf{Incorrect Fact} &
\textbf{Not Checkable} &
\textbf{Misleading} &
\textbf{Other} &
\textbf{All Categories} &
\textbf{NormAll} $\downarrow$ \\
\midrule

\multicolumn{7}{l}{\textbf{\textit{Zero-shot (ZS)}}} \\
GPT-4.1-ZS  & 0.14 & 0.47 & 0.07 & \underline{0.00} & 0.68 & 0.16 \\
Qwen3-1.7B-ZS  & 1.54 & 0.53 & 0.21 & 0.04 & 2.31 & 0.64 \\
Qwen3-8B-ZS    & 0.75 & 0.38 & 0.16 & 0.03 & 1.32 & 0.30 \\
Qwen3-32B-ZS   & 0.50 & 0.35 & 0.13 & 0.04 & 1.02 & 0.25 \\

\hdashline
\multicolumn{7}{l}{\textbf{\textit{WebNLG-SFT}}} \\
Qwen3-1.7B-SFT & 3.51 & 0.70 & 0.09 & 0.37 & 4.66 & 0.81 \\
Qwen3-8B-SFT   & 1.18 & 0.14 & 0.06 & 0.15 & 1.53 & 0.23 \\
Qwen3-32B-SFT  & \underline{0.41} & 0.08 & \underline{0.05} & 0.06 & \underline{0.61} & \underline{0.04} \\

\hdashline
\multicolumn{7}{l}{\textbf{\textit{DDKD from ZS teacher}}} \\
Qwen3-1.7B-DDKD-Base  & 1.13 & 0.34 & 0.16 & \textbf{0.04} & 1.68 & 0.45 \\
Qwen3-1.7B-DDKD-Sub   & 0.68 & 0.31 & 0.13 & 0.05 & 1.16 & 0.30 \\
Qwen3-1.7B-DDKD-Pert  & \textbf{0.66} & 0.33 & 0.13 & 0.05 & 1.16 & 0.27 \\
Qwen3-1.7B-DDKD-Mixed & \textbf{0.66} & 0.32 & 0.16 & 0.05 & 1.20 & 0.28 \\

\hdashline
\multicolumn{7}{l}{\textbf{\textit{DDKD from WebNLG-SFT teacher}}} \\
Qwen3-1.7B-DDKD-Base  & 1.88 & 0.32 & 0.10 & 0.22 & 2.52 & 0.47 \\
Qwen3-1.7B-DDKD-Sub   & \textbf{0.66} & 0.12 & \underline{\textbf{0.05}} & 0.15 & \textbf{0.98} & 0.20 \\
Qwen3-1.7B-DDKD-Pert  & 0.86 & 0.07 & 0.21 & 0.23 & 1.38 & 0.19 \\
Qwen3-1.7B-DDKD-Mixed & 0.90 & \underline{\textbf{0.05}} & 0.10 & 0.12 & 1.17 & \textbf{0.10} \\

\bottomrule
\end{tabular}
\caption{
\textbf{Qwen3 Error-Type Result Summary by GPT-5.1}. 
LLM-as-a-judge evaluation (error-type breakdown): macro-averaged number of errors per output across all target domains (\emph{lower is better}).
Results are based on GPT-5.1 as the judge.
\textbf{Bold} marks the best 1.7B system per column (ties allowed);
\underline{underline} marks the best overall system per column (ties allowed).
\textbf{NormAll} is the mean of per-domain min--max normalized \textit{All Categories} error counts, where normalization is performed independently within each domain across all systems to ensure equal domain contribution (\emph{lower is better}).
}
\label{tab:llm_judge_error_types_agg_gpt51}
\end{table*}


\begin{table*}[!htbp]
\centering
\small
\renewcommand{\arraystretch}{1.05}
\begin{tabular}{lccccc}
\toprule
\textbf{System} &
\textbf{Wikidata} &
\textbf{Ice Hockey} &
\textbf{OpenWeather} &
\textbf{GSM Arena} &
\textbf{OWID} \\
\midrule

\multicolumn{6}{l}{\textbf{\textit{Zero-shot (ZS)}}} \\
Qwen3-1.7B-ZS   & 0.98 & 1.71 & 7.19 & 1.34 & 2.57 \\
Qwen3-8B-ZS     & 0.59 & 0.62 & 4.22 & 0.69 & 1.67 \\
Qwen3-32B-ZS    & 0.53 & 0.49 & 3.69 & \underline{0.10} & 1.60 \\

\hdashline
\multicolumn{6}{l}{\textbf{\textit{WebNLG-SFT}}} \\
Qwen3-1.7B-SFT  & 0.70 & 2.42 & 10.83 & 2.67 & 2.52 \\
Qwen3-8B-SFT    & 0.46 & 0.38 & 4.96 & 1.60 & 1.32 \\
Qwen3-32B-SFT   & 0.38 & 0.12 & \underline{2.40} & 0.67 & 1.07 \\

\hdashline
\multicolumn{6}{l}{\textbf{\textit{DDKD from ZS teacher}}} \\
Qwen3-1.7B-DDKD-Base   & 0.74 & 0.86 & 5.13 & 0.92 & 2.27 \\
Qwen3-1.7B-DDKD-Sub    & 0.62 & 0.58 & 4.28 & 0.61 & 2.03 \\
Qwen3-1.7B-DDKD-Pert   & 0.51 & 0.72 & 4.13 & 0.41 & 2.17 \\
Qwen3-1.7B-DDKD-Mixed  & 0.60 & 0.55 & 4.34 & \textbf{0.32} & 1.98 \\

\hdashline
\multicolumn{6}{l}{\textbf{\textit{DDKD from WebNLG-SFT teacher}}} \\
Qwen3-1.7B-DDKD-Base   & 0.48 & 0.38 & 8.63 & 1.32 & 3.22 \\
Qwen3-1.7B-DDKD-Sub    & 0.50 & \textbf{0.11} & \textbf{4.33} & 0.86 & 1.84 \\
Qwen3-1.7B-DDKD-Pert   & 0.40 & 0.19 & 4.86 & 1.29 & 1.02 \\
Qwen3-1.7B-DDKD-Mixed  & \textbf{0.36} & 0.13 & 4.65 & 0.77 & \textbf{0.85} \\

\bottomrule
\end{tabular}
\caption{\textbf{Qwen3 Result Summary by Gemini-2.5-Pro}.
LLM-as-a-judge evaluation (\textbf{all categories}):
\textbf{average number of errors per output} on the five QUINTD-1 target domains
(\emph{lower is better}).
\textbf{Bold} indicates the best \emph{small model} (1.7B) per domain,
while \underline{underlined} values indicate the best overall system.
All error categories (Incorrect Fact, Not Checkable, Misleading, Other) are counted and summed.
Results are based on \textbf{Gemini-2.5-Pro} as the judge.
}
\label{tab:llm_judge_all_errnum_5domains_gemini}
\end{table*}

\clearpage


\begin{table*}[!htbp]
\centering
\small
\setlength{\tabcolsep}{4pt}
\renewcommand{\arraystretch}{1.05}
\begin{tabular}{lccccc}
\toprule
\textbf{System} &
\textbf{Wikidata} &
\textbf{Ice Hockey} &
\textbf{OpenWeather} &
\textbf{GSM Arena} &
\textbf{OWID} \\
\midrule

\multicolumn{6}{l}{\textbf{\textit{Zero-shot (ZS)}}} \\
Gemma3-1B-IT-ZS   & 1.71 & 2.80 & 4.66 & 4.03 & 3.14 \\
Gemma3-27B-IT-ZS    & 0.56 & 0.28 & 4.98 & 0.44 & 0.90 \\

\hdashline
\multicolumn{6}{l}{\textbf{\textit{WebNLG-SFT}}} \\
Gemma3-1B-IT-SFT  & 5.49 & 5.34 & 5.48 & 3.60 & 3.86 \\
Gemma3-27B-IT-SFT    & 0.12 & 0.20 & 2.48 & 0.37 & 0.77 \\

\hdashline
\multicolumn{6}{l}{\textbf{\textit{The best DDKD Model}}} \\
Gemma3-1B-IT-DDKD-Best   & 0.28 & 0.38 & 4.23 & 1.57 & 0.70 \\

\bottomrule
\end{tabular}
\caption{\textbf{Gemma3 Result Summary by GPT-5.1}.
LLM-as-a-judge evaluation (\textbf{all categories}):
\textbf{average number of errors per output} on the five QUINTD-1 target domains
(\emph{lower is better}).
\textbf{Bold} indicates the best \emph{small model} per domain,
while \underline{underlined} values indicate the best overall system.
All error categories (Incorrect Fact, Not Checkable, Misleading, Other) are counted and summed.
Results are based on \textbf{GPT-5.1} as the judge.
}
\label{tab:llm_judge_all_errnum_5domains_gpt_gemma3}
\end{table*}

\begin{table*}[!htbp]
\centering
\small
\renewcommand{\arraystretch}{1.05}
\begin{tabular}{lccccc}
\toprule
\textbf{System} &
\textbf{Wikidata} &
\textbf{Ice Hockey} &
\textbf{OpenWeather} &
\textbf{GSM Arena} &
\textbf{OWID} \\
\midrule

\multicolumn{6}{l}{\textbf{\textit{Zero-shot (ZS)}}} \\
Qwen3-1.7B-ZS   & \lenstat{33.17}{18.76} & \lenstat{63.40}{12.40} & \lenstat{125.86}{21.35} & \lenstat{121.59}{19.23} & \lenstat{52.61}{12.91} \\
Qwen3-8B-ZS     & \lenstat{35.57}{18.33} & \lenstat{72.35}{8.22}  & \lenstat{120.54}{13.61} & \lenstat{140.67}{15.46} & \lenstat{52.32}{13.51} \\
Qwen3-32B-ZS    & \lenstat{34.79}{18.72} & \lenstat{75.44}{11.90} & \lenstat{127.22}{22.52} & \lenstat{183.45}{34.34} & \lenstat{71.73}{24.26} \\

\hdashline
\multicolumn{6}{l}{\textbf{\textit{WebNLG-SFT}}} \\
Qwen3-1.7B-SFT  & \lenstat{29.64}{17.50} & \lenstat{116.73}{304.27} & \lenstat{767.44}{749.55} & \lenstat{776.32}{916.36} & \lenstat{189.57}{370.39} \\
Qwen3-8B-SFT    & \lenstat{27.65}{16.25} & \lenstat{95.07}{20.59}   & \lenstat{291.76}{275.24} & \lenstat{611.21}{849.10} & \lenstat{129.65}{251.24} \\
Qwen3-32B-SFT   & \lenstat{28.68}{17.45} & \lenstat{39.57}{16.16}   & \lenstat{213.48}{254.23} & \lenstat{142.14}{261.65} & \lenstat{186.44}{260.91} \\

\hdashline
\multicolumn{6}{l}{\textbf{\textit{DDKD from ZS teacher}}} \\
Qwen3-1.7B-DDKD-Base  & \lenstat{33.41}{17.57} & \lenstat{77.25}{10.67} & \lenstat{115.48}{19.55} & \lenstat{192.79}{38.71} & \lenstat{76.71}{47.49} \\
Qwen3-1.7B-DDKD-Sub   & \lenstat{33.38}{16.18} & \lenstat{73.83}{12.11} & \lenstat{130.13}{23.73} & \lenstat{189.05}{29.27} & \lenstat{72.42}{27.50} \\
Qwen3-1.7B-DDKD-Pert  & \lenstat{34.40}{17.75} & \lenstat{75.97}{13.10} & \lenstat{124.48}{26.01} & \lenstat{181.21}{35.12} & \lenstat{78.99}{22.56} \\
Qwen3-1.7B-DDKD-Mixed & \lenstat{34.12}{17.64} & \lenstat{74.01}{10.55} & \lenstat{126.47}{19.16} & \lenstat{192.31}{38.72} & \lenstat{73.78}{26.00} \\

\hdashline
\multicolumn{6}{l}{\textbf{\textit{DDKD from WebNLG-SFT teacher}}} \\
Qwen3-1.7B-DDKD-Base  & \lenstat{28.22}{16.79} & \lenstat{35.29}{12.47} & \lenstat{235.06}{214.61} & \lenstat{770.70}{1036.64} & \lenstat{200.72}{218.60} \\
Qwen3-1.7B-DDKD-Sub   & \lenstat{28.59}{16.74} & \lenstat{40.98}{16.40} & \lenstat{173.27}{210.44} & \lenstat{492.42}{890.64}  & \lenstat{162.44}{242.18} \\
Qwen3-1.7B-DDKD-Pert  & \lenstat{29.35}{17.23} & \lenstat{43.32}{16.11} & \lenstat{217.15}{228.21} & \lenstat{458.81}{787.59}  & \lenstat{60.01}{122.09} \\
Qwen3-1.7B-DDKD-Mixed & \lenstat{29.55}{17.51} & \lenstat{37.82}{13.97} & \lenstat{88.17}{63.44}   & \lenstat{340.01}{701.74}  & \lenstat{140.66}{205.04} \\

\bottomrule
\end{tabular}
\caption{\textbf{Qwen3 Outputs Length Analysis}.
Output length statistics (\textbf{tokens}): mean $\pm$ standard deviation over 100 outputs per system on each QUINTD-1 target domain.
Tokenization uses the same tokenizer as the corresponding backbone (Qwen3) in our evaluation pipeline.
}
\label{tab:output_length_tokens_5domains}
\end{table*}

\clearpage

\begin{table*}[htbp!]
\centering
\small

\begin{tabular}{lcccccc}
\toprule
\textbf{Model} & \textbf{\# Real} & \textbf{Incorrect Fact} & \textbf{Not Checkable} & \textbf{Misleading} & \textbf{Other} & \textbf{All Categories} \\
\midrule

\multicolumn{7}{l}{\textbf{\textit{Zero-Shot (ZS)}}} \\
GPT-4.1-ZS        & -- & 0.00 & 0.12 & 0.00 & 0.00 & 0.12 \\
Qwen3-1.7B-ZS     & -- & 0.18 & 0.50 & 0.02 & 0.04 & 0.74 \\
Qwen3-8B-ZS       & -- & 0.01 & 0.23 & 0.00 & 0.03 & 0.27 \\
Qwen3-32B-ZS      & -- & 0.04 & 0.39 & 0.01 & 0.01 & 0.45 \\

\hdashline
\multicolumn{7}{l}{\textbf{\textit{WebNLG-SFT}}} \\
Qwen3-1.7B-SFT    & -- & 0.08 & 0.15 & 0.05 & 0.09 & 0.37 \\
Qwen3-8B-SFT      & -- & 0.02 & 0.12 & 0.02 & 0.01 & 0.17 \\
\textbf{Qwen3-32B-SFT}
                  & -- & 0.04 & 0.03 & 0.01 & 0.01 & \textbf{0.09} \\

\hdashline
\multicolumn{7}{l}{\textbf{\textit{DDKD from ZS teacher}}} \\
Qwen3-1.7B-DDKD-Base   & 100 & 0.11 & 0.37 & 0.04 & 0.01 & 0.53 \\
Qwen3-1.7B-DDKD-Sub    & 100 & 0.08 & 0.34 & 0.01 & 0.02 & 0.45 \\
Qwen3-1.7B-DDKD-Pert   & 100 & 0.03 & 0.26 & 0.00 & 0.00 & 0.29 \\
Qwen3-1.7B-DDKD-Mixed  & 100 & 0.04 & 0.27 & 0.01 & 0.01 & 0.33 \\
Qwen3-1.7B-DDKD-Base   & 500 & 0.08 & 0.45 & 0.01 & 0.07 & 0.61 \\

\hdashline
\multicolumn{7}{l}{\textbf{\textit{DDKD from WebNLG-SFT teacher}}} \\
Qwen3-1.7B-DDKD-Base   & 100 & 0.07 & 0.09 & 0.01 & 0.08 & 0.25 \\
Qwen3-1.7B-DDKD-Sub    & 100 & 0.01 & 0.11 & 0.03 & 0.06 & 0.21 \\
Qwen3-1.7B-DDKD-Pert   & 100 & 0.05 & 0.08 & 0.01 & 0.04 & 0.18 \\
Qwen3-1.7B-DDKD-Mixed  & 100 & 0.02 & 0.06 & 0.01 & 0.01 & 0.10 \\
Qwen3-1.7B-DDKD-Base   & 500 & 0.07 & 0.07 & 0.04 & 0.08 & 0.26 \\

\bottomrule
\end{tabular}

\caption{
\textbf{Error counts on Wikidata evaluated by GPT-5.1 (Qwen3).}
We report the average number of errors per output across different error categories (lower is better).
\textbf{\# Real} denotes the number of real seed instances used for distillation.
"All Categories" represents the total error count.
\textbf{Bold} indicates the best performing model (lowest total errors).
}

\label{tab:wikidata_error_counts}
\end{table*}

\begin{table*}[htbp!]
\centering
\small
\begin{tabular}{lcccccccc}
\toprule
\textbf{Model} & \textbf{\# Real} & \textbf{Incorrect Fact} & \textbf{Not Checkable} & \textbf{Misleading} & \textbf{Other} & \textbf{All Categories}\\
\midrule
\multicolumn{6}{l}{\textbf{\textit{Zero-Shot (ZS)}}} \\
GPT-4.1-ZS & -- & 0\% & 11\% & 0\% & 0\% & 11\% \\
Qwen3-1.7B-ZS & -- & 16\% & 31\% & 2\% & 4\% & 46\% \\
Qwen3-8B-ZS & -- & 1\% & 19\% & 0\% & 3\% & 22\% \\
Qwen3-32B-ZS & -- & 4\% & 29\% & 1\% & 1\% & 33\% \\

\hdashline
\multicolumn{6}{l}{\textbf{\textit{WebNLG-SFT}}} \\
Qwen3-1.7B-SFT & -- & 8\% & 14\% & 5\% & 9\% & 26\% \\
Qwen3-8B-SFT & -- & 2\% & 12\% & 2\% & 1\% & 15\% \\
\textbf{Qwen3-32B-SFT} & -- & 4\% & 3\% & 1\% & 1\% & \textbf{8\%} \\

\hdashline
\multicolumn{6}{l}{\textbf{\textit{DDKD from ZS teacher}}} \\
Qwen3-1.7B-DDKD-Base & 100 & 10\% & 26\% & 3\% & 1\% & 34\% \\
Qwen3-1.7B-DDKD-Sub & 100 & 8\% & 23\% & 1\% & 2\% & 31\% \\
Qwen3-1.7B-DDKD-Pert & 100 & 3\% & 25\% & 0\% & 0\% & 28\% \\
Qwen3-1.7B-DDKD-Mixed & 100 & 4\% & 23\% & 1\% & 1\% & 27\% \\
Qwen3-1.7B-DDKD-Base & 500 & 8\% & 33\% & 1\% & 7\% & 43\% \\

\hdashline
\multicolumn{6}{l}{\textbf{\textit{DDKD from WebNLG-SFT teacher}}} \\
Qwen3-1.7B-DDKD-Base & 100 & 6\% & 8\% & 1\% & 8\% & 21\% \\
Qwen3-1.7B-DDKD-Sub & 100 & 1\% & 11\% & 3\% & 6\% & 21\% \\
Qwen3-1.7B-DDKD-Pert & 100 & 5\% & 7\% & 1\% & 4\% & 17\% \\
Qwen3-1.7B-DDKD-Mixed & 100 & 2\% & 6\% & 1\% & 1\% & 10\% \\
Qwen3-1.7B-DDKD-Base & 500 & 6\% & 6\% & 4\% & 8\% & 19\% \\

\bottomrule
\end{tabular}
\caption{\textbf{Error rates on Wikidata evaluated by GPT-5.1 (Qwen3).} Values represent the percentage of generated outputs containing at least one error of the specific type. \textbf{\# Real} denotes the number of real seed instances used for distillation. The "All Categories" column reports the percentage of outputs containing \textbf{any} type of error. Lower percentages indicate better performance.}
\end{table*}

\clearpage
\begin{table*}[htbp!]
\centering
\small

\begin{tabular}{lcccccc}
\toprule
\textbf{Model} & \textbf{Incorrect Fact} & \textbf{Not Checkable} & \textbf{Misleading} & \textbf{Other} & \textbf{All Categories}\\
\midrule
\multicolumn{6}{l}{\textbf{\textit{Zero-Shot (ZS)}}} \\
Qwen3-1.7B-ZS & 0.30 & 0.28 & 0.06 & 0.34 & 0.98 \\
Qwen3-8B-ZS & 0.05 & 0.20 & 0.08 & 0.26 & 0.59 \\
Qwen3-32B-ZS & 0.11 & 0.28 & 0.04 & 0.10 & 0.53 \\

\hdashline
\multicolumn{6}{l}{\textbf{\textit{WebNLG-SFT}}} \\
Qwen3-1.7B-SFT & 0.19 & 0.08 & 0.20 & 0.23 & 0.70 \\
Qwen3-8B-SFT & 0.17 & 0.06 & 0.15 & 0.08 & 0.46 \\
Qwen3-32B-SFT & 0.17 & 0.03 & 0.14 & 0.04 & 0.38 \\

\hdashline
\multicolumn{6}{l}{\textbf{\textit{DDKD from ZS teacher}}} \\
Qwen3-1.7B-DDKD-Base & 0.12 & 0.24 & 0.14 & 0.24 & 0.74 \\
Qwen3-1.7B-DDKD-Sub & 0.14 & 0.26 & 0.08 & 0.14 & 0.62 \\
Qwen3-1.7B-DDKD-Pert & 0.05 & 0.27 & 0.05 & 0.14 & 0.51 \\
Qwen3-1.7B-DDKD-Mixed & 0.14 & 0.19 & 0.03 & 0.24 & 0.60 \\

\hdashline
\multicolumn{6}{l}{\textbf{\textit{DDKD from WebNLG-SFT teacher}}} \\
Qwen3-1.7B-DDKD-Base & 0.11 & 0.07 & 0.09 & 0.21 & 0.48 \\
Qwen3-1.7B-DDKD-Sub & 0.15 & 0.02 & 0.17 & 0.16 & 0.50 \\
Qwen3-1.7B-DDKD-Pert & 0.08 & 0.06 & 0.14 & 0.12 & 0.40 \\
\textbf{Qwen3-1.7B-DDKD-Mixed} & 0.19 & 0.01 & 0.08 & 0.08 & \textbf{0.36} \\

\bottomrule
\end{tabular}
\caption{\textbf{Error counts on Wikidata evaluated by Gemini-2.5-Pro (Qwen3).} We report the average number of errors per output across different error categories (lower is better). "All Categories" represents the total error count. \textbf{Bold} indicates the best performing model (lowest total errors).}
\end{table*}

\begin{table*}[htbp!]
\centering
\small
\begin{tabular}{lcccccc}
\toprule
\textbf{Model} & \textbf{Incorrect Fact} & \textbf{Not Checkable} & \textbf{Misleading} & \textbf{Other} & \textbf{All Categories}\\
\midrule
\multicolumn{6}{l}{\textbf{\textit{Zero-Shot (ZS)}}} \\
Qwen-1.7B-ZS & 26\% & 20\% & 4\% & 29\% & 62\% \\
Qwen3-8B-ZS & 5\% & 18\% & 8\% & 20\% & 44\% \\
Qwen3-32B-ZS & 11\% & 24\% & 4\% & 9\% & 40\% \\

\hdashline
\multicolumn{6}{l}{\textbf{\textit{WebNLG-SFT}}} \\
Qwen3-1.7B-SFT & 18\% & 8\% & 14\% & 20\% & 44\% \\
Qwen3-8B-SFT & 15\% & 6\% & 11\% & 7\% & 35\% \\
Qwen3-32B-SFT & 16\% & 3\% & 6\% & 4\% & 27\% \\

\hdashline
\multicolumn{6}{l}{\textbf{\textit{DDKD from ZS teacher}}} \\
Qwen3-1.7B-DDKD-Base & 12\% & 20\% & 9\% & 21\% & 48\% \\
Qwen3-1.7B-DDKD-Sub & 14\% & 22\% & 8\% & 12\% & 46\% \\
Qwen3-1.7B-DDKD-Pert & 5\% & 24\% & 5\% & 14\% & 40\% \\
Qwen3-1.7B-DDKD-Mixed & 13\% & 17\% & 3\% & 21\% & 46\% \\

\hdashline
\multicolumn{6}{l}{\textbf{\textit{DDKD from WebNLG-SFT teacher}}} \\
Qwen3-1.7B-DDKD-Base & 10\% & 6\% & 9\% & 18\% & 35\% \\
Qwen3-1.7B-DDKD-Sub & 13\% & 2\% & 10\% & 12\% & 35\% \\
Qwen3-1.7B-DDKD-Pert & 8\% & 6\% & 7\% & 11\% & 25\% \\
Qwen3-1.7B-DDKD-Mixed & 13\% & 1\% & 6\% & 6\% & 24\% \\

\bottomrule
\end{tabular}
\caption{\textbf{Error rates on Wikidata evaluated by Gemini-2.5-Pro (Qwen3).} Values represent the percentage of generated outputs containing at least one error of the specific type. The "All Categories" column reports the percentage of outputs containing \textbf{any} type of error. Lower percentages indicate better performance.}
\end{table*}

\clearpage
\begin{table*}[htbp!]
\centering
\small
\begin{tabular}{lcccccc}
\toprule
\textbf{Model} & \textbf{Incorrect Fact} & \textbf{Not Checkable} & \textbf{Misleading} & \textbf{Other} & \textbf{All Categories}\\
\midrule
\multicolumn{6}{l}{\textbf{\textit{Zero-Shot (ZS)}}} \\
Gemma3-1B-IT-ZS & 0.17 & 1.28 & 0.06 & 0.20 & 1.71 \\
Gemma3-27B-IT-ZS & 0.04 & 0.48 & 0.00 & 0.04 & 0.56 \\

\hdashline
\multicolumn{6}{l}{\textbf{\textit{WebNLG-SFT}}} \\
Gemma3-1B-IT-SFT & 2.65 & 1.69 & 0.16 & 0.98 & 5.49 \\
Gemma3-27B-IT-SFT & 0.04 & 0.06 & 0.02 & 0.00 & 0.12 \\

\hdashline
\multicolumn{6}{l}{\textbf{\textit{The best DDKD Model}}} \\
Gemma3-1B-IT-DDKD-Best & 0.09 & 0.11 & 0.03 & 0.05 & 0.28 \\

\bottomrule
\end{tabular}
\caption{\textbf{Error counts on Wikidata evaluated by GPT-5.1 (Gemma3).} We report the average number of errors per output across different error categories (lower is better). "All Categories" represents the total error count.}
\end{table*}

\begin{table*}[htbp!]
\centering
\small
\begin{tabular}{lcccccc}
\toprule
\textbf{Model} & \textbf{Incorrect Fact} & \textbf{Not Checkable} & \textbf{Misleading} & \textbf{Other} & \textbf{All Categories}\\
\midrule
\multicolumn{6}{l}{\textbf{\textit{Zero-Shot (ZS)}}} \\
Gemma3-1B-IT-ZS & 15\% & 67\% & 6\% & 18\% & 75\% \\
Gemma3-27B-IT-ZS & 4\% & 37\% & 0\% & 4\% & 39\% \\

\hdashline
\multicolumn{6}{l}{\textbf{\textit{WebNLG-SFT}}} \\
Gemma3-1B-IT-SFT & 80\% & 76\% & 14\% & 66\% & 99\% \\
Gemma3-27B-IT-SFT & 4\% & 6\% & 2\% & 0\% & 11\% \\

\hdashline
\multicolumn{6}{l}{\textbf{\textit{The best DDKD Model}}} \\
Gemma3-1B-IT-DDKD-Best & 8\% & 10\% & 3\% & 4\% & 21\% \\

\bottomrule
\end{tabular}
\caption{\textbf{Error rates on Wikidata evaluated by GPT-5.1 (Gemma3).} Values represent the percentage of generated outputs containing at least one error of the specific type. The "All Categories" column reports the percentage of outputs containing \textbf{any} type of error. Lower percentages indicate better performance.}
\end{table*}

\begin{table*}[htbp!]
\centering
\small
\begin{tabular}{lcccccc}
\toprule
\textbf{Model} & \textbf{Incorrect Fact} & \textbf{Not Checkable} & \textbf{Misleading} & \textbf{Other} & \textbf{All Categories}\\
\midrule
\multicolumn{6}{l}{\textbf{\textit{Zero-Shot (ZS)}}} \\
Gemma3-1B-IT-ZS & 0.38 & 0.96 & 0.16 & 0.53 & 2.03 \\
Gemma3-27B-IT-ZS & 0.05 & 0.38 & 0.04 & 0.70 & 1.17 \\

\hdashline
\multicolumn{6}{l}{\textbf{\textit{WebNLG-SFT}}} \\
Gemma3-1B-IT-SFT & 5.64 & 0.35 & 0.37 & 1.08 & 7.44 \\
Gemma3-27B-IT-SFT & 0.08 & 0.05 & 0.06 & 0.06 & 0.25 \\

\hdashline
\multicolumn{6}{l}{\textbf{\textit{The best DDKD Model}}} \\
Gemma3-1B-IT-DDKD-Best & 0.20 & 0.10 & 0.09 & 0.14 & 0.53 \\

\bottomrule
\end{tabular}
\caption{\textbf{Error counts on Wikidata evaluated by Gemini-2.5-Pro (Gemma3).} We report the average number of errors per output across different error categories (lower is better). "All Categories" represents the total error count.}
\end{table*}

\begin{table*}[htbp!]
\centering
\small
\begin{tabular}{lcccccc}
\toprule
\textbf{Model} & \textbf{Incorrect Fact} & \textbf{Not Checkable} & \textbf{Misleading} & \textbf{Other} & \textbf{All Categories}\\
\midrule
\multicolumn{6}{l}{\textbf{\textit{Zero-Shot (ZS)}}} \\
Gemma3-1B-IT-ZS & 34\% & 58\% & 15\% & 50\% & 93\% \\
Gemma3-27B-IT-ZS & 5\% & 34\% & 4\% & 63\% & 77\% \\

\hdashline
\multicolumn{6}{l}{\textbf{\textit{WebNLG-SFT}}} \\
Gemma3-1B-IT-SFT & 98\% & 14\% & 29\% & 84\% & 100\% \\
Gemma3-27B-IT-SFT & 7\% & 5\% & 6\% & 6\% & 22\% \\

\hdashline
\multicolumn{6}{l}{\textbf{\textit{The best DDKD Model}}} \\
Gemma3-1B-IT-DDKD-Best & 17\% & 10\% & 8\% & 13\% & 42\% \\

\bottomrule
\end{tabular}
\caption{\textbf{Error rates on Wikidata evaluated by Gemini-2.5-Pro (Gemma3).} Values represent the percentage of generated outputs containing at least one error of the specific type. The "All Categories" column reports the percentage of outputs containing \textbf{any} type of error. Lower percentages indicate better performance.}
\label{tab:error_rate_wikidata_gemini_gemma3}
\end{table*}

\clearpage

\begin{table*}[htbp!]
\centering
\small
\begin{tabular}{lccccccc}
\toprule
\textbf{Model} & \textbf{\# Real} & \textbf{Incorrect Fact} & \textbf{Not Checkable} & \textbf{Misleading} & \textbf{Other} & \textbf{All Categories}\\
\midrule
\multicolumn{6}{l}{\textbf{\textit{Zero-Shot (ZS)}}} \\
GPT-4.1-ZS & -- & 0.01 & 0.36 & 0.00 & 0.00 & 0.37 \\
Qwen3-1.7B-ZS & -- & 0.73 & 0.58 & 0.06 & 0.01 & 1.38 \\
Qwen3-8B-ZS  & -- & 0.30 & 0.27 & 0.02 & 0.00 & 0.59 \\
Qwen3-32B-ZS  & -- & 0.07 & 0.54 & 0.02 & 0.00 & 0.63 \\

\hdashline
\multicolumn{6}{l}{\textbf{\textit{WebNLG-SFT}}} \\
Qwen3-1.7B-SFT  & -- & 1.21 & 0.13 & 0.05 & 0.10 & 1.49 \\
Qwen3-8B-SFT  & -- & 0.06 & 0.09 & 0.00 & 0.01 & 0.16 \\
\textbf{Qwen3-32B-SFT}  & -- & 0.01 & 0.01 & 0.00 & 0.02 & \textbf{0.02} \\

\hdashline
\multicolumn{6}{l}{\textbf{\textit{DDKD from ZS teacher}}} \\
Qwen3-1.7B-DDKD-Base  & 100 & 0.33 & 0.56 & 0.01 & 0.01 & 0.91 \\
Qwen3-1.7B-DDKD-Sub  & 100 & 0.07 & 0.58 & 0.01 & 0.00 & 0.66 \\
Qwen3-1.7B-DDKD-Pert  & 100 & 0.09 & 0.75 & 0.00 & 0.05 & 0.89 \\
Qwen3-1.7B-DDKD-Mixed  & 100 & 0.15 & 0.54 & 0.00 & 0.04 & 0.73 \\
Qwen3-1.7B-DDKD-Base  & 500 & 0.10 & 0.77 & 0.01 & 0.12  & 1.00 \\

\hdashline
\multicolumn{6}{l}{\textbf{\textit{DDKD from WebNLG-SFT teacher}}} \\
Qwen3-1.7B-DDKD-Base  & 100 & 0.26 & 0.00 & 0.00 & 0.00 & 0.26 \\
Qwen3-1.7B-DDKD-Sub  & 100 & 0.02 & 0.02 & 0.00 & 0.00 & 0.04 \\
Qwen3-1.7B-DDKD-Pert  & 100 & 0.10 & 0.02 & 0.00 & 0.00 & 0.12 \\
Qwen3-1.7B-DDKD-Mixed  & 100 & 0.01 & 0.11 & 0.00 & 0.02 & 0.03 \\
Qwen3-1.7B-DDKD-Base  & 500 & 0.04  & 0.01 & 0.00 & 0.00 & 0.04 \\

\bottomrule
\end{tabular}
\caption{\textbf{Error counts on Ice Hockey evaluated by GPT-5.1 (Qwen3).} We report the average number of errors per output across different error categories (lower is better). \textbf{\# Real} denotes the number of real seed instances used for distillation. "All Categories" represents the total error count. \textbf{Bold} indicates the best performing model (lowest total errors).}
\label{tab:error_count_icehockey_gpt_qwen3}
\end{table*}

\begin{table*}[htbp!]
\centering
\small
\begin{tabular}{lcccccc}
\toprule
\textbf{Model} & \textbf{\# Real} & \textbf{Incorrect Fact} & \textbf{Not Checkable} & \textbf{Misleading} & \textbf{Other} & \textbf{All Categories}\\
\midrule
\multicolumn{6}{l}{\textbf{\textit{Zero-Shot (ZS)}}} \\
GPT-4.1-ZS & -- & 1\% & 25\% & 0\% & 0\% & 26\% \\
Qwen3-1.7B-ZS & -- & 44\% & 47\% & 6\% & 1\% & 73\% \\
Qwen3-8B-ZS & -- & 24\% & 20\% & 2\% & 0\% & 40\% \\
Qwen3-32B-ZS & -- & 5\% & 45\% & 2\% & 0\% & 47\% \\

\hdashline
\multicolumn{6}{l}{\textbf{\textit{WebNLG-SFT}}} \\
Qwen3-1.7B-SFT & -- & 63\% & 11\% & 5\% & 10\% & 75\% \\
Qwen3-8B-SFT & -- & 6\% & 9\% & 0\% & 1\% & 14\% \\
\textbf{Qwen3-32B-SFT} & -- & 1\% & 1\% & 0\% & 0\% & \textbf{2\%} \\

\hdashline
\multicolumn{6}{l}{\textbf{\textit{DDKD from ZS teacher}}} \\
Qwen3-1.7B-DDKD-Base & 100 & 23\% & 47\% & 1\% & 1\% & 55\% \\
Qwen3-1.7B-DDKD-Sub & 100 & 7\% & 50\% & 1\% & 0\% & 54\% \\
Qwen3-1.7B-DDKD-Pert & 100 & 8\% & 55\% & 0\% & 4\% & 60\% \\
Qwen3-1.7B-DDKD-Mixed & 100 & 8\% & 39\% & 0\% & 4\% & 46\% \\
Qwen3-1.7B-DDKD-Base & 500 & 9\% & 53\% & 1\% & 12\% & 60\% \\

\hdashline
\multicolumn{6}{l}{\textbf{\textit{DDKD from WebNLG-SFT teacher}}} \\
Qwen3-1.7B-DDKD-Base & 100 & 21\% & 0\% & 0\% & 0\% & 21\% \\
Qwen3-1.7B-DDKD-Sub & 100 & 2\% & 2\% & 0\% & 0\% & 4\% \\
Qwen3-1.7B-DDKD-Pert & 100 & 7\% & 2\% & 0\% & 1\% & 9\% \\
\textbf{Qwen3-1.7B-DDKD-Mixed} & 100 & 1\% & 0\% & 0\% & 1\% & \textbf{2\%} \\
Qwen3-1.7B-DDKD-Base & 500 & 3\% & 1\% & 0\% & 0\% & 4\% \\

\bottomrule
\end{tabular}
\caption{\textbf{Error rates on Ice Hockey evaluated by GPT-5.1 (Qwen3).} Values represent the percentage of generated outputs containing at least one error of the specific type. \textbf{\# Real} denotes the number of real seed instances used for distillation. The "All Categories" column reports the percentage of outputs containing \textbf{any} type of error. Lower percentages indicate better performance.}
\end{table*}

\clearpage

\begin{table*}[htbp!]
\centering
\small
\begin{tabular}{lcccccc}
\toprule
\textbf{Model} & \textbf{Incorrect Fact} & \textbf{Not Checkable} & \textbf{Misleading} & \textbf{Other} & \textbf{All Categories}\\
\midrule
\multicolumn{6}{l}{\textbf{\textit{Zero-Shot (ZS)}}} \\
Qwen3-1.7B-ZS & 0.88 & 0.27 & 0.19 & 0.37 & 1.71 \\
Qwen3-8B-ZS  & 0.39 & 0.07 & 0.06 & 0.10 & 0.62 \\
Qwen3-32B-ZS  & 0.10 & 0.19 & 0.05 & 0.15 & 0.49 \\

\hdashline
\multicolumn{6}{l}{\textbf{\textit{WebNLG-SFT}}} \\
Qwen3-1.7B-SFT  & 1.59 & 0.12 & 0.27 & 0.44 & 2.42 \\
Qwen3-8B-SFT  & 0.13 & 0.05 & 0.09 & 0.11 & 0.38 \\
Qwen3-32B-SFT  & 0.04 & 0.00 & 0.07 & 0.01 & 0.12 \\

\hdashline
\multicolumn{6}{l}{\textbf{\textit{DDKD from ZS teacher}}} \\
Qwen3-1.7B-DDKD-Base  & 0.38 & 0.22 & 0.11 & 0.15 & 0.86 \\
Qwen3-1.7B-DDKD-Sub  & 0.09 & 0.21 & 0.13 & 0.15 & 0.58 \\
Qwen3-1.7B-DDKD-Pert  & 0.13 & 0.37 & 0.05 & 0.17 & 0.72 \\
Qwen3-1.7B-DDKD-Mixed  & 0.18 & 0.19 & 0.07 & 0.11 & 0.55 \\

\hdashline
\multicolumn{6}{l}{\textbf{\textit{DDKD from WebNLG-SFT teacher}}} \\
Qwen3-1.7B-DDKD-Base  & 0.25 & 0.02 & 0.11 & 0.00 & 0.38 \\
\textbf{Qwen3-1.7B-DDKD-Sub}  & 0.08 & 0.00 & 0.03 & 0.00 & \textbf{0.11} \\
Qwen3-1.7B-DDKD-Pert  & 0.13 & 0.01 & 0.04 & 0.01 & 0.19 \\
Qwen3-1.7B-DDKD-Mixed  & 0.03 & 0.01 & 0.09 & 0.00 & 0.13 \\

\bottomrule
\end{tabular}
\caption{\textbf{Error counts on Ice Hockey evaluated by Gemini-2.5-Pro (Qwen3).} We report the average number of errors per output across different error categories (lower is better). "All Categories" represents the total error count. \textbf{Bold} indicates the best performing model (lowest total errors).}
\end{table*}

\begin{table*}[htbp!]
\centering
\small
\begin{tabular}{lcccccc}
\toprule
\textbf{Model} & \textbf{Incorrect Fact} & \textbf{Not Checkable} & \textbf{Misleading} & \textbf{Other} & \textbf{All Categories}\\
\midrule
\multicolumn{6}{l}{\textbf{\textit{Zero-Shot (ZS)}}} \\
Qwen3-1.7B-ZS & 53\% & 25\% & 17\% & 35\% & 91\% \\
Qwen3-8B-ZS & 33\% & 7\% & 4\% & 10\% & 46\% \\
Qwen3-32B-ZS & 8\% & 17\% & 5\% & 12\% & 33\% \\

\hdashline
\multicolumn{6}{l}{\textbf{\textit{WebNLG-SFT}}} \\
Qwen3-1.7B-SFT & 79\% & 11\% & 26\% & 39\% & 97\% \\
Qwen3-8B-SFT & 13\% & 5\% & 7\% & 8\% & 32\% \\
Qwen3-32B-SFT & 3\% & 0\% & 7\% & 1\% & 11\% \\

\hdashline
\multicolumn{6}{l}{\textbf{\textit{DDKD from ZS teacher}}} \\
Qwen3-1.7B-DDKD-Base & 27\% & 20\% & 11\% & 13\% & 51\% \\
Qwen3-1.7B-DDKD-Sub & 9\% & 17\% & 13\% & 15\% & 42\% \\
Qwen3-1.7B-DDKD-Pert & 10\% & 33\% & 5\% & 15\% & 46\% \\
Qwen3-1.7B-DDKD-Mixed & 10\% & 15\% & 6\% & 10\% & 36\% \\

\hdashline
\multicolumn{6}{l}{\textbf{\textit{DDKD from WebNLG-SFT teacher}}} \\
Qwen3-1.7B-DDKD-Base & 21\% & 2\% & 11\% & 0\% & 34\% \\
\textbf{Qwen3-1.7B-DDKD-Sub} & 7\% & 0\% & 3\% & 0\% & \textbf{10\%} \\
Qwen3-1.7B-DDKD-Pert & 10\% & 1\% & 4\% & 1\% & 16\% \\
Qwen3-1.7B-DDKD-Mixed & 2\% & 1\% & 8\% & 0\% & 11\% \\

\bottomrule
\end{tabular}
\caption{\textbf{Error rates on Ice Hockey evaluated by Gemini-2.5-Pro (Qwen3).} Values represent the percentage of generated outputs containing at least one error of the specific type. The "All Categories" column reports the percentage of outputs containing \textbf{any} type of error. Lower percentages indicate better performance.}
\end{table*}

\clearpage
\begin{table*}[htbp!]
\centering
\small
\begin{tabular}{lcccccc}
\toprule
\textbf{Model} & \textbf{Incorrect Fact} & \textbf{Not Checkable} & \textbf{Misleading} & \textbf{Other} & \textbf{All Categories}\\
\midrule
\multicolumn{6}{l}{\textbf{\textit{Zero-Shot (ZS)}}} \\
Gemma3-1B-IT-ZS & 2.33 & 0.34 & 0.01 & 0.12 & 2.80 \\
Gemma3-27B-IT-ZS & 0.13 & 0.14 & 0.01 & 0.00 & 0.28\\

\hdashline
\multicolumn{6}{l}{\textbf{\textit{WebNLG-SFT}}} \\
Gemma3-1B-IT-SFT & 2.69 & 1.14 & 0.31 & 1.20 & 5.34 \\
Gemma3-27B-IT-SFT & 0.11 & 0.09 & 0.00  & 0.00 & 0.20 \\

\hdashline
\multicolumn{6}{l}{\textbf{\textit{The best DDKD Model}}} \\
Gemma3-1B-IT-DDKD-Best & 0.35 & 0.03 & 0.00 & 0.00 & 0.38 \\

\bottomrule
\end{tabular}
\label{tab:avg_num_err_ice_gemma3_gpt}
\caption{\textbf{Error counts on Ice Hockey evaluated by GPT-5.1 (Gemma3).} We report the average number of errors per output across different error categories (lower is better). "All Categories" represents the total error count.}
\end{table*}

\begin{table*}[htbp!]
\centering
\small
\begin{tabular}{lcccccc}
\toprule
\textbf{Model} & \textbf{Incorrect Fact} & \textbf{Not Checkable} & \textbf{Misleading} & \textbf{Other} & \textbf{All Categories}\\
\midrule
\multicolumn{6}{l}{\textbf{\textit{Zero-Shot (ZS)}}} \\
Gemma3-1B-IT-ZS & 92\% & 26\% & 1\% & 12\% & 94\% \\
Gemma3-27B-IT-ZS & 12\% & 13\% & 1\% & 0\% & 24\% \\

\hdashline
\multicolumn{6}{l}{\textbf{\textit{WebNLG-SFT}}} \\
Gemma3-1B-IT-SFT & 88\% & 62\% & 24\% & 78\% & 100\% \\
Gemma3-27B-IT-SFT & 10\% & 9\% & 0\% & 0\% & 19\% \\

\hdashline
\multicolumn{6}{l}{\textbf{\textit{The best DDKD Model}}} \\
Gemma3-1B-IT-DDKD-Best & 30\% & 3\% & 0\% & 0\% & 32\% \\

\bottomrule
\end{tabular}
\caption{\textbf{Error rates on Ice Hockey evaluated by GPT-5.1 (Gemma3).} Values represent the percentage of generated outputs containing at least one error of the specific type. The "All Categories" column reports the percentage of outputs containing \textbf{any} type of error. Lower percentages indicate better performance.}
\end{table*}

\begin{table*}[htbp!]
\centering
\small
\label{tab:avg_num_err_ice_gemma3_gemini}
\begin{tabular}{lcccccc}
\toprule
\textbf{Model} & \textbf{Incorrect Fact} & \textbf{Not Checkable} & \textbf{Misleading} & \textbf{Other} & \textbf{All Categories}\\
\midrule
\multicolumn{6}{l}{\textbf{\textit{Zero-Shot (ZS)}}} \\
Gemma3-1B-IT-ZS & 2.46 & 0.13 & 0.17 & 0.33 & 3.09 \\
Gemma3-27B-IT-ZS & 0.21 & 0.10 & 0.05 & 0.19 & 0.55 \\

\hdashline
\multicolumn{6}{l}{\textbf{\textit{WebNLG-SFT}}} \\
Gemma3-1B-IT-SFT & 3.09 & 0.51 & 0.33 & 1.96 & 5.89 \\
Gemma3-27B-IT-SFT & 0.17 & 0.05 & 0.06 & 0.32 & 0.60 \\

\hdashline
\multicolumn{6}{l}{\textbf{\textit{The best DDKD Model}}} \\
Gemma3-1B-IT-DDKD-Best & 0.40 & 0.03 & 0.13 & 0.31 & 0.87 \\

\bottomrule
\end{tabular}
\caption{\textbf{Error counts on Ice Hockey evaluated by Gemini-2.5-Pro (Gemma3).}}
\end{table*}

\begin{table*}[htbp!]
\centering
\small
\begin{tabular}{lcccccc}
\toprule
\textbf{Model} & \textbf{Incorrect Fact} & \textbf{Not Checkable} & \textbf{Misleading} & \textbf{Other} & \textbf{All Categories}\\
\midrule
\multicolumn{6}{l}{\textbf{\textit{Zero-Shot (ZS)}}} \\
Gemma3-1B-IT-ZS & 92\% & 10\% & 14\% & 31\% & 98\% \\
Gemma3-27B-IT-ZS & 17\% & 10\% & 5\% & 15\% & 40\% \\

\hdashline
\multicolumn{6}{l}{\textbf{\textit{WebNLG-SFT}}} \\
Gemma3-1B-IT-SFT & 93\% & 32\% & 26\% & 96\% & 99\% \\
Gemma3-27B-IT-SFT & 16\% & 5\% & 6\% & 29\% & 50\% \\

\hdashline
\multicolumn{6}{l}{\textbf{\textit{The best DDKD Model}}} \\
Gemma3-1B-IT-DDKD-Best & 34\% & 3\% & 12\% & 24\% & 56\% \\

\bottomrule
\end{tabular}
\caption{\textbf{Error rates on Ice Hockey evaluated by Gemini-2.5-Pro (Gemma3).}}
\label{tab:error_rate_icehockey_gemini_gemma3}
\end{table*}


\clearpage

\begin{table*}[htbp!]
\centering
\small
\begin{tabular}{lccccccc}
\toprule
\textbf{Model} & \textbf{\# Real} & \textbf{Incorrect Fact} & \textbf{Not Checkable} & \textbf{Misleading} & \textbf{Other} & \textbf{All Categories}\\
\midrule
\multicolumn{6}{l}{\textbf{\textit{Zero-Shot (ZS)}}} \\
GPT-4.1-ZS & -- & 0.34 & 0.32 & 0.10 & 0.00 & 0.76 \\
Qwen3-1.7B-ZS & -- & 5.06 & 0.50 & 0.52 & 0.02 & 6.10 \\
Qwen3-8B-ZS & -- & 2.52 & 0.34 & 0.46 & 0.05 & 3.37 \\
Qwen3-32B-ZS & -- & 1.70 & 0.35 & 0.31 & 0.17 & 2.53 \\

\hdashline
\multicolumn{6}{l}{\textbf{\textit{WebNLG-SFT}}} \\
Qwen3-1.7B-SFT  & -- & 13.49 & 2.42 & 0.20 & 0.55 & 16.66 \\
Qwen3-8B-SFT  & -- & 4.78 & 0.31 & 0.08 & 0.04 & 5.21 \\
\textbf{Qwen3-32B-SFT}  & -- & 1.52 & 0.11 & 0.16 & 0.07 & \textbf{1.86} \\

\hdashline
\multicolumn{6}{l}{\textbf{\textit{DDKD from ZS teacher}}} \\
Qwen3-1.7B-DDKD-Base  & 100 & 3.45 & 0.19 & 0.36 & 0.05 & 4.05 \\
Qwen3-1.7B-DDKD-Sub  & 100 & 2.22 & 0.10 & 0.20 & 0.16 & 2.68 \\
Qwen3-1.7B-DDKD-Pert  & 100 & 2.04 & 0.19 & 0.33 & 0.12 & 2.68 \\
Qwen3-1.7B-DDKD-Mixed  & 100 & 2.15 & 0.22 & 0.34 & 0.17 & 2.88 \\
Qwen3-1.7B-DDKD-Base  & 500 & 3.38 & 0.22 & 0.43 & 0.05 & 4.08 \\

\hdashline
\multicolumn{6}{l}{\textbf{\textit{DDKD from WebNLG-SFT teacher}}} \\
Qwen3-1.7B-DDKD-Base  & 100 & 5.74 & 0.66 & 0.32 & 0.06 & 6.78 \\
Qwen3-1.7B-DDKD-Sub  & 100 & 1.55 & 0.28 & 0.16 & 0.02 & 2.01 \\
Qwen3-1.7B-DDKD-Pert   & 100 & 3.44 & 0.04 & 0.97 & 0.43 & 4.88 \\
Qwen3-1.7B-DDKD-Mixed  & 100 & 3.99 & 0.04 & 0.39 & 0.05 & 4.47 \\
Qwen3-1.7B-DDKD-Base  & 500 & 2.14 & 0.28 & 0.27 & 0.10 & 2.79 \\

\bottomrule
\end{tabular}
\caption{\textbf{Error counts on OpenWeather evaluated by GPT-5.1 (Qwen3).} We report the average number of errors per output across different error categories (lower is better). \textbf{\# Real} denotes the number of real seed instances used for distillation. "All Categories" represents the total error count. \textbf{Bold} indicates the best performing open-source model (lowest total errors).}
\label{tab:error_count_weather_gpt_qwen3}
\end{table*}

\begin{table*}[htbp!]
\centering
\small
\begin{tabular}{lccccccc}
\toprule
\textbf{Model} & \textbf{\# Real} & \textbf{Incorrect Fact} & \textbf{Not Checkable} & \textbf{Misleading} & \textbf{Other} & \textbf{All Categories}\\
\midrule
\multicolumn{6}{l}{\textbf{\textit{Zero-Shot (ZS)}}} \\
GPT-4.1-ZS & -- & 29\% & 23\% & 10\% & 5\% & 49\% \\
Qwen3-1.7B-ZS & -- & 98\% & 23\% & 36\% & 2\% & 99\% \\
Qwen3-8B-ZS & -- & 94\% & 26\% & 37\% & 5\% & 98\% \\
Qwen3-32B-ZS & -- & 83\% & 20\% & 25\% & 17\% & 92\% \\

\hdashline
\multicolumn{6}{l}{\textbf{\textit{WebNLG-SFT}}} \\
Qwen3-1.7B-SFT & -- & 93\% & 29\% & 16\% & 43\% & 96\% \\
Qwen3-8B-SFT & -- & 67\% & 10\% & 6\% & 4\% & 72\% \\
\textbf{Qwen3-32B-SFT} & -- & 53\% & 10\% & 14\% & 7\% & \textbf{69\%} \\

\hdashline
\multicolumn{6}{l}{\textbf{\textit{DDKD from ZS teacher}}} \\
Qwen3-1.7B-DDKD-Base & 100 & 97\% & 13\% & 29\% & 5\% & 99\% \\
Qwen3-1.7B-DDKD-Sub & 100 & 89\% & 9\% & 16\% & 16\% & 92\% \\
Qwen3-1.7B-DDKD-Pert & 100 & 85\% & 16\% & 27\% & 12\% & 91\% \\
Qwen3-1.7B-DDKD-Mixed & 100 & 81\% & 16\% & 29\% & 16\% & 90\% \\
Qwen3-1.7B-DDKD-Base & 500 & 95\% & 18\% & 40\% & 5\% & 97\% \\

\hdashline
\multicolumn{6}{l}{\textbf{\textit{DDKD from WebNLG-SFT teacher}}} \\
Qwen3-1.7B-DDKD-Base & 100 & 81\% & 11\% & 30\% & 6\% & 92\% \\
Qwen3-1.7B-DDKD-Sub & 100 & 61\% & 11\% & 15\% & 2\% & 70\% \\
Qwen3-1.7B-DDKD-Pert & 100 & 62\% & 13\% & 5\% & 7\% & 70\% \\
\textbf{Qwen3-1.7B-DDKD-Mixed} & 100 & 63\% & 2\% & 16\% & 1\% & \textbf{69\%} \\
Qwen3-1.7B-DDKD-Base & 500 & 59\% & 21\% & 23\% & 6\% & 76\% \\

\bottomrule
\end{tabular}
\caption{\textbf{Error rates on OpenWeather evaluated by GPT-5.1 (Qwen3).} Values represent the percentage of generated outputs containing at least one error of the specific type. \textbf{\# Real} denotes the number of real seed instances used for distillation. The "All Categories" column reports the percentage of outputs containing \textbf{any} type of error. Lower percentages indicate better performance.}
\end{table*}

\clearpage

\begin{table*}[htbp!]
\centering
\small
\begin{tabular}{lcccccc}
\toprule
\textbf{Model} & \textbf{Incorrect Fact} & \textbf{Not Checkable} & \textbf{Misleading} & \textbf{Other} & \textbf{All Categories}\\
\midrule
\multicolumn{6}{l}{\textbf{\textit{Zero-Shot (ZS)}}} \\
Qwen3-1.7B-ZS & 6.30 & 0.10 & 0.74 & 0.05 & 7.19 \\
Qwen3-8B-ZS  & 3.61 & 0.06 & 0.52 & 0.03 & 4.22 \\
Qwen3-32B-ZS  & 2.84 & 0.07 & 0.75 & 0.03 & 3.69 \\

\hdashline
\multicolumn{6}{l}{\textbf{\textit{WebNLG-SFT}}} \\
Qwen3-1.7B-SFT  & 9.54 & 0.90 & 0.61 & 0.59 & 10.83 \\
Qwen3-8B-SFT  & 4.28 & 0.00 & 0.46 & 0.22 & 4.96 \\
\textbf{Qwen3-32B-SFT}  & 1.46 & 0.01 & 0.89 & 0.04 & \textbf{2.40} \\

\hdashline
\multicolumn{6}{l}{\textbf{\textit{DDKD from ZS teacher}}} \\
Qwen3-1.7B-DDKD-Base  & 4.48 & 0.03 & 0.61 & 0.01 & 5.13 \\
Qwen3-1.7B-DDKD-Sub  & 3.36 & 0.05 & 0.78 & 0.09 & 4.28 \\
Qwen3-1.7B-DDKD-Pert  & 3.20 & 0.06 & 0.85 & 0.02 & 4.13 \\
Qwen3-1.7B-DDKD-Mixed  & 3.48 & 0.04 & 0.77 & 0.05 & 4.34 \\

\hdashline
\multicolumn{6}{l}{\textbf{\textit{DDKD from WebNLG-SFT teacher}}} \\
Qwen3-1.7B-DDKD-Base  & 7.58 & 0.00 & 0.85 & 0.20 & 8.63 \\
Qwen3-1.7B-DDKD-Sub  & 2.67 & 0.10 & 1.54 & 0.11 & 4.33 \\
Qwen3-1.7B-DDKD-Pert  & 3.42 & 0.04 & 1.04 & 0.36 & 4.86 \\
Qwen3-1.7B-DDKD-Mixed  & 4.05 & 0.02 & 0.52 & 0.06 & 4.65 \\

\bottomrule
\end{tabular}
\caption{\textbf{Error counts on OpenWeather evaluated by Gemini-2.5-Pro (Qwen3).} We report the average number of errors per output across different error categories (lower is better). "All Categories" represents the total error count. \textbf{Bold} indicates the best performing model (lowest total errors).}
\end{table*}

\begin{table*}[htbp!]
\centering
\small
\begin{tabular}{lcccccc}
\toprule
\textbf{Model} & \textbf{Incorrect Fact} & \textbf{Not Checkable} & \textbf{Misleading} & \textbf{Other} & \textbf{All Categories}\\
\midrule
\multicolumn{6}{l}{\textbf{\textit{Zero-Shot (ZS)}}} \\
Qwen3-1.7B-ZS & 100\% & 8\% & 42\% & 5\% & 100\% \\
Qwen3-8B-ZS & 100\% & 6\% & 42\% & 3\% & 100\% \\
Qwen3-32B-ZS & 96\% & 6\% & 47\% & 3\% & 97\% \\

\hdashline
\multicolumn{6}{l}{\textbf{\textit{WebNLG-SFT}}} \\
Qwen3-1.7B-SFT & 95\% & 7\% & 31\% & 46\% & 99\% \\
Qwen3-8B-SFT & 78\% & 0\% & 24\% & 17\% & 90\% \\
\textbf{Qwen3-32B-SFT} & 66\% & 1\% & 38\% & 4\% & \textbf{81\%} \\

\hdashline
\multicolumn{6}{l}{\textbf{\textit{DDKD from ZS teacher}}} \\
Qwen3-1.7B-DDKD-Base & 99\% & 3\% & 44\% & 1\% & 100\% \\
Qwen3-1.7B-DDKD-Sub  & 98\% & 5\% & 49\% & 6\% & 99\% \\
Qwen3-1.7B-DDKD-Pert & 96\% & 6\% & 49\% & 2\% & 99\% \\
Qwen3-1.7B-DDKD-Mixed & 99\% & 4\% & 54\% & 3\% & 100\% \\

\hdashline
\multicolumn{6}{l}{\textbf{\textit{DDKD from WebNLG-SFT teacher}}} \\
Qwen3-1.7B-DDKD-Base & 97\% & 0\% & 40\% & 13\% & 99\% \\
Qwen3-1.7B-DDKD-Sub & 80\% & 1\% & 50\% & 11\% & 96\% \\
Qwen3-1.7B-DDKD-Pert & 80\% & 3\% & 35\% & 30\% & 94\% \\
Qwen3-1.7B-DDKD-Mixed & 90\% & 2\% & 29\% & 6\% & 95\% \\

\bottomrule
\end{tabular}
\caption{\textbf{Error rates on OpenWeather evaluated by Gemini-2.5-Pro (Qwen3).} Values represent the percentage of generated outputs containing at least one error of the specific type. The "All Categories" column reports the percentage of outputs containing \textbf{any} type of error. Lower percentages indicate better performance.}
\end{table*}

\clearpage
\begin{table*}[htbp!]
\centering
\small
\begin{tabular}{lcccccc}
\toprule
\textbf{Model} & \textbf{Incorrect Fact} & \textbf{Not Checkable} & \textbf{Misleading} & \textbf{Other} & \textbf{All Categories}\\
\midrule
\multicolumn{6}{l}{\textbf{\textit{Zero-Shot (ZS)}}} \\
Gemma3-1B-IT-ZS & 3.02 & 1.05 & 0.57 & 0.02 & 4.66 \\
Gemma3-27B-IT-ZS & 3.57 & 1.00 & 0.34 & 0.07 & 4.98 \\

\hdashline
\multicolumn{6}{l}{\textbf{\textit{WebNLG-SFT}}} \\
Gemma3-1B-IT-SFT & 2.65 & 1.69 & 0.16 & 0.98 & 5.48 \\
Gemma3-27B-IT-SFT & 2.26 & 0.06 & 0.12 & 0.04 & 2.48 \\

\hdashline
\multicolumn{6}{l}{\textbf{\textit{The best DDKD Model}}} \\
Gemma3-1B-IT-DDKD-Best & 3.94 & 0.05 & 0.16 & 0.08 & 4.23 \\

\bottomrule
\end{tabular}
\caption{\textbf{Error counts on OpenWeather evaluated by GPT-5.1 (Gemma3).} We report the average number of errors per output across different error categories (lower is better). "All Categories" represents the total error count. \textbf{Bold} indicates the best performing model (lowest total errors).}
\end{table*}

\begin{table*}[htbp!]
\centering
\small
\begin{tabular}{lcccccc}
\toprule
\textbf{Model} & \textbf{Incorrect Fact} & \textbf{Not Checkable} & \textbf{Misleading} & \textbf{Other} & \textbf{All Categories}\\
\midrule
\multicolumn{6}{l}{\textbf{\textit{Zero-Shot (ZS)}}} \\
Gemma3-1B-IT-ZS & 94\% & 64\% & 45\% & 2\% & 100\% \\
Gemma3-27B-IT-ZS & 93\% & 27\% & 25\% & 7\% & 97\% \\

\hdashline
\multicolumn{6}{l}{\textbf{\textit{WebNLG-SFT}}} \\
Gemma3-1B-IT-SFT & 80\% & 76\% & 14\% & 66\% & 99\% \\
Gemma3-27B-IT-SFT & 31\% & 6\% & 4\% & 4\% & 43\% \\

\hdashline
\multicolumn{6}{l}{\textbf{\textit{The best DDKD Model}}} \\
Gemma3-1B-IT-DDKD-Best & 70\% & 4\% & 8\% & 7\% & 75\% \\

\bottomrule
\end{tabular}
\caption{\textbf{Error rates on OpenWeather evaluated by GPT-5.1 (Gemma3).} Values represent the percentage of generated outputs containing at least one error of the specific type. The "All Categories" column reports the percentage of outputs containing \textbf{any} type of error. Lower percentages indicate better performance.}
\end{table*}

\begin{table*}[htbp!]
\centering
\small
\begin{tabular}{lcccccc}
\toprule
\textbf{Model} & \textbf{Incorrect Fact} & \textbf{Not Checkable} & \textbf{Misleading} & \textbf{Other} & \textbf{All Categories}\\
\midrule
\multicolumn{6}{l}{\textbf{\textit{Zero-Shot (ZS)}}} \\
Gemma3-1B-IT-ZS & 4.06 & 0.65 & 0.57 & 0.09 & 5.37 \\
Gemma3-27B-IT-ZS & 5.72 & 0.10 & 0.69 & 0.10 & 6.52 \\

\hdashline
\multicolumn{6}{l}{\textbf{\textit{WebNLG-SFT}}} \\
Gemma3-1B-IT-SFT & 5.64 & 0.35 & 0.37 & 1.08 & 7.44 \\
Gemma3-27B-IT-SFT & 2.86 & 0.05 & 1.61 & 0.33 & 4.85 \\

\hdashline
\multicolumn{6}{l}{\textbf{\textit{The best DDKD Model}}} \\
Gemma3-1B-IT-DDKD-Best & 3.55 & 0.03 & 1.99 & 0.23 & 5.80 \\

\bottomrule
\end{tabular}
\caption{\textbf{Error counts on OpenWeather evaluated by Gemini-2.5-Pro (Gemma3).}}
\end{table*}

\begin{table*}[htbp!]
\centering
\small
\begin{tabular}{lcccccc}
\toprule
\textbf{Model} & \textbf{Incorrect Fact} & \textbf{Not Checkable} & \textbf{Misleading} & \textbf{Other} & \textbf{All Categories}\\
\midrule
\multicolumn{6}{l}{\textbf{\textit{Zero-Shot (ZS)}}} \\
Gemma3-1B-IT-ZS & 100\% & 41\% & 46\% & 9\% & 100\% \\
Gemma3-27B-IT-ZS & 100\% & 9\% & 46\% & 1\% & 100\% \\

\hdashline
\multicolumn{6}{l}{\textbf{\textit{WebNLG-SFT}}} \\
Gemma3-1B-IT-SFT & 98\% & 14\% & 29\% & 84\% & 100\% \\
Gemma3-27B-IT-SFT & 52\% & 1\% & 45\% & 26\% & 89\% \\

\hdashline
\multicolumn{6}{l}{\textbf{\textit{The best DDKD Model}}} \\
Gemma3-1B-IT-DDKD-Best & 84\% & 2\% & 55\% & 15\% & 99\% \\

\bottomrule
\end{tabular}
\caption{\textbf{Error rates on OpenWeather evaluated by Gemini-2.5-Pro (Gemma3).} }
\label{tab:error_rate_weather_gemini_gemma3}
\end{table*}


\clearpage

\begin{table*}[htbp!]
\centering
\small
\begin{tabular}{lccccccc}
\toprule
\textbf{Model} & \textbf{\# Real} & \textbf{Incorrect Fact} & \textbf{Not Checkable} & \textbf{Misleading} & \textbf{Other} & \textbf{All Categories}\\
\midrule
\multicolumn{6}{l}{\textbf{\textit{Zero-Shot (ZS)}}} \\
GPT-4.1-ZS & -- & 0.01 & 1.04 & 0.00 & 0.00 & 1.05 \\
Qwen3-1.7B-ZS & -- & 0.42 & 0.61 & 0.08 & 0.11 & 1.22 \\
Qwen3-8B-ZS  & -- & 0.21 & 0.79 & 0.10 & 0.70 & 1.17 \\
Qwen3-32B-ZS  & -- & 0.03 & 0.42 & 0.04 & 0.00 & 0.49 \\

\hdashline
\multicolumn{6}{l}{\textbf{\textit{WebNLG-SFT}}} \\
Qwen3-1.7B-SFT  & -- & 0.90 & 0.31 & 0.06 & 0.89 & 2.16 \\
Qwen3-8B-SFT  & -- & 0.51 & 0.15 & 0.11 & 0.59 & 1.36 \\
Qwen3-32B-SFT  & -- & 0.20 & 0.23 & 0.06 & 0.16 & 0.65 \\

\hdashline
\multicolumn{6}{l}{\textbf{\textit{DDKD from ZS teacher}}} \\
Qwen3-1.7B-DDKD-Base  & 100 & 0.38 & 0.49 & 0.04 & 0.12 & 1.03 \\
Qwen3-1.7B-DDKD-Sub  & 100& 0.19 & 0.42 & 0.10 & 0.05 & 0.67 \\
\textbf{Qwen3-1.7B-DDKD-Pert}  & 100& 0.09 & 0.30 & 0.05 & 0.03 & \textbf{0.47} \\
Qwen3-1.7B-DDKD-Mixed  & 100& 0.08 & 0.50 & 0.04 & 0.03 & 0.65 \\
Qwen3-1.7B-DDKD-Base  & 500 & 0.09 & 0.51 & 0.04 & 0.08 & 0.72 \\

\hdashline
\multicolumn{6}{l}{\textbf{\textit{DDKD from WebNLG-SFT teacher}}} \\
Qwen3-1.7B-DDKD-Base  & 100& 0.48 & 0.14 & 0.14 & 0.62 & 1.38 \\
Qwen3-1.7B-DDKD-Sub  & 100& 0.43 & 0.19 & 0.03 & 0.47 & 1.12 \\
Qwen3-1.7B-DDKD-Pert  & 100& 0.50 & 0.18 & 0.07 & 0.50 & 1.25 \\
Qwen3-1.7B-DDKD-Mixed  & 100& 0.30 & 0.15 & 0.08 & 0.34 & 0.87 \\
Qwen3-1.7B-DDKD-Base  & 500 & 0.42 & 0.24 & 0.02 & 0.46 & 1.14 \\

\bottomrule
\end{tabular}
\caption{\textbf{Error counts on GSM Arena evaluated by GPT-5.1 (Qwen3).} We report the average number of errors per output across different error categories (lower is better). \textbf{\# Real} denotes the number of real seed instances used for distillation. "All Categories" represents the total error count. \textbf{Bold} indicates the best performing model (lowest total errors).}
\label{tab:error_count_gsmarena_gpt_qwen3}
\end{table*}

\begin{table*}[htbp!]
\centering
\small
\label{tab:error_rate_gsm_gpt}
\begin{tabular}{lcccccc}
\toprule
\textbf{Model} & \textbf{\# Real} & \textbf{Incorrect Fact} & \textbf{Not Checkable} & \textbf{Misleading} & \textbf{Other} & \textbf{All Categories}\\
\midrule
\multicolumn{6}{l}{\textbf{\textit{Zero-Shot (ZS)}}} \\
GPT-4.1-ZS & -- & 1\% & 57\% & 0\% & 0\% & 57\% \\
Qwen3-1.7B-ZS & -- & 33\% & 45\% & 8\% & 11\% & 71\% \\
Qwen3-8B-ZS & -- & 19\% & 53\% & 9\% & 6\% & 69\% \\
\textbf{Qwen3-32B-ZS} & -- & 3\% & 29\% & 4\% & 0\% & \textbf{34\%} \\

\hdashline
\multicolumn{6}{l}{\textbf{\textit{WebNLG-SFT}}} \\
Qwen3-1.7B-SFT & -- & 56\% & 16\% & 6\% & 52\% & 80\% \\
Qwen3-8B-SFT & -- & 41\% & 14\% & 11\% & 44\% & 76\% \\
Qwen3-32B-SFT & -- & 13\% & 14\% & 6\% & 14\% & 36\% \\

\hdashline
\multicolumn{6}{l}{\textbf{\textit{DDKD from ZS teacher}}} \\
Qwen3-1.7B-DDKD-Base & 100 & 31\% & 36\% & 4\% & 12\% & 62\% \\
Qwen3-1.7B-DDKD-Sub & 100 & 19\% & 33\% & 1\% & 5\% & 47\% \\
Qwen3-1.7B-DDKD-Pert & 100 & 8\% & 26\% & 5\% & 3\% & 39\% \\
Qwen3-1.7B-DDKD-Mixed & 100 & 7\% & 37\% & 4\% & 3\% & 38\% \\
Qwen3-1.7B-DDKD-Base & 500 & 9\% & 35\% & 4\% & 8\% & 49\% \\

\hdashline
\multicolumn{6}{l}{\textbf{\textit{DDKD from WebNLG-SFT teacher}}} \\
Qwen3-1.7B-DDKD-Base & 100 & 36\% & 11\% & 13\% & 39\% & 58\% \\
Qwen3-1.7B-DDKD-Sub & 100 & 24\% & 10\% & 3\% & 35\% & 52\% \\
Qwen3-1.7B-DDKD-Pert & 100 & 22\% & 11\% & 5\% & 33\% & 51\% \\
Qwen3-1.7B-DDKD-Mixed & 100 & 21\% & 11\% & 8\% & 27\% & 51\% \\
Qwen3-1.7B-DDKD-Base & 500 & 22\% & 12\% & 2\% & 34\% & 50\% \\

\bottomrule
\end{tabular}
\caption{\textbf{Error rates on GSM Arena evaluated by GPT-5.1 (Qwen3).} Values represent the percentage of generated outputs containing at least one error of the specific type. \textbf{\# Real} denotes the number of real seed instances used for distillation. The "All Categories" column reports the percentage of outputs containing \textbf{any} type of error. Lower percentages indicate better performance.}
\end{table*}

\clearpage

\begin{table*}[htbp!]
\centering
\small
\begin{tabular}{lcccccc}
\toprule
\textbf{Model} & \textbf{Incorrect Fact} & \textbf{Not Checkable} & \textbf{Misleading} & \textbf{Other} & \textbf{All Categories}\\
\midrule
\multicolumn{6}{l}{\textbf{\textit{Zero-Shot (ZS)}}} \\
Qwen3-1.7B-ZS & 0.65 & 0.25 & 0.29 & 0.15 & 1.34 \\
Qwen3-8B-ZS  & 0.33 & 0.24 & 0.06 & 0.06 & 0.69 \\
\textbf{Qwen3-32B-ZS}  & 0.04 & 0.03 & 0.03 & 0.00 & \textbf{0.10} \\

\hdashline
\multicolumn{6}{l}{\textbf{\textit{WebNLG-SFT}}} \\
Qwen3-1.7B-SFT  & 1.12 & 0.20 & 0.20 & 1.15 & 2.67 \\
Qwen3-8B-SFT  & 0.61 & 0.04 & 0.18 & 0.77 & 1.60 \\
Qwen3-32B-SFT  & 0.26 & 0.13 & 0.13 & 0.15 & 0.67 \\

\hdashline
\multicolumn{6}{l}{\textbf{\textit{DDKD from ZS teacher}}} \\
Qwen3-1.7B-DDKD-Base  & 0.53 & 0.16 & 0.10 & 0.13 & 0.92 \\
Qwen3-1.7B-DDKD-Sub  & 0.29 & 0.12 & 0.12 & 0.08 & 0.61 \\
Qwen3-1.7B-DDKD-Pert  & 0.24 & 0.09 & 0.07 & 0.01 & 0.41 \\
Qwen3-1.7B-DDKD-Mixed & 0.11 & 0.11 & 0.10 & 0.00 & 0.32 \\

\hdashline
\multicolumn{6}{l}{\textbf{\textit{DDKD from WebNLG-SFT teacher}}} \\
Qwen3-1.7B-DDKD-Base  & 0.54 & 0.06 & 0.18 & 0.54 & 1.32 \\
Qwen3-1.7B-DDKD-Sub  & 0.39 & 0.09 & 0.10 & 0.28 & 0.86 \\
Qwen3-1.7B-DDKD-Pert  & 0.58 & 0.08 & 0.16 & 0.47 & 1.29 \\
Qwen3-1.7B-DDKD-Mixed  & 0.29 & 0.08 & 0.12 & 0.28 & 0.77 \\

\bottomrule
\end{tabular}
\caption{\textbf{Error counts on GSM Arena evaluated by Gemini-2.5-Pro (Qwen3).} We report the average number of errors per output across different error categories (lower is better). "All Categories" represents the total error count. \textbf{Bold} indicates the best performing model (lowest total errors).}
\end{table*}

\begin{table*}[htbp!]
\centering
\small
\begin{tabular}{lcccccc}
\toprule
\textbf{Model} & \textbf{Incorrect Fact} & \textbf{Not Checkable} & \textbf{Misleading} & \textbf{Other} & \textbf{All Categories}\\
\midrule
\multicolumn{6}{l}{\textbf{\textit{Zero-Shot (ZS)}}} \\
Qwen3-1.7B-ZS & 46\% & 20\% & 25\% & 13\% & 72\% \\
Qwen3-8B-ZS & 30\% & 21\% & 6\% & 6\% & 47\% \\
\textbf{Qwen3-32B-ZS} & 4\% & 3\% & 3\% & 0\% & \textbf{10\%} \\

\hdashline
\multicolumn{6}{l}{\textbf{\textit{WebNLG-SFT}}} \\
Qwen3-1.7B-SFT & 65\% & 12\% & 17\% & 76\% & 94\% \\
Qwen3-8B-SFT & 45\% & 3\% & 17\% & 54\% & 86\% \\
Qwen3-32B-SFT & 18\% & 9\% & 13\% & 13\% & 45\% \\

\hdashline
\multicolumn{6}{l}{\textbf{\textit{DDKD from ZS teacher}}} \\
Qwen3-1.7B-DDKD-Base & 38\% & 12\% & 9\% & 13\% & 50\% \\
Qwen3-1.7B-DDKD-Sub  & 28\% & 11\% & 12\% & 8\% & 45\% \\
Qwen3-1.7B-DDKD-Pert & 20\% & 6\% & 7\% & 1\% & 31\% \\
Qwen3-1.7B-DDKD-Mixed & 10\% & 9\% & 9\% & 0\% & 26\% \\

\hdashline
\multicolumn{6}{l}{\textbf{\textit{DDKD from WebNLG-SFT teacher}}} \\
Qwen3-1.7B-DDKD-Base & 39\% & 5\% & 16\% & 43\% & 68\% \\
Qwen3-1.7B-DDKD-Sub & 26\% & 4\% & 8\% & 25\% & 43\% \\
Qwen3-1.7B-DDKD-Pert & 27\% & 4\% & 16\% & 36\% & 57\% \\
Qwen3-1.7B-DDKD-Mixed & 22\% & 5\% & 11\% & 24\% & 51\% \\

\bottomrule
\end{tabular}
\caption{\textbf{Error rates on GSM Arena evaluated by Gemini-2.5-Pro (Qwen3).} Values represent the percentage of generated outputs containing at least one error of the specific type. The "All Categories" column reports the percentage of outputs containing \textbf{any} type of error. Lower percentages indicate better performance.}
\end{table*}

\clearpage
\begin{table*}[htbp!]
\centering
\small
\begin{tabular}{lcccccc}
\toprule
\textbf{Model} & \textbf{Incorrect Fact} & \textbf{Not Checkable} & \textbf{Misleading} & \textbf{Other} & \textbf{All Categories}\\
\midrule
\multicolumn{6}{l}{\textbf{\textit{Zero-Shot (ZS)}}} \\
Gemma3-1B-IT-ZS & 1.25 & 2.64 & 0.07 & 0.07 & 4.03 \\
Gemma3-27B-IT-ZS & 0.07 & 0.34 & 0.03 & 0.00 & 0.44 \\

\hdashline
\multicolumn{6}{l}{\textbf{\textit{WebNLG-SFT}}} \\
Gemma3-1B-IT-SFT & 1.90 & 0.75 & 0.13 & 0.82 & 3.60 \\
Gemma3-27B-IT-SFT & 0.15 & 0.13 & 0.07 & 0.02 & 0.37 \\

\hdashline
\multicolumn{6}{l}{\textbf{\textit{The best DDKD Model}}} \\
Gemma3-1B-IT-DDKD-Best & 0.87 & 0.45 & 0.09 & 0.16 & 1.57 \\

\bottomrule
\end{tabular}
\caption{\textbf{Error counts on GSM Arena evaluated by GPT-5.1 (Gemma3).} We report the average number of errors per output across different error categories (lower is better). "All Categories" represents the total error count. \textbf{Bold} indicates the best performing model (lowest total errors).}
\end{table*}

\begin{table*}[htbp!]
\centering
\small
\begin{tabular}{lcccccc}
\toprule
\textbf{Model} & \textbf{Incorrect Fact} & \textbf{Not Checkable} & \textbf{Misleading} & \textbf{Other} & \textbf{All Categories}\\
\midrule
\multicolumn{6}{l}{\textbf{\textit{Zero-Shot (ZS)}}} \\
Gemma3-1B-IT-ZS & 83\% & 89\% & 7\% & 7\% & 98\% \\
Gemma3-27B-IT-ZS & 6\% & 30\% & 3\% & 0\% & 44\% \\

\hdashline
\multicolumn{6}{l}{\textbf{\textit{WebNLG-SFT}}} \\
Gemma3-1B-IT-SFT & 86\% & 47\% & 13\% & 59\% & 96\% \\
Gemma3-27B-IT-SFT & 12\% & 12\% & 7\% & 2\% & 29\% \\

\hdashline
\multicolumn{6}{l}{\textbf{\textit{The best DDKD Model}}} \\
Gemma3-1B-IT-DDKD-Best & 34\% & 33\% & 8\% & 14\% & 59\% \\

\bottomrule
\end{tabular}
\caption{\textbf{Error rates on GSM Arena evaluated by GPT-5.1 (Gemma3).} Values represent the percentage of generated outputs containing at least one error of the specific type. The "All Categories" column reports the percentage of outputs containing \textbf{any} type of error. Lower percentages indicate better performance.}
\end{table*}

\begin{table*}[htbp!]
\centering
\small
\begin{tabular}{lcccccc}
\toprule
\textbf{Model} & \textbf{Incorrect Fact} & \textbf{Not Checkable} & \textbf{Misleading} & \textbf{Other} & \textbf{All Categories}\\
\midrule
\multicolumn{6}{l}{\textbf{\textit{Zero-Shot (ZS)}}} \\
Gemma3-1B-IT-ZS & 1.48 & 1.47 & 0.18 & 0.08 & 3.21 \\
Gemma3-27B-IT-ZS & 0.10 & 0.07 & 0.01 & 0.00 & 0.18 \\

\hdashline
\multicolumn{6}{l}{\textbf{\textit{WebNLG-SFT}}} \\
Gemma3-1B-IT-SFT & 2.29 & 0.45 & 0.27 & 1.13 & 4.14 \\
Gemma3-27B-IT-SFT & 0.23 & 0.08 & 0.05 & 0.03 & 0.39 \\

\hdashline
\multicolumn{6}{l}{\textbf{\textit{The best DDKD Model}}} \\
Gemma3-1B-IT-DDKD-Best & 0.99 & 0.20 & 0.21 & 0.39 & 1.79 \\

\bottomrule
\end{tabular}
\caption{\textbf{Error counts on GSM Arena evaluated by Gemini-2.5-Pro (Gemma3).}}
\end{table*}

\begin{table*}[htbp!]
\centering
\small
\begin{tabular}{lcccccc}
\toprule
\textbf{Model} & \textbf{Incorrect Fact} & \textbf{Not Checkable} & \textbf{Misleading} & \textbf{Other} & \textbf{All Categories}\\
\midrule
\multicolumn{6}{l}{\textbf{\textit{Zero-Shot (ZS)}}} \\
Gemma3-1B-IT-ZS & 84\% & 67\% & 14\% & 8\% & 99\% \\
Gemma3-27B-IT-ZS & 10\% & 7\% & 1\% & 0\% & 15\% \\

\hdashline
\multicolumn{6}{l}{\textbf{\textit{WebNLG-SFT}}} \\
Gemma3-1B-IT-SFT & 90\% & 31\% & 23\% & 69\% & 96\% \\
Gemma3-27B-IT-SFT & 19\% & 8\% & 5\% & 2\% & 32\% \\

\hdashline
\multicolumn{6}{l}{\textbf{\textit{The best DDKD Model}}} \\
Gemma3-1B-IT-DDKD-Best & 44\% & 14\% & 17\% & 22\% & 65\% \\

\bottomrule
\end{tabular}
\caption{\textbf{Error rates on GSM Arena evaluated by Gemini-2.5-Pro (Gemma3).}}
\label{tab:error_rate_gsmarena_gemini_gemma3}
\end{table*}


\clearpage

\begin{table*}[htbp!]
\centering
\small
\begin{tabular}{lccccccc}
\toprule
\textbf{Model} & \textbf{\# Real} & \textbf{Incorrect Fact} & \textbf{Not Checkable} & \textbf{Misleading} & \textbf{Other} & \textbf{All Categories}\\
\midrule
\multicolumn{6}{l}{\textbf{\textit{Zero-Shot (ZS)}}} \\
GPT-4.1-ZS & -- & 0.21 & 0.16 & 0.16 & 0.00 & 0.53 \\
Qwen3-1.7B-ZS & -- & 1.29 & 0.44 & 0.36 & 0.02 & 2.11 \\
Qwen3-8B-ZS  & -- & 0.71 & 0.25 & 0.24 & 0.00 & 1.20 \\
Qwen3-32B-ZS  & -- & 0.68 & 0.05 & 0.25 & 0.01 & 0.99 \\

\hdashline
\multicolumn{6}{l}{\textbf{\textit{WebNLG-SFT}}} \\
Qwen3-1.7B-SFT  & -- & 1.86 & 0.47 & 0.07 & 0.21 & 2.61 \\
Qwen3-8B-SFT  & -- & 0.52 & 0.04 & 0.09 & 0.08 & 0.73 \\
Qwen3-32B-SFT  & -- & 0.28 & 0.03 & 0.04 & 0.08 & 0.43 \\

\hdashline
\multicolumn{6}{l}{\textbf{\textit{DDKD from ZS teacher}}} \\
Qwen3-1.7B-DDKD-Base  & 100 & 1.40 & 0.10 & 0.36 & 0.02 & 1.88 \\
Qwen3-1.7B-DDKD-Sub  & 100 & 0.82 & 0.09 & 0.41 & 0.02 & 1.34 \\
Qwen3-1.7B-DDKD-Pert  & 100 & 1.04 & 0.14 & 0.27 & 0.04 & 1.49 \\
Qwen3-1.7B-DDKD-Mixed  & 100 & 0.88 & 0.09 & 0.42 & 0.00 & 1.39 \\
Qwen3-1.7B-DDKD-Base  & 500 & 0.94 & 0.04 & 0.31 & 0.00 & 1.29 \\

\hdashline
\multicolumn{6}{l}{\textbf{\textit{DDKD from WebNLG-SFT teacher}}} \\
Qwen3-1.7B-DDKD-Base  & 100 & 2.87 & 0.72 & 0.03 & 0.32 & 3.94 \\
Qwen3-1.7B-DDKD-Sub  & 100 & 1.29 & 0.01 & 0.03 & 0.19 & 1.52 \\
Qwen3-1.7B-DDKD-Pert  & 100 & 0.22 & 0.04 & 0.02 & 0.17 & 0.45 \\
Qwen3-1.7B-DDKD-Mixed  & 100 & 0.19 & 0.00 & 0.00 & 0.17 & 0.36 \\
Qwen3-1.7B-DDKD-Base  & 500 & 0.85 & 0.08 & 0.10 & 0.09 & 1.12 \\

\bottomrule
\end{tabular}
\caption{\textbf{Error counts on OWID evaluated by GPT-5.1 (Qwen3).} We report the average number of errors per output across different error categories (lower is better). \textbf{\# Real} denotes the number of real seed instances used for distillation. "All Categories" represents the total error count. \textbf{Bold} indicates the best performing model (lowest total errors).}
\label{tab:error_count_owid_gpt_qwen3}
\end{table*}

\begin{table*}[htbp!]
\centering
\small
\label{tab:accuracy_owid}
\begin{tabular}{lccccccc}
\toprule
\textbf{Model} & \textbf{\# Real} & \textbf{Incorrect Fact} & \textbf{Not Checkable} & \textbf{Misleading} & \textbf{Other} & \textbf{All Categories}\\
\midrule
\multicolumn{6}{l}{\textbf{\textit{Zero-Shot (ZS)}}} \\
GPT-4.1-ZS & -- & 21\% & 12\% & 15\% & 0\% & 38\% \\
Qwen3-1.7B-ZS & -- & 68\% & 42\% & 35\% & 2\% & 87\% \\
Qwen3-8B-ZS & -- & 51\% & 22\% & 22\% & 0\% & 73\% \\
Qwen3-32B-ZS & -- & 48\% & 5\% & 24\% & 1\% & 61\% \\

\hdashline
\multicolumn{6}{l}{\textbf{\textit{WebNLG-SFT}}} \\
Qwen3-1.7B-SFT & -- & 67\% & 18\% & 7\% & 17\% & 77\% \\
Qwen3-8B-SFT & -- & 19\% & 4\% & 9\% & 8\% & 27\% \\
Qwen3-32B-SFT & -- & 12\% & 3\% & 1\% & 8\% & 22\% \\

\hdashline
\multicolumn{6}{l}{\textbf{\textit{DDKD from ZS teacher}}} \\
Qwen3-1.7B-DDKD-Base & 100 & 59\% & 9\% & 28\% & 2\% & 71\% \\
Qwen3-1.7B-DDKD-Sub & 100 & 56\% & 9\% & 33\% & 2\% & 70\% \\
Qwen3-1.7B-DDKD-Pert & 100 & 56\% & 12\% & 26\% & 3\% & 70\% \\
Qwen3-1.7B-DDKD-Mixed & 100 & 50\% & 9\% & 29\% & 0\% & 68\% \\
Qwen3-1.7B-DDKD-Base & 500 & 56\% & 4\% & 29\% & 0\% & 72\% \\

\hdashline
\multicolumn{6}{l}{\textbf{\textit{DDKD from WebNLG-SFT teacher}}} \\
Qwen3-1.7B-DDKD-Base & 100 & 28\% & 13\% & 2\% & 30\% & 47\% \\
Qwen3-1.7B-DDKD-Sub & 100 & 14\% & 1\% & 3\% & 18\% & 31\% \\
Qwen3-1.7B-DDKD-Pert & 100 & 13\% & 4\% & 2\% & 17\% & 32\% \\
Qwen3-1.7B-DDKD-Mixed & 100 & 14\% & 0\% & 0\% & 16\% & 28\% \\
Qwen3-1.7B-DDKD-Base & 500 & 35\% & 6\% & 6\% & 8\% & 42\% \\

\bottomrule
\end{tabular}
\caption{\textbf{Error rates on OWID evaluated by GPT-5.1 (Qwen3).} Values represent the percentage of generated outputs containing at least one error of the specific type. \textbf{\# Real} denotes the number of real seed instances used for distillation. The "All Categories" column reports the percentage of outputs containing \textbf{any} type of error. Lower percentages indicate better performance.}
\end{table*}

\clearpage

\begin{table*}[htbp!]
\centering
\small
\begin{tabular}{lcccccc}
\toprule
\textbf{Model} & \textbf{Incorrect Fact} & \textbf{Not Checkable} & \textbf{Misleading} & \textbf{Other} & \textbf{All Categories}\\
\midrule
\multicolumn{6}{l}{\textbf{\textit{Zero-Shot (ZS)}}} \\
Qwen3-1.7B-ZS & 1.75 & 0.33 & 0.43 & 0.06 & 2.57 \\
Qwen3-8B-ZS  & 1.12 & 0.13 & 0.42 & 0.00 & 1.67 \\
Qwen3-32B-ZS  & 1.18 & 0.03 & 0.37 & 0.02 & 1.60 \\

\hdashline
\multicolumn{6}{l}{\textbf{\textit{WebNLG-SFT}}} \\
Qwen3-1.7B-SFT  & 2.00 & 0.08 & 0.13 & 0.31 & 2.52 \\
Qwen3-8B-SFT  & 0.96 & 0.01 & 0.18 & 0.17 & 1.32 \\
Qwen3-32B-SFT  & 0.64 & 0.00 & 0.08 & 0.35 & 1.07 \\

\hdashline
\multicolumn{6}{l}{\textbf{\textit{DDKD from ZS teacher}}} \\
Qwen3-1.7B-DDKD-Base  & 1.79 & 0.05 & 0.38 & 0.05 & 2.27 \\
Qwen3-1.7B-DDKD-Sub  & 1.45 & 0.07 & 0.47 & 0.04 & 2.03 \\
Qwen3-1.7B-DDKD-Pert  & 1.63 & 0.03 & 0.45 & 0.06 & 2.17 \\
Qwen3-1.7B-DDKD-Mixed  & 1.39 & 0.03 & 0.46 & 0.10 & 1.98 \\

\hdashline
\multicolumn{6}{l}{\textbf{\textit{DDKD from WebNLG-SFT teacher}}} \\
Qwen3-1.7B-DDKD-Base  & 2.60 & 0.02 & 0.08 & 0.52 & 3.22 \\
Qwen3-1.7B-DDKD-Sub  & 1.26 & 0.01 & 0.11 & 0.46 & 1.84 \\
Qwen3-1.7B-DDKD-Pert  & 0.49 & 0.00 & 0.14 & 0.39 & 1.02 \\
Qwen3-1.7B-DDKD-Mixed  & 0.34 & 0.00 & 0.09 & 0.42 & 0.85 \\

\bottomrule
\end{tabular}
\caption{\textbf{Error counts on OWID evaluated by Gemini-2.5-Pro (Qwen3).} We report the average number of errors per output across different error categories (lower is better). "All Categories" represents the total error count. \textbf{Bold} indicates the best performing model (lowest total errors).}
\end{table*}

\begin{table*}[htbp!]
\centering
\small
\label{tab:accuracy_owid_gemini}
\begin{tabular}{lcccccc}
\toprule
\textbf{Model} & \textbf{Incorrect Fact} & \textbf{Not Checkable} & \textbf{Misleading} & \textbf{Other} & \textbf{All Categories}\\
\midrule
\multicolumn{6}{l}{\textbf{\textit{Zero-Shot (ZS)}}} \\
Qwen3-1.7B-ZS & 84\% & 32\% & 40\% & 6\% & 96\% \\
Qwen3-8B-ZS & 65\% & 13\% & 37\% & 0\% & 85\% \\
Qwen3-32B-ZS & 67\% & 3\% & 28\% & 2\% & 79\% \\

\hdashline
\multicolumn{6}{l}{\textbf{\textit{WebNLG-SFT}}} \\
Qwen3-1.7B-SFT & 77\% & 6\% & 12\% & 30\% & 86\% \\
Qwen3-8B-SFT & 32\% & 1\% & 18\% & 15\% & 55\% \\
Qwen3-32B-SFT & 64\% & 0\% & 8\% & 35\% & 52\% \\

\hdashline
\multicolumn{6}{l}{\textbf{\textit{DDKD from ZS teacher}}} \\
Qwen3-1.7B-DDKD-Base & 70\% & 5\% & 32\% & 5\% & 83\% \\
Qwen3-1.7B-DDKD-Sub  & 78\% & 7\% & 39\% & 4\% & 90\% \\
Qwen3-1.7B-DDKD-Pert & 74\% & 3\% & 36\% & 6\% & 86\% \\
Qwen3-1.7B-DDKD-Mixed & 69\% & 3\% & 35\% & 10\% & 85\% \\

\hdashline
\multicolumn{6}{l}{\textbf{\textit{DDKD from WebNLG-SFT teacher}}} \\
Qwen3-1.7B-DDKD-Base & 44\% & 2\% & 8\% & 43\% & 71\% \\
Qwen3-1.7B-DDKD-Sub & 36\% & 1\% & 10\% & 40\% & 68\% \\
Qwen3-1.7B-DDKD-Pert & 30\% & 0\% & 14\% & 33\% & 57\% \\
Qwen3-1.7B-DDKD-Mixed & 23\% & 0\% & 9\% & 37\% & 58\% \\

\bottomrule
\end{tabular}
\caption{\textbf{Error rates on OWID evaluated by Gemini-2.5-Pro (Qwen3).} Values represent the percentage of generated outputs containing at least one error of the specific type. The "All Categories" column reports the percentage of outputs containing \textbf{any} type of error. Lower percentages indicate better performance.}
\end{table*}

\clearpage
\begin{table*}[htbp!]
\centering
\small
\begin{tabular}{lcccccc}
\toprule
\textbf{Model} & \textbf{Incorrect Fact} & \textbf{Not Checkable} & \textbf{Misleading} & \textbf{Other} & \textbf{All Categories}\\
\midrule
\multicolumn{6}{l}{\textbf{\textit{Zero-Shot (ZS)}}} \\
Gemma3-1B-IT-ZS & 2.40 & 0.35 & 0.39 & 0.00 & 3.14 \\
Gemma3-27B-IT-ZS & 0.74 & 0.04 & 0.12 & 0.00 & 0.90 \\

\hdashline
\multicolumn{6}{l}{\textbf{\textit{WebNLG-SFT}}} \\
Gemma3-1B-IT-SFT & 2.76 & 0.28 & 0.32 & 0.50 & 3.86 \\
Gemma3-27B-IT-SFT & 0.60 & 0.00 & 0.03 & 0.14 & 0.77 \\

\hdashline
\multicolumn{6}{l}{\textbf{\textit{The best DDKD Model}}} \\
Gemma3-1B-IT-DDKD-Best & 0.48 & 0.06 & 0.04 & 0.12 & 0.70 \\

\bottomrule
\end{tabular}
\caption{\textbf{Error counts on OWID evaluated by GPT-5.1 (Gemma3).} We report the average number of errors per output across different error categories (lower is better). "All Categories" represents the total error count. \textbf{Bold} indicates the best performing model (lowest total errors).}
\end{table*}

\begin{table*}[htbp!]
\centering
\small
\begin{tabular}{lcccccc}
\toprule
\textbf{Model} & \textbf{Incorrect Fact} & \textbf{Not Checkable} & \textbf{Misleading} & \textbf{Other} & \textbf{All Categories}\\
\midrule
\multicolumn{6}{l}{\textbf{\textit{Zero-Shot (ZS)}}} \\
Gemma3-1B-IT-ZS & 93\% & 28\% & 30\% & 0\% & 97\% \\
Gemma3-27B-IT-ZS & 53\% & 3\% & 12\% & 0\% & 59\% \\

\hdashline
\multicolumn{6}{l}{\textbf{\textit{WebNLG-SFT}}} \\
Gemma3-1B-IT-SFT & 50\% & 23\% & 32\% & 34\% & 92\% \\
Gemma3-27B-IT-SFT & 10\% & 0\% & 3\% & 14\% & 23\% \\

\hdashline
\multicolumn{6}{l}{\textbf{\textit{The best DDKD Model}}} \\
Gemma3-1B-IT-DDKD-Best & 6\% & 5\% & 3\% & 12\% & 22\% \\

\bottomrule
\end{tabular}
\caption{\textbf{Error rates on OWID evaluated by GPT-5.1 (Gemma3).} Values represent the percentage of generated outputs containing at least one error of the specific type. The "All Categories" column reports the percentage of outputs containing \textbf{any} type of error. Lower percentages indicate better performance.}
\end{table*}

\begin{table*}[htbp!]
\centering
\small
\begin{tabular}{lcccccc}
\toprule
\textbf{Model} & \textbf{Incorrect Fact} & \textbf{Not Checkable} & \textbf{Misleading} & \textbf{Other} & \textbf{All Categories}\\
\midrule
\multicolumn{6}{l}{\textbf{\textit{Zero-Shot (ZS)}}} \\
Gemma3-1B-IT-ZS & 2.55 & 0.22 & 0.49 & 0.05 & 3.31 \\
Gemma3-27B-IT-ZS & 1.28 & 0.10 & 0.29 & 0.05 & 1.63 \\

\hdashline
\multicolumn{6}{l}{\textbf{\textit{WebNLG-SFT}}} \\
Gemma3-1B-IT-SFT & 1.90 & 0.11 & 0.39 & 0.56 & 2.96 \\
Gemma3-27B-IT-SFT & 0.51 & 0.00 & 0.16 & 0.56 & 1.23 \\

\hdashline
\multicolumn{6}{l}{\textbf{\textit{The best DDKD Model}}} \\
Gemma3-1B-IT-DDKD-Best & 0.59 & 0.03 & 0.10 & 0.41 & 1.13 \\

\bottomrule
\end{tabular}
\caption{\textbf{Error counts on OWID evaluated by Gemini-2.5-Pro (Gemma3).}}
\end{table*}

\begin{table*}[htbp!]
\centering
\small
\begin{tabular}{lcccccc}
\toprule
\textbf{Model} & \textbf{Incorrect Fact} & \textbf{Not Checkable} & \textbf{Misleading} & \textbf{Other} & \textbf{All Categories}\\
\midrule
\multicolumn{6}{l}{\textbf{\textit{Zero-Shot (ZS)}}} \\
Gemma3-1B-IT-ZS & 97\% & 20\% & 44\% & 5\% & 100\% \\
Gemma3-27B-IT-ZS & 72\% & 1\% & 26\% & 5\% & 83\% \\

\hdashline
\multicolumn{6}{l}{\textbf{\textit{WebNLG-SFT}}} \\
Gemma3-1B-IT-SFT & 72\% & 10\% & 37\% & 48\% & 94\% \\
Gemma3-27B-IT-SFT & 28\% & 0\% & 16\% & 42\% & 68\% \\

\hdashline
\multicolumn{6}{l}{\textbf{\textit{The best DDKD Model}}} \\
Gemma3-1B-IT-DDKD-Best & 23\% & 3\% & 9\% & 36\% & 59\% \\

\bottomrule
\end{tabular}
\caption{\textbf{Error rates on OWID evaluated by Gemini-2.5-Pro (Gemma3).}}
\label{tab:error_rate_owid_gemini_gemma3}
\end{table*}

\clearpage

\begin{table*}[htbp!]
\centering
\small
\setlength{\tabcolsep}{4.0pt}
\renewcommand{\arraystretch}{1.08}
\begin{tabular}{lccccccccccc}
\toprule
\multirow{2}{*}{\textbf{System}} &
\multicolumn{2}{c}{\textbf{Wikidata}} &
\multicolumn{2}{c}{\textbf{IceHockey}} &
\multicolumn{2}{c}{\textbf{GSM}} &
\multicolumn{2}{c}{\textbf{Weather}} &
\multicolumn{2}{c}{\textbf{OWID}} &
\multirow{2}{*}{\textbf{NormAvg}$^\dagger$} \\
\cmidrule(lr){2-3}\cmidrule(lr){4-5}\cmidrule(lr){6-7}\cmidrule(lr){8-9}\cmidrule(lr){10-11}
& \textbf{Score} & \textbf{Ratio}
& \textbf{Score} & \textbf{Ratio}
& \textbf{Score} & \textbf{Ratio}
& \textbf{Score} & \textbf{Ratio}
& \textbf{Score} & \textbf{Ratio}
&  \\
\midrule
Qwen3-1.7B-ZS
& 4.61 & 0.912
& 3.82 & 0.736
& 3.20 & 0.601
& \textbf{2.78} & \textbf{0.463}
& 2.93 & 0.501
& 0.5795 \\
Qwen3-1.7B-SFT
& 4.56 & 0.899
& 3.42 & 0.628
& 3.00 & 0.551
& 2.20 & 0.294
& 2.52 & 0.385
& 0.0000 \\
Qwen3-32B-SFT / ZS
& 4.61 & 0.916
& \textbf{3.99} & \textbf{0.776}
& 3.84 & \textbf{0.761}
& 2.76 & 0.456
& \textbf{3.41} & \textbf{0.609}
& \textbf{0.8908} \\
Qwen3-1.7B-DDKD-Best
& \textbf{4.66} & \textbf{0.926}
& 3.97 & 0.766
& \textbf{3.85} & 0.754
& 2.57 & 0.404
& 2.94 & 0.493
& 0.8150 \\
\bottomrule
\end{tabular}
\caption{
Content coverage sanity check using \textbf{GPT-5.1} as the judge.
Each cell reports mean Coverage Score (1--5) and mean Estimated Coverage Ratio (0--1).
For Wikidata (entity description), the judge evaluates completeness over all input facts; for other domains, it evaluates coverage over task-relevant important content (salient entities/attributes and numeric patterns) without exhaustive row-level matching for long tables.
\textit{Qwen3-32B-SFT/ZS} denotes the teacher variant selected per domain (Table~\ref{tab:final_ddkd_selection}).
$^\dagger$\textbf{NormAvg} is the average across domains of min--max normalized \emph{coverage score} within each domain (normalization computed over the systems in this table; higher is better).
}
\label{tab:coverage_gpt51}
\end{table*}

\begin{table*}[!htbp]
\centering
\small
\setlength{\tabcolsep}{4.0pt}
\renewcommand{\arraystretch}{1.08}
\begin{tabular}{lccccccccccc}
\toprule
\multirow{2}{*}{\textbf{System}} &
\multicolumn{2}{c}{\textbf{Wikidata}} &
\multicolumn{2}{c}{\textbf{IceHockey}} &
\multicolumn{2}{c}{\textbf{GSM}} &
\multicolumn{2}{c}{\textbf{Weather}} &
\multicolumn{2}{c}{\textbf{OWID}} &
\multirow{2}{*}{\textbf{NormAvg}$^\dagger$} \\
\cmidrule(lr){2-3}\cmidrule(lr){4-5}\cmidrule(lr){6-7}\cmidrule(lr){8-9}\cmidrule(lr){10-11}
& \textbf{Score} & \textbf{Ratio}
& \textbf{Score} & \textbf{Ratio}
& \textbf{Score} & \textbf{Ratio}
& \textbf{Score} & \textbf{Ratio}
& \textbf{Score} & \textbf{Ratio}
&  \\
\midrule
Qwen3-1.7B-ZS
& 4.48 & 0.900
& 3.27 & 0.650
& 3.24 & 0.673
& 2.55 & 0.439
& 2.80 & 0.524
& 0.4833 \\
Qwen3-1.7B-SFT
& 4.51 & 0.897
& 2.74 & 0.512
& 3.14 & 0.642
& 1.99 & 0.228
& 2.43 & 0.421
& 0.0857 \\
Qwen3-32B-SFT / ZS
& \textbf{4.55} & \textbf{0.913}
& \textbf{3.48} & \textbf{0.699}
& \textbf{3.85} & \textbf{0.801}
& \textbf{2.64} & \textbf{0.448}
& \textbf{2.96} & \textbf{0.537}
& \textbf{1.0000} \\
Qwen3-1.7B-DDKD-Best
& 4.51 & 0.909
& 3.46 & 0.690
& 3.68 & 0.767
& 2.52 & 0.415
& 2.85 & 0.498
& 0.7540 \\
\bottomrule
\end{tabular}
\caption{
Content coverage sanity check using \textbf{Gemini-2.5-Pro} as the judge.
Each cell reports mean Coverage Score (1--5) and mean Estimated Coverage Ratio (0--1).
For Wikidata (entity description), the judge evaluates completeness over all input facts; for other domains, it evaluates coverage over task-relevant important content (salient entities/attributes and numeric patterns) without exhaustive row-level matching for long tables.
\textit{Qwen3-32B-SFT/ZS} denotes the teacher variant selected per domain (Table~\ref{tab:final_ddkd_selection}).
$^\dagger$\textbf{NormAvg} is the average across domains of min--max normalized \emph{coverage scores} within each domain (normalization computed over the systems in this table; higher is better).
}
\label{tab:coverage_gemini25}
\end{table*}

\end{document}